\PassOptionsToPackage{table}{xcolor}
\documentclass[10pt,twocolumn,letterpaper]{article}

\usepackage{cvpr}              

\usepackage[dvipsnames]{xcolor}

\usepackage{multirow}
\usepackage{cuted}
\usepackage{mathtools}
\usepackage{tikz}
\RequirePackage{snapshot} 

\definecolor{cvprblue}{rgb}{0.21,0.49,0.74}
\usepackage[pagebackref,breaklinks,colorlinks,citecolor=cvprblue]{hyperref}
\usepackage[table,dvipsnames]{xcolor} 
\usepackage{colortbl} 
\usepackage{graphicx}

\newcommand{\parsection}[1]{\noindent\textbf{#1} }

\newcommand{\colorpred}{\hat{\mathbf{C}}}
\newcommand{\colvec}{\mathbf{c}}
\newcommand{\seq}{s}
\newcommand{\seqs}{S}
\newcommand{\rays}{\mathcal{R}}
\newcommand{\ray}{\mathbf{r}}
\newcommand{\loc}{\mathbf{x}}
\newcommand{\dir}{\mathbf{d}}
\newcommand{\ori}{\mathbf{o}}
\newcommand{\propnet}{\sigma_\text{prop}}
\newcommand{\egopose}{\mathbf{P}}
\newcommand{\objpose}{\xi}
\newcommand{\objdim}{\mathbf{s}}
\newcommand{\extrinsic}{\mathbf{T}}
\newcommand{\intrinsic}{\mathbf{K}}
\newcommand{\track}{\mathcal{T}}
\newcommand{\transmit}{U}
\newcommand{\static}{\phi}
\newcommand{\dynamic}{\ensuremath{\psi}}
\newcommand{\hashgrid}[1]{\operatorname{H}_#1}
\newcommand{\mlp}{\operatorname{MLP}}
\newcommand{\loss}{\mathcal{L}}
\newcommand{\real}{\mathbb{R}}
\newcommand{\se}[1]{\mathfrak{se}(#1)}
\newcommand{\SE}[1]{\mathbf{SE}(#1)}
\newcommand{\scenegraph}{\mathcal{G}}
\newcommand{\nodes}{\mathcal{V}}
\newcommand{\edges}{\mathcal{E}}
\newcommand{\appearance}{\mathbf{A}}
\newcommand{\transient}{\mathbf{G}} 
\newcommand{\rot}{\mathbf{R}}
\newcommand{\trans}{\mathbf{t}}

\colorlet{colorFst}{Green!25}       
\colorlet{colorSnd}{SpringGreen!45} 
\colorlet{colorTrd}{Yellow!30}      
\newcommand{\fs}{\cellcolor{colorFst}\bf}   
\newcommand{\nd}{\cellcolor{colorSnd}}      
\newcommand{\rd}{\cellcolor{colorTrd}}      

\def\paperID{6585} 
\def\confName{CVPR}
\def\confYear{2024}

\title{Multi-Level Neural Scene Graphs for Dynamic Urban Environments}

\author{Tobias Fischer$^{1}$ \qquad
 Lorenzo Porzi$^{2}$ \qquad
 Samuel Rota Bul\`{o}$^{2}$ \qquad\\
 Marc Pollefeys$^{1}$ \qquad 
 Peter Kontschieder$^{2}$ \vspace{0.05cm}\\
{ $^{1}$ ETH Z{\"u}rich \quad $^{2}$ Meta Reality Labs}\vspace{0.05cm} \\
   \small{\url{https://tobiasfshr.github.io/pub/ml-nsg/} }
}

\begin{document}
\maketitle

\begin{strip}
\vspace{-42pt}
\centering

   \includegraphics[width=1.0\linewidth]{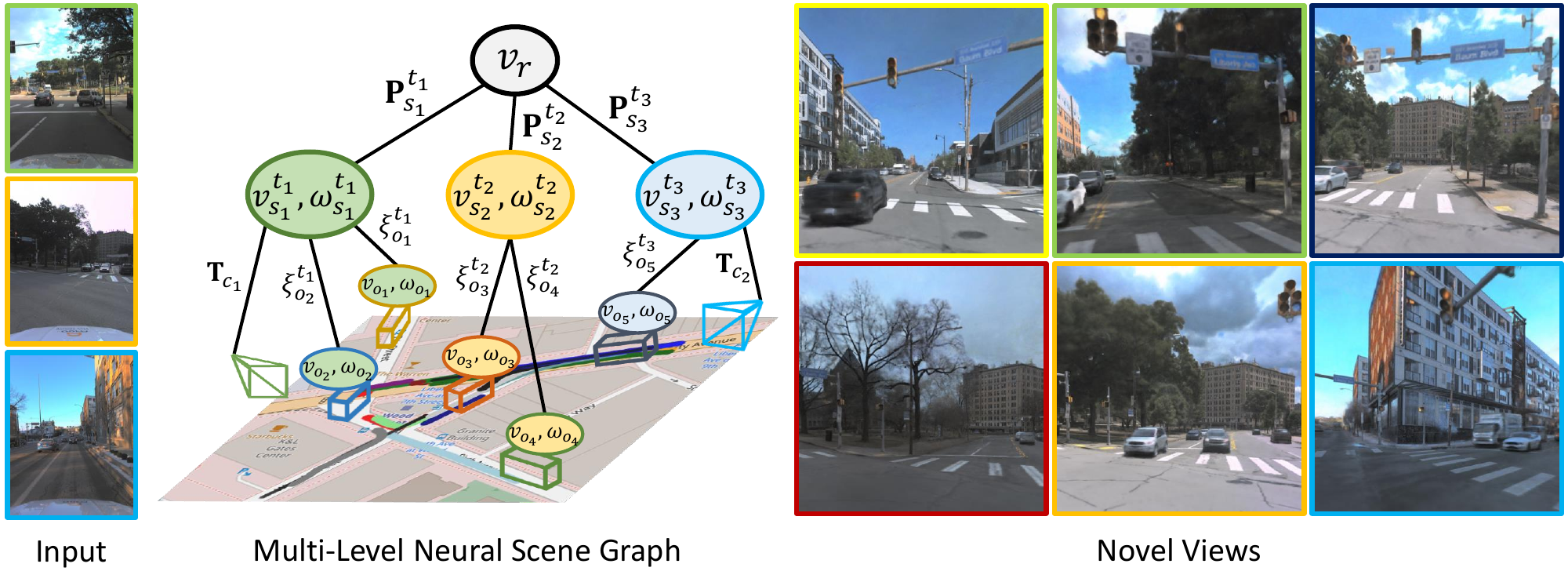}\vspace{-2mm}
   \captionof{figure}{\textbf{Overview.} We represent sequences captured from moving vehicles in a shared geographic area with a multi-level scene graph. Each dynamic object $v_o$ is associated with a sequence node $v_\seq^t$ and time $t$. The sequence nodes are registered in a common world frame at the root node $v_r$ through the vehicle poses $\egopose_\seq^t$, while the dynamic objects are localized w.r.t. the sequence node with pose $\objpose_o^t$. Each camera $c$ is associated with an ego-vehicle position, \ie node $v_\seq^t$, through the extrinsic calibration $\extrinsic_c$. The sequence and object nodes hold latent codes $\omega$ that condition the radiance field, synthesizing novel views in various conditions with distinct dynamic objects.}
   \vspace{-2mm}
\label{fig:teaser}
\end{strip}

\begin{abstract}
We estimate the radiance field of large-scale dynamic areas from multiple vehicle captures under varying environmental conditions.
Previous works in this domain are either restricted to static environments, do not scale to more than a single short video, or struggle to separately represent dynamic object instances. 
To this end, we present a novel, decomposable radiance field approach for dynamic urban environments. 
We propose a multi-level neural scene graph representation that scales to thousands of images from dozens of sequences with hundreds of fast-moving objects.
To enable efficient training and rendering of our representation, we develop a fast composite ray sampling and rendering scheme.
To test our approach in urban driving scenarios, we introduce a new, novel view synthesis benchmark.
We show that our approach outperforms prior art by a significant margin on both established and our proposed benchmark while being faster in training and rendering.

\end{abstract}    
\section{Introduction}
\label{sec:intro}

Estimating the radiance field of a dynamic urban environment from data collected by sensor-equipped vehicles is a core challenge in closed-loop simulation for robotics and in mixed reality. 
It is particularly relevant for applications like autonomous driving and city-scale mapping, where capture vehicles can be frequently deployed.
The growing amount of available data provides great opportunities for creating up-to-date digital twins of entire cities but also poses unique challenges as increasingly heterogeneous data sources must be processed.
In particular, limited scene coverage, different lighting, weather, seasonal conditions, and varying geometry due to distinct dynamic and transient objects make radiance field reconstruction of dynamic urban environments extremely challenging.

Recently, Neural Radiance Fields (NeRFs) have enabled significant progress in achieving realistic novel view synthesis from a set of input views~\cite{sitzmann2019scene, niemeyer2020differentiable,  mildenhall2021nerf, muller2022instant, barron2022mip}. 
These methods represent a static 3D scene with fully implicit~\cite{mescheder2019occupancy, sitzmann2019scene, mildenhall2021nerf} or low-level explicit structures~\cite{liu2020neural, fridovich2022plenoxels, muller2022instant, sun2022direct, xu2022point, gao2023surfelnerf, riegler2021stable, kerbl20233d} such as voxels, points, surfels, meshes or 3D gaussians.
In parallel,  implicit~\cite{li2021neural, xian2021space, park2021hypernerf, wu2022d2nerf, song2023nerfplayer} and low-level explicit~\cite{TiNeuVox, park2023temporal, park2023pointdynrf, luiten2023dynamic} neural representations for dynamic 4D scenes have been investigated.
However, fewer works have focused on decomposing scenes into higher-level entities~\cite{salas2013slam, tulsiani2018factoring, luiten2020track, kundu2022panoptic}.
In computer graphics and 3D mapping, scene graphs have been used to represent complex scenes in a multi-level hierarchy~\cite{sowizral2000scene, armeni20193d, rosinol20203d}.
In view synthesis, Ost~\etal~\cite{ost2021neural} apply this concept by describing actors in a dynamic scene as entities of a scene graph. However, their representation lacks the ability of ~\cite{armeni20193d, rosinol20203d} to represent scenes at multiple levels, thereby limiting their representation to short, single sequences.

While earlier methods for view synthesis focused on object-centric scenes with controlled camera trajectories, recent works move towards radiance field reconstruction of large-scale environments from in-the-wild captures. 
Among these, many methods focus on static environments, removing dynamic actors from the input data~\cite{martin2021nerf, rematas2022urban, tancik2022block, turki2022mega, rudnev2022nerf, lu2023urban, liu2023real}. A few works explicitly model dynamic actors~\cite{ost2021neural, kundu2022panoptic, yang2023unisim}, but either do not scale to more than a single, short video~\cite{ost2021neural, kundu2022panoptic, yang2023unisim}, or struggle to accurately represent individual object instances~\cite{turki2023suds}. 
This complicates the evaluation of these methods for real-world applications because existing benchmarks neither scale to large urban areas~\cite{ost2021neural, kundu2022panoptic} nor reflect realistic capturing conditions~\cite{gao2022monocular}.

To address these issues, we propose a multi-level neural scene graph representation that spans large geographic areas with hundreds of dynamic objects. In contrast to previous works, our multi-level scene graph formulation allows us to distinguish and represent dynamic object instances effectively and further to represent a scene under varying conditions. To make our representation viable for large-scale dynamic environments, we develop a composite ray sampling and rendering scheme that enables fast training and rendering of our method. To test this hypothesis, we introduce a benchmark for radiance field reconstruction in dynamic urban environments based on~\cite{wilson2023argoverse}. We fuse data from dozens of vehicle captures under varying conditions amounting to more than ten thousand images with several hundreds of dynamic objects per reconstructed area. 
We summarize our contributions as follows:
\begin{itemize}
    \item We propose a multi-level neural scene graph formulation that scales to dozens of sequences with hundreds of fast-moving objects under varying environmental conditions.
    \item We develop an efficient composite ray sampling and rendering scheme that enables fast training and rendering of our representation.
    \item We present a benchmark that provides a realistic, application-driven evaluation of radiance field reconstruction in dynamic urban environments.
\end{itemize}
We show state-of-the-art view synthesis results on both established benchmarks~\cite{kitti, cabon2020vkitti2} and our proposed benchmark.

\section{Related Work}
\label{sec:related_work}

\parsection{3D and 4D scene representations.}
Finding the right scene representation is a core issue in 3D computer vision and graphics~\cite{cadena2016past}.
Over the years, a wide variety of options have been explored~\cite{polygonsurface, Bloomenthal1997IntroductionTI, Kaess2015SimultaneousLA, Fairfield2007RealTimeSW, Lu2015VisualNU, Pollefeys2007DetailedRU, salas2013slam, armeni20193d, mescheder2019occupancy, schmied2023r3d3}, and, more recently, neural rendering~\cite{niemeyer2020differentiable} has been used to enable a new generation of scene models that support photo-realistic novel view synthesis.
Scene representations for neural rendering can be roughly classified as ``implicit''~\cite{mescheder2019occupancy, sitzmann2019scene, mildenhall2021nerf}, which store most of the information in the weights of a neural network, or ``explicit'', which use low-level spatial primitives such as voxels~\cite{liu2020neural, fridovich2022plenoxels, muller2022instant, sun2022direct}, points~\cite{xu2022point}, surfels~\cite{gao2023surfelnerf}, meshes~\cite{riegler2021stable}, or 3D gaussians~\cite{kerbl20233d}.

Similarly, different 4D dynamic scene representations for view synthesis have been investigated, including both implicit~\cite{pumarola2021d, tretschk2021nonrigid, xian2021space, park2021hypernerf, li2021neural, gao2021dynamic, wu2022d2nerf, turki2023suds} and explicit~\cite{TiNeuVox, park2023pointdynrf, luiten2023dynamic, yang2023unisim} approaches. Dynamics are generally modeled as deformations of a canonical volume~\cite{pumarola2021d, tretschk2021nonrigid, TiNeuVox, park2021hypernerf, wu2022d2nerf}, a separate scene motion function~\cite{xian2021space, li2021neural, gao2021dynamic, turki2023suds, li2023dynibar}, or rigid transformations of local geometric primitives~\cite{luiten2023dynamic}.

Another line of work investigates the decomposition of scenes into higher-level entities~\cite{salas2013slam, tulsiani2018factoring, luiten2020track, kundu2022panoptic}. To express the composition of different entities into a complex scene, classical computer graphics literature~\cite{cunningham2001lessons} uses scene graphs. In particular, entities are represented as nodes in a hierarchical graph and are connected through edges defined by coordinate frame transformations. Thus, global transformations can be acquired by traversing the graph from its root node.
For indoor 3D mapping, this concept was proposed by Armeni~\etal~\cite{armeni20193d} to represent static scenes at multiple levels of hierarchy, \ie buildings, rooms and objects, and was later extended to dynamic scenes by Rosinol~\etal~\cite{rosinol20203d}.
Recently, Ost~\etal~\cite{ost2021neural} have revisited this concept for view synthesis, describing multi-object scenes with a graph that represents the dynamic actors in the scene. However, their representation lacks the ability of~\cite{armeni20193d, rosinol20203d} to represent scenes at multiple levels of hierarchy and is thus inherently limited to single, short video clips.
On the contrary, we present a scalable, multi-level scene graph representation that spans large geographic areas with hundreds of dynamic objects. 

\parsection{Representing large-scale urban scenes.}
Compared to controlled captures of small scenes, in-the-wild captures of large-scale scenes pose distinct challenges. Limited viewpoint coverage, inaccurate camera poses, far-away buildings and sky, fast-moving objects, complex lighting, and auto exposure make radiance field reconstruction challenging.
Therefore, previous works aid the reconstruction by using depth priors from \eg LiDAR, refining camera parameters, adding information about camera exposure, and using specialized sky and light modeling components~\cite{martin2021nerf, rematas2022urban, tancik2022block, liu2023real, xie2023s, wang2023neural, rudnev2022nerf}.
While many of these works simply remove dynamic actors, some methods explicitly model scene dynamics~\cite{turki2023suds, ost2021neural, kundu2022panoptic, yang2023unisim}. However, these methods either struggle to scale to more than a single sequence~\cite{ost2021neural, kundu2022panoptic, yang2023unisim} or to accurately represent and distinguish dynamic actors~\cite{turki2023suds}. 
To address these issues, we present a decomposed, scalable scene representation for dynamic urban environments.

\parsection{Efficient scene rendering.}
Especially for large-scale scenes, the efficiency of a scene representation in training and inference is crucial. To alleviate the burden of ray traversal in volumetric rendering, many techniques for efficient sampling~\cite{muller2022instant, barron2022mip, fridovich2022plenoxels, turki2022mega} have been proposed. Recently, researchers have exploited more efficient forms of rendering, \eg rasterization of meshes~\cite{lu2023urban, liu2023real} or other 3D primitives~\cite{kerbl20233d}. While these approaches are focused on static scenes, we extend the approach of~\cite{barron2022mip} to our scene graph representation and thus to dynamic scenes.
\section{Data}
\label{sec:data}

\begin{figure}[t] 
  \centering
   \includegraphics[width=1.0\linewidth]{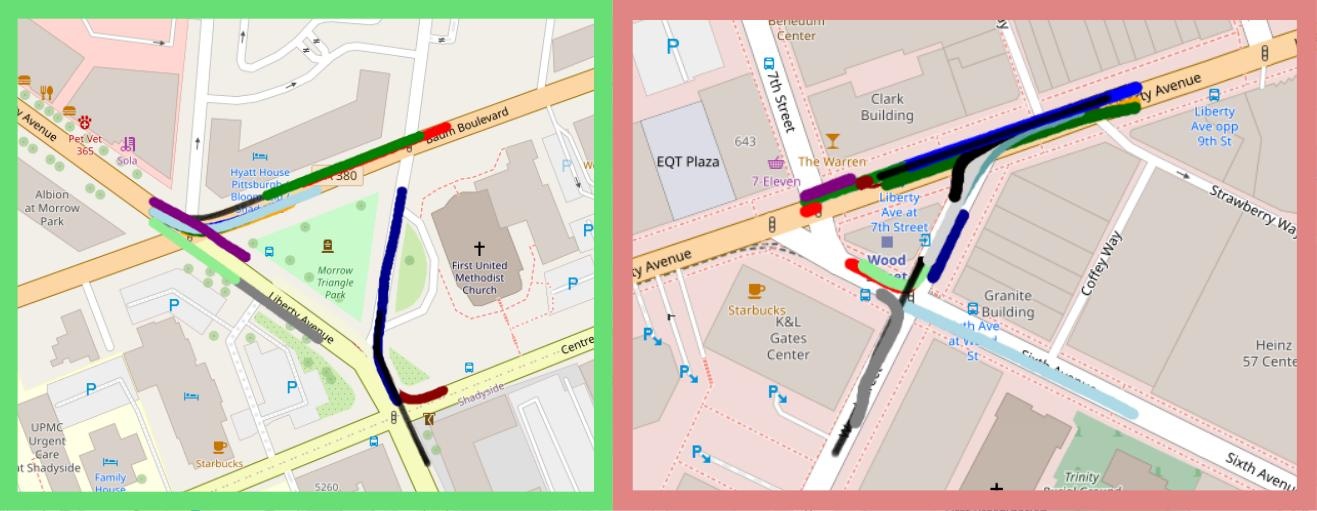}
   \caption{\textbf{Ego-vehicle trajectories of our benchmark.} We show the residential (left) and the downtown (right) areas, trajectories superimposed on 2D maps obtained from OpenStreetMap~\cite{osm}.}
   \label{fig:av2_data_bevs}
\end{figure}

We investigate how to synthesize novel views of a dynamic urban environment from a set of captures taken from moving vehicles. Specifically, we are interested in captures spanning the same geographic area under varying conditions.
This data presents unique challenges due to limited viewpoint coverage, different lighting, weather, and season, and even varying geometry due to distinct dynamic and transient objects. 

Previous benchmarks in this area~\cite{ost2021neural, kundu2022panoptic} are limited to short, single video clips with simple ego trajectories and a few dynamic objects.
Therefore, we present a benchmark that better reflects the aforementioned challenges. We base our benchmark on Argoverse 2~\cite{wilson2023argoverse} which provides a rich set of captures from a fleet of vehicles deployed in multiple US cities that span different weather, season, and time of day.
The vehicles are equipped with a surround-view camera rig with seven global shutter cameras, a LiDAR sensor, and a GPS. 
Furthermore, the timings of the different sensors are provided, so that one can relate the LiDAR-based 3D bounding box annotations to camera timestamps.

\begin{figure}[t] 
  \centering
   \includegraphics[width=1.0\linewidth]{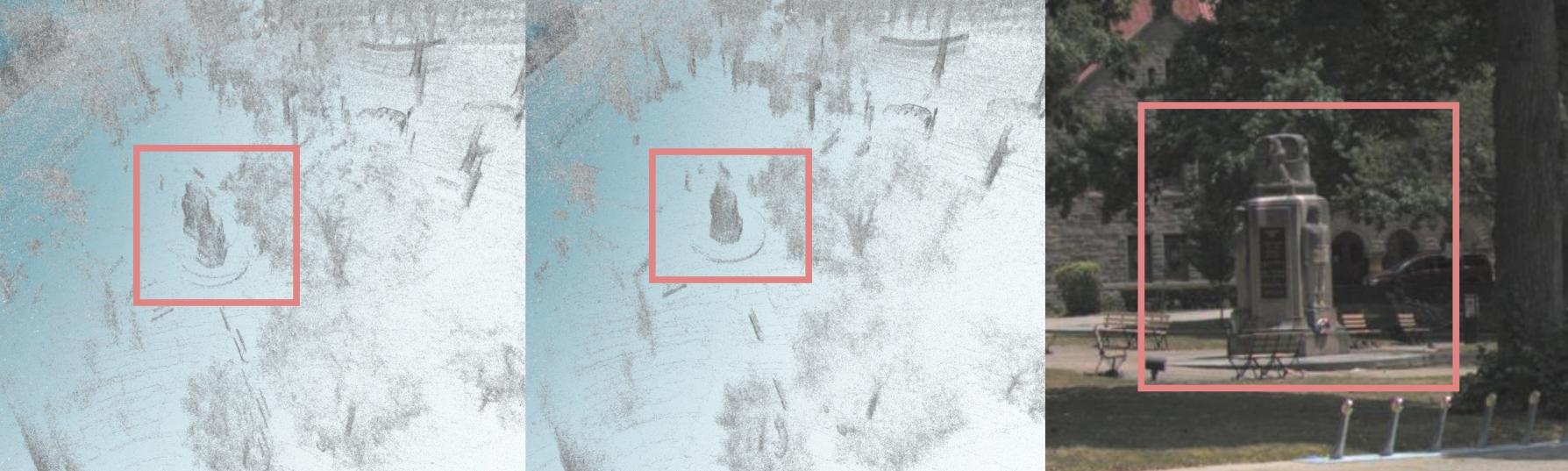}
   \caption{\textbf{Sequence alignment visualization.} The initial GPS-based alignment is imprecise, as evidenced by the duplicated structures in the overlaid LiDAR point clouds (left). After our ICP alignment, the area is well reconstructed (middle) according to its real geometry (right, from Argoverse 2~\cite{wilson2023argoverse}).}
   \label{fig:data_alignment}
\end{figure}

We leverage the GPS information to globally align the sequences and to identify regions that we are interested in mapping. We then extract the specific vehicle captures.
Since GPS-based localization accuracy is only coarsely precise in urban areas, we align the captures via a global, offline iterative-closest-point (ICP) procedure applied to the LiDAR point clouds of all sequences to achieve satisfactory alignment for novel view synthesis purposes (see Fig.~\ref{fig:data_alignment}).

Following this procedure, we build a benchmark that enables real-world evaluation of novel view synthesis from diverse vehicle captures. It is composed of 37 vehicle captures split into two geographic regions as illustrated in Fig.~\ref{fig:av2_data_bevs}. The regions cover a residential and a downtown area to resemble the different characteristics of urban environments.
The residential area spans 14 captures with 10493 training and 1162 testing images with more than 700 distinct moving objects.
The downtown area spans 23 captures with 16933 training and 1876 testing images with more than 600 distinct moving objects.

\section{Method}
\label{sec:method}

\parsection{Problem setup.}
We are given a set of sequences $\seqs$ captured from moving vehicles in different conditions. Each sequence $\seq \in \seqs$ consists of a set of images taken from cameras $C_s$ mounted on the vehicle at different timesteps indexed by $T_\seq$.
We assume the vehicle's sensors are calibrated with respect to a common vehicle and a global world frame. In particular, we assume given sensor extrinsic $\extrinsic_c = [\rot_c | \trans_c ] \in \SE3$ for each camera $c\in C = \bigcup_\seq C_\seq$, and ego-vehicle poses $\egopose_\seq^t = [ \rot_\seq^t | \trans_\seq^t ] \in \SE3$ for each timestamp $t\in T_\seq$ and sequence $\seq \in \seqs$. Furthermore, we assume all cameras to have known intrinsics $\intrinsic_c$.
Each sequence $\seq$ entails a set $O_\seq$ of dynamic objects. For each object $o\in O_\seq$, we assume to have an estimated 3D bounding box track $\track_o$ consisting of object poses $\{ \objpose_o^{t_0}, ..., \objpose_o^{t_n} \} \subset \SE3$ w.r.t. the ego-vehicle frame, where $t_i\in T_s$, and the 3D object dimensions $\objdim_o \in \real^{3}_{>0}$.
The images span a common geographic area for which we would like to estimate a radiance field 
\begin{equation}
    f_\theta(\loc, \dir, t, \seq) = ( \sigma(\loc, t, \seq) , \, \colvec(\loc, \dir, t, \seq))
\end{equation}
that outputs volume density $\sigma \in \real_{\geq 0}$ and color $\colvec \in [0, 1]^3$ conditioned on 3D location $\loc$, viewing direction $\dir$, time $t$ and sequence $\seq$.

\subsection{Multi-Level Neural Scene Graph}
\parsection{Overview.} 
We illustrate our representation in Fig.~\ref{fig:teaser}. We decompose the scene into a graph  $\scenegraph = (\nodes, \edges)$ where the set of nodes $\mathcal V$ is composed by a root node $v_r$ that defines the global coordinate system, \emph{camera} nodes $\{v_c\}_{c\in C}$, and, for each sequence $s\in S$, \emph{sequence} nodes $\{v_\seq^t\}_{t\in T_s}$ and dynamic \emph{object} nodes $\{v_o\}_{o\in O_s}$. The nodes are connected through oriented edges $e \in \edges$ that represent rigid transformations between the coordinate frames of the nodes, consistently with the edge direction. 
In addition, each node $v \in \nodes$ can have an associated latent vector $\omega$.
Given the scene graph $\scenegraph$, we model $f_\theta$ using two radiance fields, namely $\static$ for largely static and $\dynamic$ for highly dynamic scene parts. We use the latent vectors $\omega$ to condition the radiance fields. In particular, we use latent vectors $\omega_\seq^t$ of a sequence node $v_\seq^t$ to condition $\static$:
\begin{equation}
    \static (\loc, \dir, \omega_\seq^t) = ( \sigma_\static(\loc, \omega_\seq^t) , \, \colvec_\static(\loc, \dir, \omega_\seq^t) ) .
\end{equation}
Since the graph $\scenegraph$ has multiple levels, a node associated with a dynamic object will naturally fall into the sequence $s$ it appears in. Therefore, the radiance field $\dynamic$ is conditioned on latent vectors $\omega_\seq^t$ and $\omega_o$:
\begin{equation}
    \dynamic (\loc, \dir, \omega_\seq^t, \omega_o) = ( \sigma_{\dynamic}(\loc, \omega_\seq^t, \omega_o) , \, \colvec_{\dynamic}(\loc, \dir, \omega_\seq^t, \omega_o) )
\end{equation}
of the corresponding sequence and object nodes $v_\seq^t$ and $v_o$.

\parsection{Appearance and geometry variation.}
Reconstructing an environment from multiple captures is challenging from two perspectives: in addition to varying dynamic objects, there is i) varying appearance across captures, and ii) slow-moving or static \textit{transient} geometry such as tree leaves or construction sites.
Since both transient geometry and appearance can vary across captures, but are usually smooth \textit{within} a sequence, we model these phenomena as smooth functions over time
\begin{equation}
   \omega_\seq^t = [\appearance_\seq\mathcal{F}(t), \, \transient_\seq\mathcal{F}(t)]
\end{equation}
where $\mathbf{A}_\seq$ is an appearance matrix, $\mathbf{G}_\seq$ is a transient geometry matrix and $\mathcal{F}(\cdot)$ is a 1D basis function of sines and cosines with linearly increasing frequencies at log-scale~\cite{yang2022banmo, turki2023suds}. We normalize time $t$ into the interval $[-1, 1]$ using the maximum sequence length $\operatorname{max}_{\seq \in \seqs}|T_\seq|$.
Crucially, this allows to model near-static, but sequence-specific regions of the input, as well as appearance changes due to \eg auto-exposure. The degree of variation across time can be controlled by the number of frequencies.

\parsection{Sequence nodes.}
The sequence nodes $v_\seq^t$ are connected to the root node $v_r$ with an edge $e_{v_\seq^t v_r} = \egopose_\seq^t$, \ie the sequences $S$ share a common global world frame. Each sequence node holds the latent vector $\omega_\seq^t$ that conditions the radiance field $\static$. We model $\static$ with a multi-scale 3D hash grid representation~\cite{muller2022instant} and lightweight MLP heads:
\begin{align}
    \mathbf{f}_\loc = \hashgrid{3}(\loc) \\
    \mathbf{h}_{\sigma_\static}, {\sigma_\static} = \mlp_{\sigma_\static}(\mathbf{f}_\loc) \\
    \colvec_\static = \mlp_{\colvec_\static}(\mathbf{h}_{\sigma_\static}, \gamma_\text{SH}(\dir), \mathbf{A}_\seq\mathcal{F}(t)) \label{eq:static_rf} \\
    \sigma_\transient, \colvec_\transient = \mlp_\transient (\mathbf{h}_{\sigma_\static}, \mathbf{G}_\seq\mathcal{F}(t)) \label{eq:transient_rf}
\end{align}
where $\gamma_\text{SH}(\cdot)$ is a spherical harmonics encoding~\cite{muller2022instant}. The final colors and densities are computed as a mixture of static and transient output (analogous to Eq.~\ref{eq:dc_mixture}, see supp. mat.).

\parsection{Dynamic nodes.}
The dynamic nodes $v_o$ are connected to the sequence nodes $v_\seq^t$ with edges $e_{v_o v_\seq^t} = \objpose^t_o$.
The dynamic node $v_o$ associated with object $o$ holds a latent vector $\omega_o$ that conditions the radiance field $\dynamic$. We model $\dynamic$ following~\cite{mildenhall2021nerf} with an MLP conditioned on $\omega_o$ to represent different instances with the same network~\cite{ost2021neural, yang2023unisim}
\begin{align}
    \mathbf{h}_{\sigma_\dynamic}, {\sigma_\dynamic} = \mlp_{\sigma_\dynamic}(\gamma_\text{PE}(\loc), \omega_o) \\
    \colvec_\dynamic = \mlp_{\colvec_\dynamic}(\mathbf{h}_{\sigma_\dynamic}, \gamma_\text{PE}(\dir), \omega_o, \omega_\seq^t) 
\end{align}
where $\gamma_\text{PE}(\cdot)$ is a positional encoding~\cite{mildenhall2021nerf}. Note that 3D position $\loc$ and viewing direction $\dir$ are transformed into the local object coordinate frame with $(\egopose_\seq^t\objpose^t_o)^{-1}\mathbf{I}_3(1 / \operatorname{max}(\objdim_o))$. We condition $\dynamic$ on both the scene and object-dependent latent vectors. This allows us to disentangle scene-dependent appearance from the object texture and thus to transfer objects across sequences. We illustrate this process in Fig.~\ref{fig:car_appearance}.

\parsection{Camera nodes.}
The cameras $c \in C_s$ are connected to the sequence nodes $v_\seq^t$ through edges $e_{v_c v_\seq^t} = \extrinsic_c$, \ie the calibration of camera $c$ w.r.t. the ego-vehicle frame. This way, we can tie camera poses to a specific ego-vehicle pose.

\def \cropcloudl {0px}
\def \cropcloudb {600px}
\def \cropcloudr {0px}
\def \cropcloudt {200px}
\begin{figure}[t]
\centering
\footnotesize
\setlength{\tabcolsep}{1pt}
\resizebox{\linewidth}{!}{
\begin{tabular}{@{}c|c|c@{}}
\includegraphics[width=0.25\linewidth]{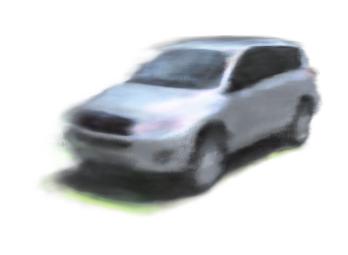} &
\includegraphics[width=0.25\linewidth]{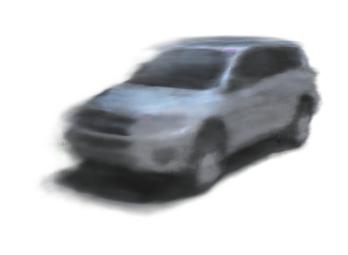} &
\includegraphics[width=0.25\linewidth]{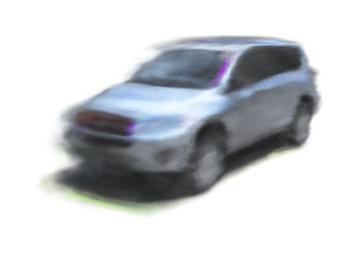}  \\ 
\includegraphics[width=0.22\linewidth]{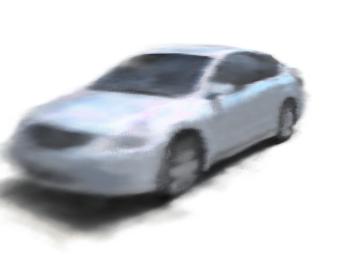} &
\includegraphics[width=0.22\linewidth]{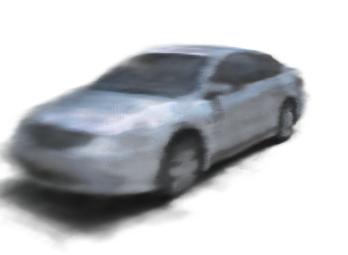} &
\includegraphics[width=0.22\linewidth]{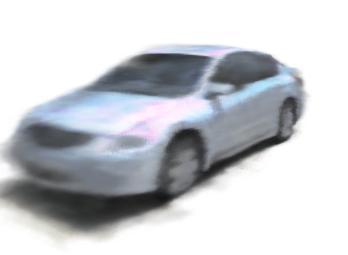}  \\ 
\includegraphics[width=0.33\linewidth,trim={{\cropcloudl} {\cropcloudb} {\cropcloudr} {\cropcloudt}},clip]{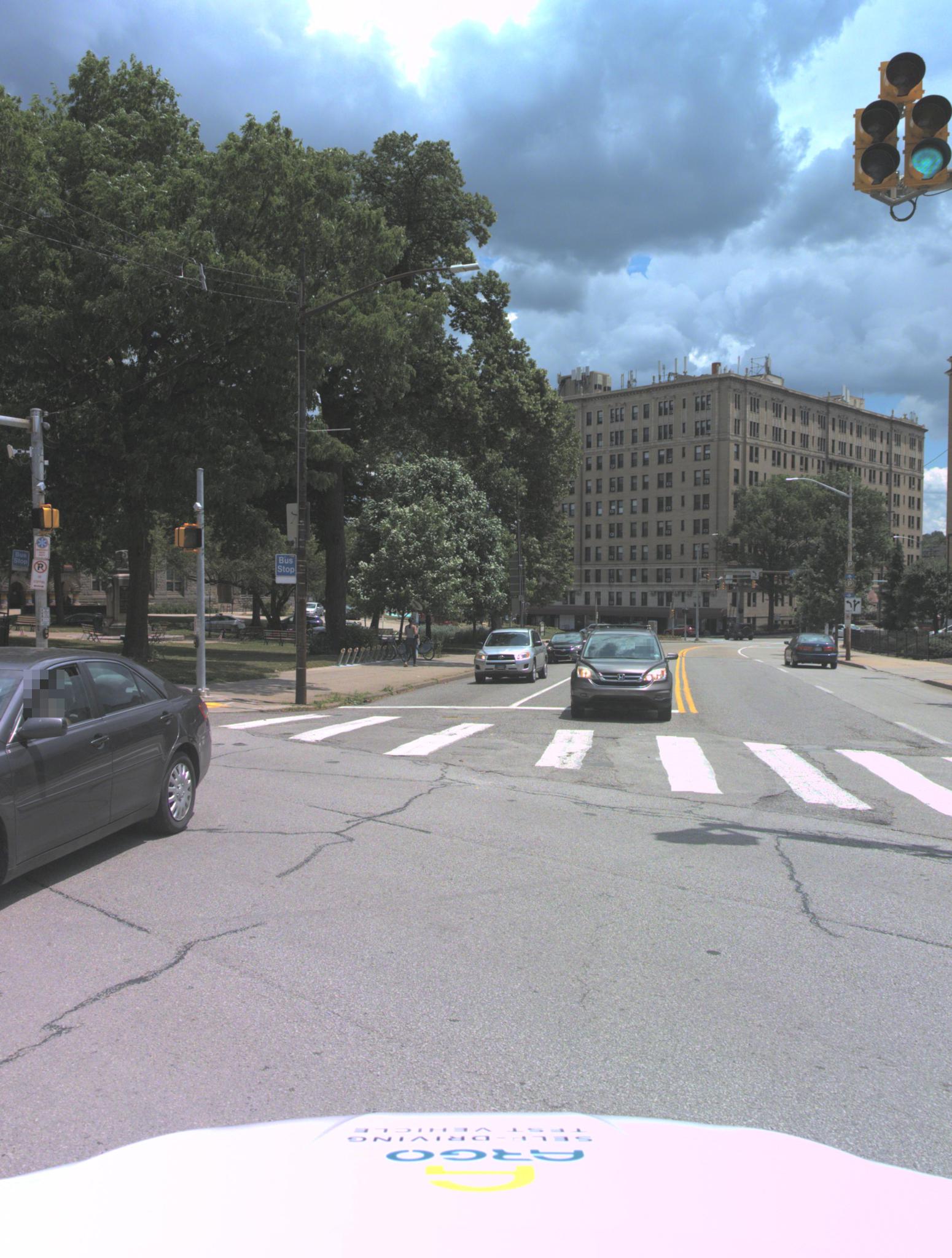} &
\includegraphics[width=0.33\linewidth,trim={{\cropcloudl} {\cropcloudb} {\cropcloudr} {\cropcloudt}},clip]{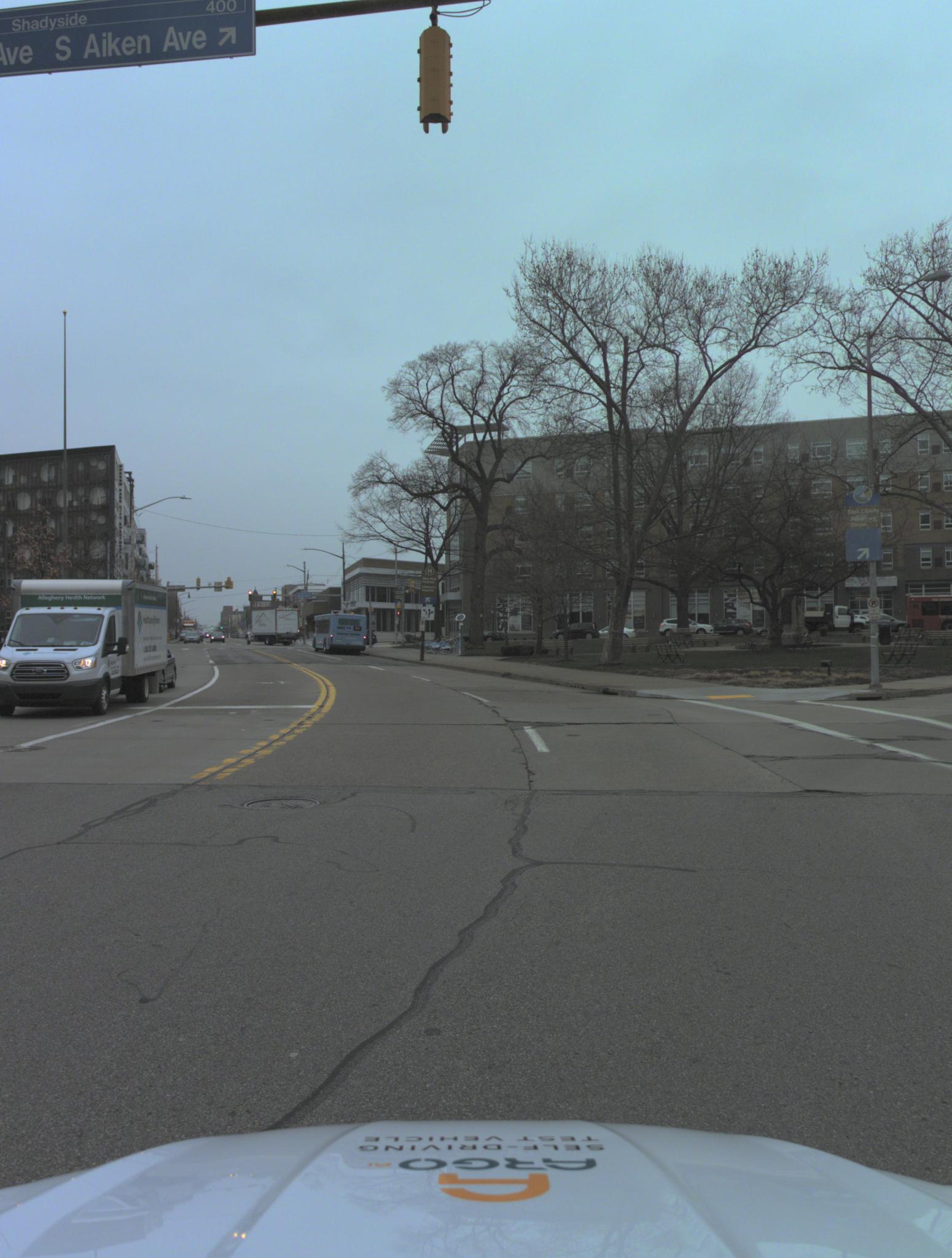} &
\includegraphics[width=0.33\linewidth,trim={{\cropcloudl} {\cropcloudb} {\cropcloudr} {\cropcloudt}},clip]{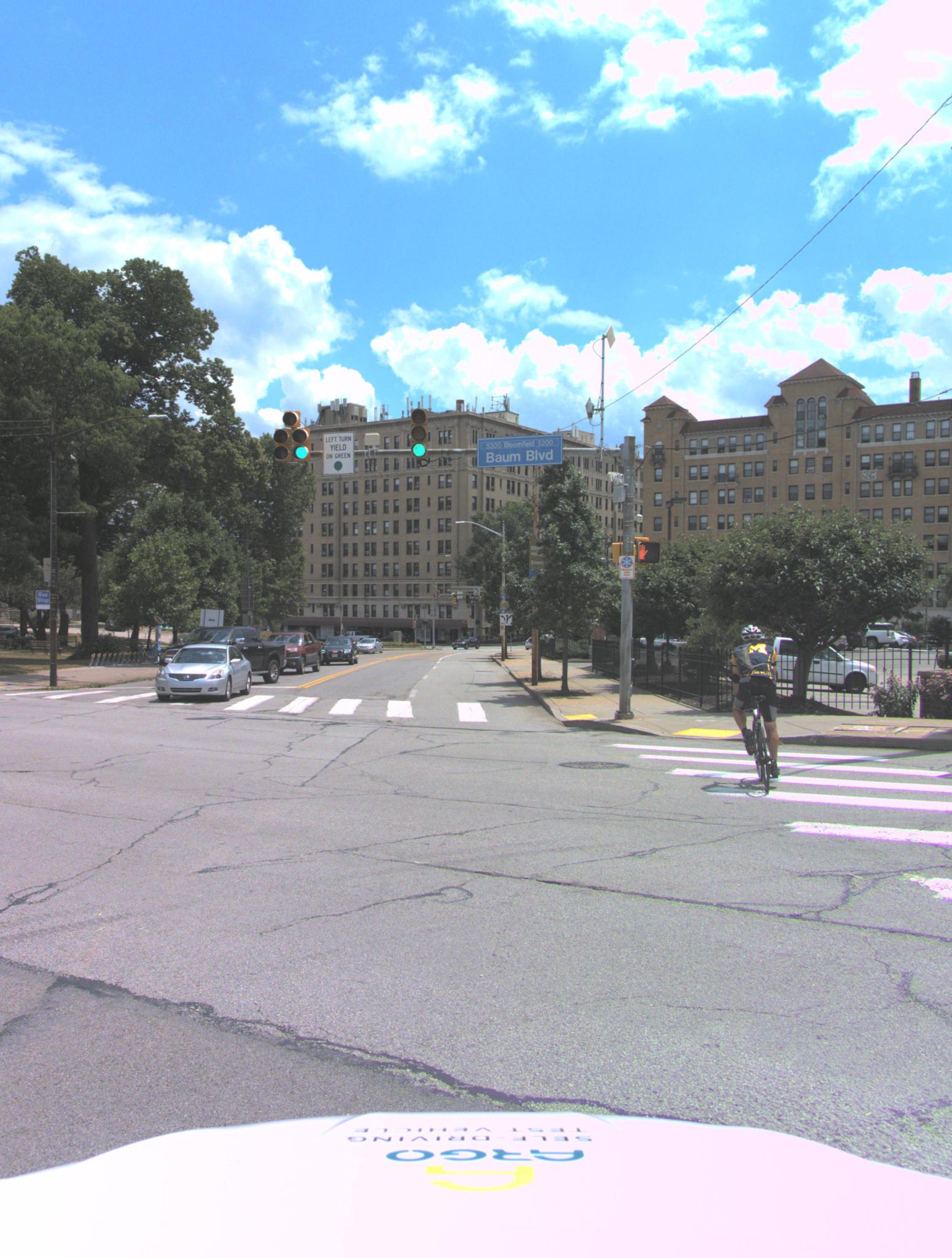} \\ 
\end{tabular}}\vspace{-2mm}
\caption{\textbf{Modifying car appearance with scene appearance.} We exchange $\omega_s^t$ for different car instances. The car's appearance in a rendered, car-centric view (top) changes according to the environmental conditions visible in the sequence $\seq$ (bottom).}
\label{fig:car_appearance}
\vspace{-5mm}
\end{figure}

\subsection{Scene Graph Rendering}

We describe how we render our scene graph $\scenegraph$ for a given set of rays $\rays$. The sampling locations along a ray $(\ray, t, s) \in \rays$ at time $t$ in sequence $s$ are defined as $\ray (u) = \ori + u \dir$ with $\ori = \rot_\seq^t \trans_c + \trans_\seq^t$ and $\dir = \rot_\seq^t \rot_c \intrinsic_c^{-1} (p_x,p_y, 1)^{T}$.

\parsection{Continuous-time pose.}
In order to realistically render videos at different frame rates, we treat the dynamic object poses $\{ \objpose_o^{t_0}, ..., \objpose_o^{t_n} \}$ as a continuous function of time $\objpose_o(t)$. We compute $\objpose_o(t)$ by interpolating between the two nearest poses at $t_a \leq t < t_b$ to time $t$. This allows us also to synchronize estimated object poses originating from the LiDAR measurements with the camera timestamps. 

\parsection{Ray-node intersection.}
To render the dynamic nodes, we measure the intersection of their 3D bounding boxes with each $(\ray, t, s) \in \rays$. 
In particular, given the sequence $\seq$ and time $t$ the ray $\ray$ is associated with, we first traverse the graph $\scenegraph$ to retrieve all relevant nodes $v_o$ and their 3D bounding boxes $\mathbf{b}_o^t = [\objpose_o(t), \objdim_o]$ at time $t$.
Then, we transform $\ray$ into each local node coordinate system and subsequently use AABB-ray intersection~\cite{majercik2018ray} to compute the entry and exit locations $u_o^\text{in}, u_o^\text{out}$ along ray $\ray$.

\parsection{Composite rendering.}
To render the color $\colorpred$ of ray $\ray$ in sequence $\seq$ at time $t$, we use volumetric rendering~\cite{mildenhall2021nerf}
\begin{align}
    \colorpred (\ray, t, \seq) =& \int_{u_n}^{u_f} \transmit(u) \sigma(\ray(u), t, \seq) \colvec(\ray(u), \dir, t, \seq) \,du \notag \\
    \text{where} \, &\transmit(u) = \operatorname{exp}\left(-\int_{u_n}^u\sigma(u') du'\right) .
\end{align}
We obtain density $\sigma$ and color $\colvec$ at sampling location $\ray(u)$ as mixture of the radiance fields $\static$ and $\dynamic$
\begin{equation}
\label{eq:dc_mixture}
    \sigma =  \sigma_\static + \sigma_\dynamic \, , \,
    \colvec = \frac{\sigma_\static}{\sigma_\static + \sigma_\dynamic}\colvec_\static + \frac{\sigma_\dynamic}{\sigma_\static + \sigma_\dynamic}\colvec_\dynamic  .
\end{equation}
Crucially, we set $\sigma_\dynamic$ to zero when $\ray(u)$ does not lie within a 3D bounding box of a dynamic node given the calculated entry and exit points $u^\text{in}, u^\text{out}$.

\begin{figure}[t]
\centering
   \includegraphics[width=0.9\linewidth]{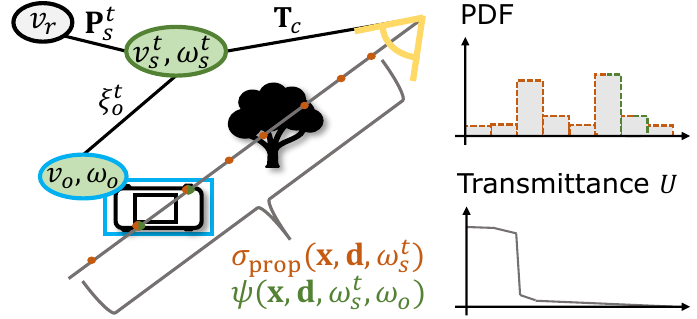}\vspace{-2mm}
   \caption{\textbf{Composite ray sampling.} If a ray intersects with an object $v_o$, we sample from both proposal network $\propnet$ and radiance field $\dynamic$, and $\propnet$ otherwise. We condition each with the latents $\omega$ of the respective nodes. The PDF is a mixture of all node densities that intersect with the ray. The transmittance $\transmit$ drops at the first surface intersection (tree) where further samples will concentrate.}
\label{fig:rendering}
\vspace{-5mm}
\end{figure}

\parsection{Composite ray sampling.}
Instead of densely sampling the space~\cite{kundu2022panoptic, rematas2022urban} or leveraging separate ray sampling mechanisms per node~\cite{ost2021neural}, we use a composite ray sampling strategy illustrated in Fig.~\ref{fig:rendering}.
We extend the proposal sampling mechanism introduced in~\cite{barron2022mip} by joint sampling from a computationally efficient density field $\propnet$ and dynamic nodes $v_o$ at time $t$.
In particular, as in Eq.~\ref{eq:dc_mixture}, we represent $\sigma(\ray(u), t, s)$ by a mixture of $\propnet(\ray(u), \omega_\seq^t)$ and $\sigma_\dynamic(\ray(u), \omega_\seq^t, \omega_o)$. However, analogous to our rendering step, we constrain sampling from $\sigma_\dynamic$ to $[u^\text{in}, u^\text{out}]$.
This allows us to skip empty space efficiently by distilling the static density $\sigma_\static$ into $\propnet$. At the same time, sparsely querying $\sigma_\dynamic$ when $\ray(u)$ falls into a dynamic node enables us to still accurately represent the full, dynamic $\sigma$.

Since there are only few samples that fall into dynamic nodes when performing uniform sampling, the computational overhead in the first ray sampling iteration is negligible.
In the second iteration, we apply inverse transform sampling given the CDF $F(u) = 1 - U(u)$ along ray $\ray$ and thus the samples are concentrated at the first surface intersection. Hence, only for rays that fall in the line of sight of an object surface will we sample $\sigma_\dynamic$ more than a few times. In total, we use two ray sampling iterations as in~\cite{tancik2023nerfstudio}.

\parsection{Space contraction.}
Following~\cite{barron2022mip, tancik2023nerfstudio}, we contract the unbounded scene space into a fixed-size bounding box, normalizing the scene with bounds computed from the LiDAR point clouds and the ego-vehicle poses $\egopose$ into a unit cube.

\vspace{-1mm}
\section{Optimization}
\vspace{-1mm}

\begin{table*}[!htbp]
\centering
\scriptsize
\begin{tabular}{l|cccccc|ccc|c}
\toprule
\multirow{2}{*}{Method} & \multicolumn{3}{c}{Residential} & \multicolumn{3}{c|}{Downtown} & \multicolumn{3}{c|}{Mean} & \multirow{2}{*}{Train time (h)} \\
 &  PSNR $\uparrow$ &SSIM $\uparrow$ & LPIPS $\downarrow$ &  PSNR $\uparrow$ &SSIM $\uparrow$ & LPIPS $\downarrow$ &  PSNR $\uparrow$ &SSIM $\uparrow$ & LPIPS $\downarrow$ \\ \midrule
Nerfacto + Emb. & 19.83 & 0.637 & 0.562  & 18.05 & 0.655 & \rd 0.625 & 18.94 & 0.646 & \rd 0.594 & 8.0 \\
Nerfacto + Emb. + Time & \rd 20.05 & \rd 0.641 & \rd 0.562 & \rd 18.66 & \rd 0.656 & \nd 0.603 & \rd 19.36  & \rd 0.654 & \nd 0.583 & 13.2 \\
SUDS~\cite{turki2023suds} & \nd 21.76 & \nd 0.659 & \nd 0.556 & \nd 19.91 & \nd 0.665 & 0.645 & \nd 20.84 & \nd 0.662 & 0.601 & 54.8\\
\textbf{Ours} & \fs 22.29 & \fs 0.678 & \fs 0.523 & \fs 20.01 & \fs 0.681 & \fs 0.586 & \fs 21.15 & \fs 0.680 & \fs 0.555 & 17.2 \\
\bottomrule
\end{tabular}
\vspace{-2mm}
\caption{\textbf{Novel View Synthesis on Argoverse 2.}  While the static Nerfacto baseline has the weakest performance, the dynamic variant Nerfacto + Time improves only marginally upon it. The state-of-the-art method SUDS exhibits stronger view synthesis results but takes more than $3\times$ longer to train. Our method outperforms all methods across all metrics and exhibits competitive training speed.}
\label{tab:av2_nvs}
\vspace{-4mm}
\end{table*}


\begin{table*}[!htbp]
\centering
\scriptsize
\begin{tabular}
{l|ccccccccc}
\toprule
\multicolumn{1}{l|}{\multirow{2}[3]{*}{Method}} &\multicolumn{3}{c}{KITTI [75\%]} & \multicolumn{3}{c}{KITTI [50\%]} & \multicolumn{3}{c}{KITTI [25\%]} \\ \cmidrule(lr){2-4}\cmidrule(lr){5-7}\cmidrule(lr){8-10}
& PSNR $\uparrow$ & SSIM $\uparrow$ & LPIPS $\downarrow$ & PSNR $\uparrow$ & SSIM $\uparrow$ & LPIPS $\downarrow$ & PSNR $\uparrow$ & SSIM $\uparrow$ & LPIPS $\downarrow$ \\ \midrule
NeRF~\cite{mildenhall2021nerf}\xspace & 18.56 & 0.557 & 0.554 & 19.12 & 0.587 & 0.497 & 18.61 & 0.570 & 0.510 \\
NeRF + Time\xspace & 21.01 & 0.612 & 0.492 & 21.34 & 0.635 & 0.448 & 19.55 & 0.586 & 0.505 \\
NSG~\cite{ost2021neural}\xspace & 21.53 & 0.673 & 0.254 & 21.26 & 0.659 & 0.266 & 20.00 & 0.632 & 0.281 \\
Nerfacto + Emb. & 22.75 &	\rd 0.801 &	\rd 0.156 &	22.38 &	0.793 &	0.160 &	\rd 21.24 &	\rd 0.758 &	\rd 0.178 \\
Nerfacto + Emb. + Time & \nd 23.19 &	\nd 0.804	& \nd 0.155 &	\nd 23.18 &	\rd 0.803 &	\rd 0.155	& \nd 21.98 &	\nd 0.777 &	\nd 0.172 \\
SUDS~\cite{turki2023suds}\xspace  & \rd 22.77 & 0.797 & 0.171 & \rd 23.12 & \nd 0.821 & \nd 0.135 & 20.76 & 0.747  & 0.198 \\
\textbf{Ours} &  \fs 28.38 &	\fs 0.907 &	\fs 0.052 & \fs 27.51 & \fs 0.898 & \fs 0.055 & \fs 26.51 &	\fs 0.887 & \fs	0.060 \\
\midrule
\multirow{2}[3]{*}{Method}&\multicolumn{3}{c}{VKITTI2 [75\%]} & \multicolumn{3}{c}{VKITTI2 [50\%]} & \multicolumn{3}{c}{VKITTI2 [25\%]} \\ \cmidrule(lr){2-4}\cmidrule(lr){5-7}\cmidrule(lr){8-10}
& PSNR $\uparrow$ & SSIM $\uparrow$ & LPIPS $\downarrow$ & PSNR $\uparrow$ & SSIM $\uparrow$ & LPIPS $\downarrow$ & PSNR $\uparrow$ & SSIM $\uparrow$ & LPIPS $\downarrow$ \\ \midrule
NeRF~\cite{mildenhall2021nerf}\xspace & 18.67 & 0.548 & 0.634 & 18.58 & 0.544 & 0.635 & 18.17 & 0.537 & 0.644 \\
NeRF + Time\xspace & 19.03 & 0.574 & 0.587 & 18.90 & 0.565 & 0.610 & 18.04 & 0.545 & 0.626 \\
NSG~\cite{ost2021neural}\xspace & \rd 23.41 & 0.689 & 0.317 & \rd 23.23 & 0.679 & 0.325 & \rd 21.29 & 0.666 & 0.317 \\
Nerfacto + Emb. & 22.15 &	\rd 0.847 &	\rd 0.145 &	21.88 & 0.843 &	0.148 &	21.28 &	\rd 0.827 &	\nd 0.155 \\
Nerfacto + Emb. + Time & 22.11 & \nd 0.849 & \nd 0.144 &	21.78 &	\rd 0.844 & \rd	0.147 &	21.00 & 0.825 &	\rd 0.157 \\
SUDS~\cite{turki2023suds}\xspace & \nd 23.87 & 0.846 &  0.150 & \nd 23.78 & \nd 0.851 & \nd 0.142 & \nd 22.18 & \nd 0.829 &  0.160 \\
\textbf{Ours} & \fs 29.73 &	\fs 0.912 &	\fs 0.065 & \fs 29.19 & \fs 0.906 & \fs 0.066 & \fs 28.29 & \fs 0.901 & \fs 0.067 \\
\bottomrule
\end{tabular}
\vspace{-2mm}
\caption{\textbf{Novel View Synthesis on KITTI and VKITTI2.} We compare our method to prior art, following the experimental protocol in~\cite{turki2023suds}. We test the view synthesis performance of the methods with varying fractions of training views and observe that fewer training views generally result in lower performance. Our method outperforms previous works by a large margin on all settings. }
\label{tab:kitti_nvs}
\vspace{-5mm}
\end{table*}

We optimize the parameters $\theta$ of the radiance field $f_\theta$ with 
\begin{equation}
\label{eq:loss}
\begin{split}
    \loss =& \sum_{(\ray, t, s) \in \mathcal{R}} \loss_\text{rgb}(\ray, t, s) + \lambda_\text{dist} \loss_\text{dist}(\ray, t, s) + \lambda_\text{prop}  \\  & \loss_\text{prop}(\ray, t, s)  + \lambda_\text{dep} \loss_\text{dep}(\ray, t, s) + \lambda_\text{entr} \loss_\text{entr}(\ray, t, s)
\end{split}
\end{equation}
where $\loss_\text{dist}$ and $\loss_\text{prop}$ follow~\cite{barron2022mip}. We supervise $\propnet$ with $\sigma_\static$ \textit{only} to learn effective composite ray sampling.

\parsection{Photometric loss.}
We compare the rendered training rays with their ground truth color
\begin{equation}
    \loss_\text{rgb}(\ray, t, \seq) = || \mathbf{C}(\ray, t, \seq) - \hat{\mathbf{C}}(\ray, t, \seq) ||_2^2 .
\end{equation}

\parsection{Expected depth loss.}
We render the expected depth values for each ray and compare it with the ground truth depth
\begin{equation}
    \loss_\text{dep}(\ray, t, \seq) = || \mathbf{D}(\ray, t, \seq) - \hat{\mathbf{D}}(\ray, t, \seq) ||_2^2 
\end{equation}
where the expected depth is calculated via integrating the sampling values $\hat{\mathbf{D}} (\ray, t, \seq) = \int_{u_n}^{u_f} u\transmit(u) \sigma(\ray(u), t, \seq) \,du$.

\parsection{Entropy regularization.}
While static and dynamic scene parts can overlap, \ie inside a 3D bounding box $\mathbf{b}_o^t$ there could be a part of the street or sidewalk, their density should be strictly separated w.r.t. a single sampling location $\ray(u)$. We leverage an entropy regularization loss~\cite{wu2022d2nerf} to encourage clear separation between entities in $\static$ and $\dynamic$
\begin{equation}
     \loss_\text{entr}(\ray, t, \seq) = \int_{u_n}^{u_f}  \mathcal{H}\left(\dfrac{ \sigma_{\dynamic}(\textbf{r}(u), t, \seq)}{\sigma_\static(\textbf{r}(u), \seq) + \sigma_{\dynamic}(\textbf{r}(u), t, \seq)}\right) \: du
\end{equation}
where $\mathcal{H}(\cdot)$ is the Shannon entropy and we use $t, \seq$ as a replacement for the vectors $\omega_\seq^t, \omega_o$ for ease of notation.

\parsection{Hierarchical pose optimization.}
Alongside scene geometry, we optimize edges $e_{v_\seq^t v_r}$ and $e_{v_o v_\seq^t}$ in our graph, \ie we refine ego-vehicle poses $\egopose_\seq^t \in \SE3$ and object poses $\objpose_o^t \in \SE3$.
In particular, we optimize for pose residuals $\delta \egopose_\seq^t \in \se3$ and $\delta \objpose_o^t \in \se2$. We constrain the object pose residual to $\se2$ since objects are usually upright and move along the ground plane.
Given each residual, we update the ego-vehicle pose with $\hat{\egopose_\seq^t} = \operatorname{expmap}(\delta\egopose_\seq^t) \egopose_\seq^t$ and the object pose with $\hat{\objpose_o^t} = \operatorname{expmap}(\delta\objpose_o^t)_{\SE2 \rightarrow \SE3}\objpose_o^t$,
where $\operatorname{expmap}(\cdot)$ is the exponential map of each lie group.

Compared to naively optimizing camera and object poses, optimizing the edges along our scene graph has two key advantages.
First, we leverage multi-camera constraints, \ie we keep the camera extrinsics $\extrinsic_c$ fixed and optimize the ego-vehicle poses $\egopose_\seq^t$ only. This is in contrast to prior art that generally treats each camera pose as independent.
Second, the residual $\delta\egopose_\seq^t$ propagates to the object poses since they are defined w.r.t. the ego-vehicle coordinate frame instead of the global world frame.
Given that optimizing a radiance field as well as camera and object poses jointly is notoriously difficult~\cite{lin2021barf}, these constraints are vital to view synthesis quality (see Tab.~\ref{tab:pose_optim}).
\vspace{-1mm}
\section{Experiments}
\vspace{-1mm}
\parsection{Datasets.}
To evaluate against competing methods on our proposed benchmark on Argoverse 2~\cite{wilson2023argoverse}, we hold out every 10th sample in uniform time intervals where a sample corresponds to seven ring-camera images.
To compare our methods against prior art on KITTI~\cite{kitti} and VKITTI2~\cite{cabon2020vkitti2}, we follow the experimental protocol and data splits in~\cite{turki2023suds}. We use the provided 3D bounding box annotations for all datasets. We provide results using an off-the-shelf 3D tracker~\cite{yin2021center} in the supplemental material.

\parsection{Metrics.}
Following~\cite{turki2023suds}, we measure image synthesis quality with PSNR, SSIM~\cite{wang2004image}, and LPIPS (AlexNet)~\cite{zhang2018unreasonable}.

\parsection{Implementation details.} 
We implement our method in PyTorch~\cite{paszke2019pytorch}, accelerating the time-consuming ray-node intersection with a custom CUDA implementation. 
We train our model on a single RTX 3090 GPU for 250,000 steps on Argoverse 2 and 100,000 steps on KITTI and VKITTI2, with 8192 rays per batch.
All model parameters are optimized using Adam~\cite{kingma2014adam} with $\beta_1 = 0.9, \beta_2=0.999$, and an exponential decay learning rate schedule: from $10^{-5}$ to $10^{-6}$ for pose parameters, and from $10^{-3}$ to $10^{-4}$ for all others.
In order to counter pose drift, we further apply weight decay with a factor $10^{-2}$ to $\delta\egopose$ and $\delta\objpose$.
The static radiance field $\static$ is trained from scratch, while $\dynamic$ is initialized from a semantic prior~\cite{chang2015shapenet} following~\cite{kundu2022panoptic}.

\parsection{Baselines.}
We compare our method against prior work in dynamic outdoor scene representation, \ie SUDS~\cite{turki2023suds} and NSG~\cite{ost2021neural}. Further, we include results obtained by augmenting Nerfacto~\cite{tancik2023nerfstudio}, the closest state-of-the-art NeRF architecture to ours, with components designed to handle multi-capture reconstruction and scene dynamics.
In particular, we consider two variants: ``Nerfacto + Emb.'', where we incorporate our appearance embedding $\appearance_\seq\mathcal{F}(t)$ and the expected depth loss; and ``Nerfacto + Emb. + Time'', where we further include time modeling with 4D hash tables following~\cite{park2023temporal, xian2021space}.
For both Nerfacto variants, we set hash table and MLP sizes to be aligned with our method.

\begin{table}
\centering
\resizebox{1.0\linewidth}{!}{
\begin{tabular}{@{}lccccccc}
\toprule
& SRN~\cite{sitzmann2019scene} & NeRF~\cite{mildenhall2021nerf}  & NeRF + Time & NSG~\cite{ost2021neural}   & PNF~\cite{kundu2022panoptic} & SUDS~\cite{turki2023suds} & \textbf{Ours} \\ \midrule
PSNR $\uparrow$ & 18.83 & 23.34 & 24.18 & 26.66 & \rd 27.48 & \nd 28.31 & \fs 29.36 \\
SSIM $\uparrow$ & 0.590 & 0.662 & 0.677 & 0.806 & \rd 0.870 & \nd 0.876 & \fs 0.911 \\\bottomrule
\end{tabular}
}\vspace{-2mm}
\caption{\textbf{Image reconstruction on KITTI.} We outperform prior art in image reconstruction, \ie our method can better fit the training views. We follow the experimental protocol in~\cite{ost2021neural, kundu2022panoptic, turki2023suds}.}
\label{tab:kitti_imrec}
\vspace{-3mm}
\end{table}

\begin{figure}[t]
\centering
\footnotesize
\setlength{\tabcolsep}{1pt}
\resizebox{1.0\linewidth}{!}{
\begin{tabular}{@{}cc|c@{}}
\rotatebox{90}{\hspace{3mm}Ours} & \includegraphics[width=0.45\linewidth]{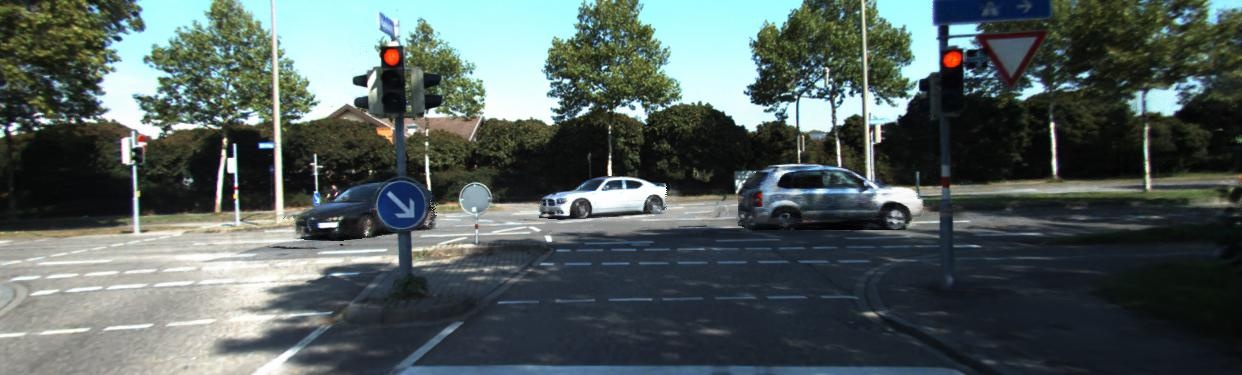} & \includegraphics[width=0.45\linewidth]{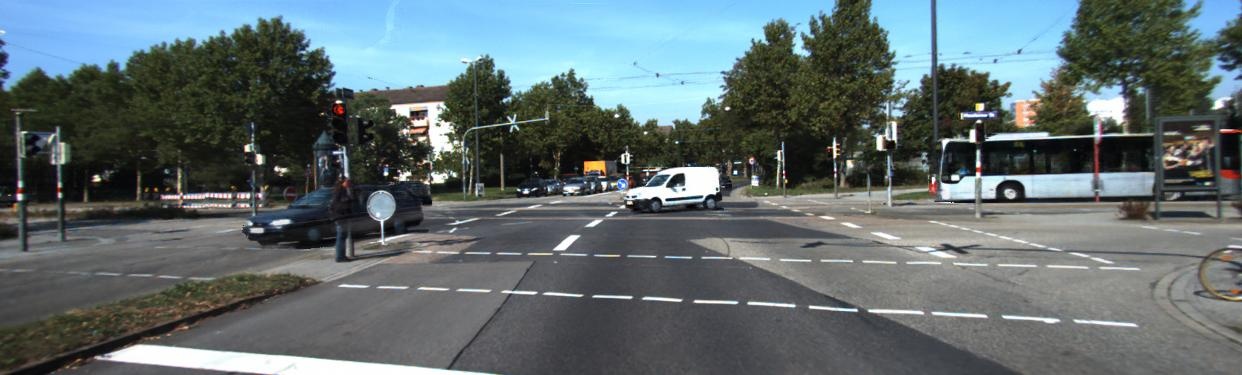}  \\
\rotatebox{90}{\hspace{4mm}GT} & \includegraphics[width=0.45\linewidth]{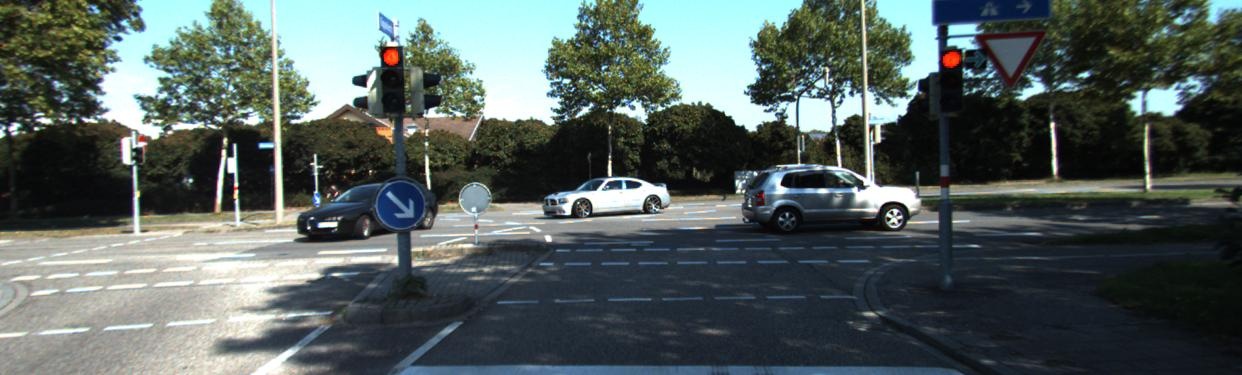} &  \includegraphics[width=0.45\linewidth]{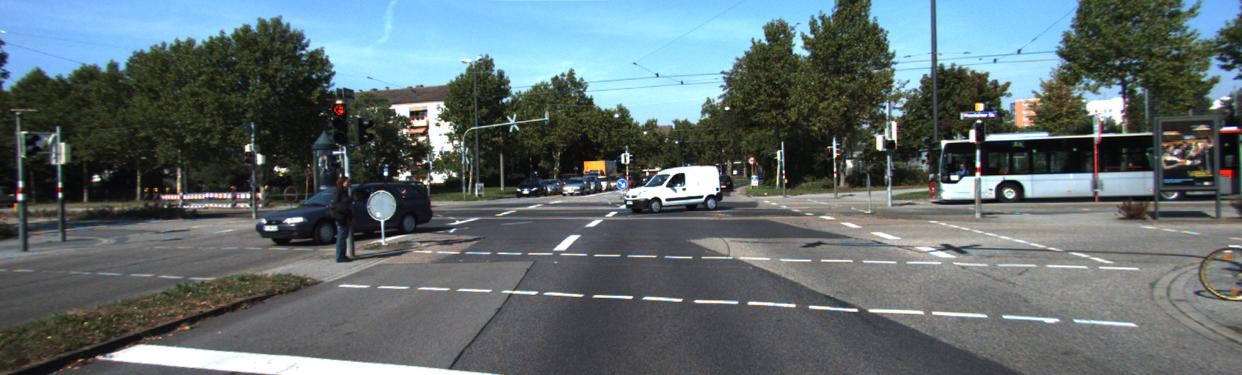} \\
\end{tabular}}\vspace{-2mm}
\caption{\textbf{Qualitative results on KITTI.} With only 25\%  of views (approx. 15-20) for training, we can still synthesize sharp and realistic novel views with dynamic objects rendered at high quality.}
\label{fig:kitti_qualitative}
\vspace{-5mm}
\end{figure}

\subsection{Comparison to state-of-the-art}
We first compare to prior work on our proposed benchmark in Tab~\ref{tab:av2_nvs}. We observe that the static Nerfacto + Emb. has the weakest performance, which slightly increases by adding time modeling in Nerfacto + Emb. + Time. Meanwhile, the current state-of-the-art method SUDS~\cite{turki2023suds} outperforms the Nerfacto variants, while being much slower to train. Our method performs the best on all metrics. The improvement is particularly pronounced in the perceptual quality metrics (SSIM and LPIPS). The training speed of our method is competitive to Nerfacto + Emb. + Time and more than $3\times$ faster than SUDS. 

\begin{figure*}[t]
\centering
\footnotesize
\setlength{\tabcolsep}{1pt}
\resizebox{1.0\linewidth}{!}{

\begin{tabular}{@{}ccc|cc|cc@{}}

  \rotatebox{90}{\hspace{5mm}Ground Truth} & \includegraphics[width=0.2\linewidth]{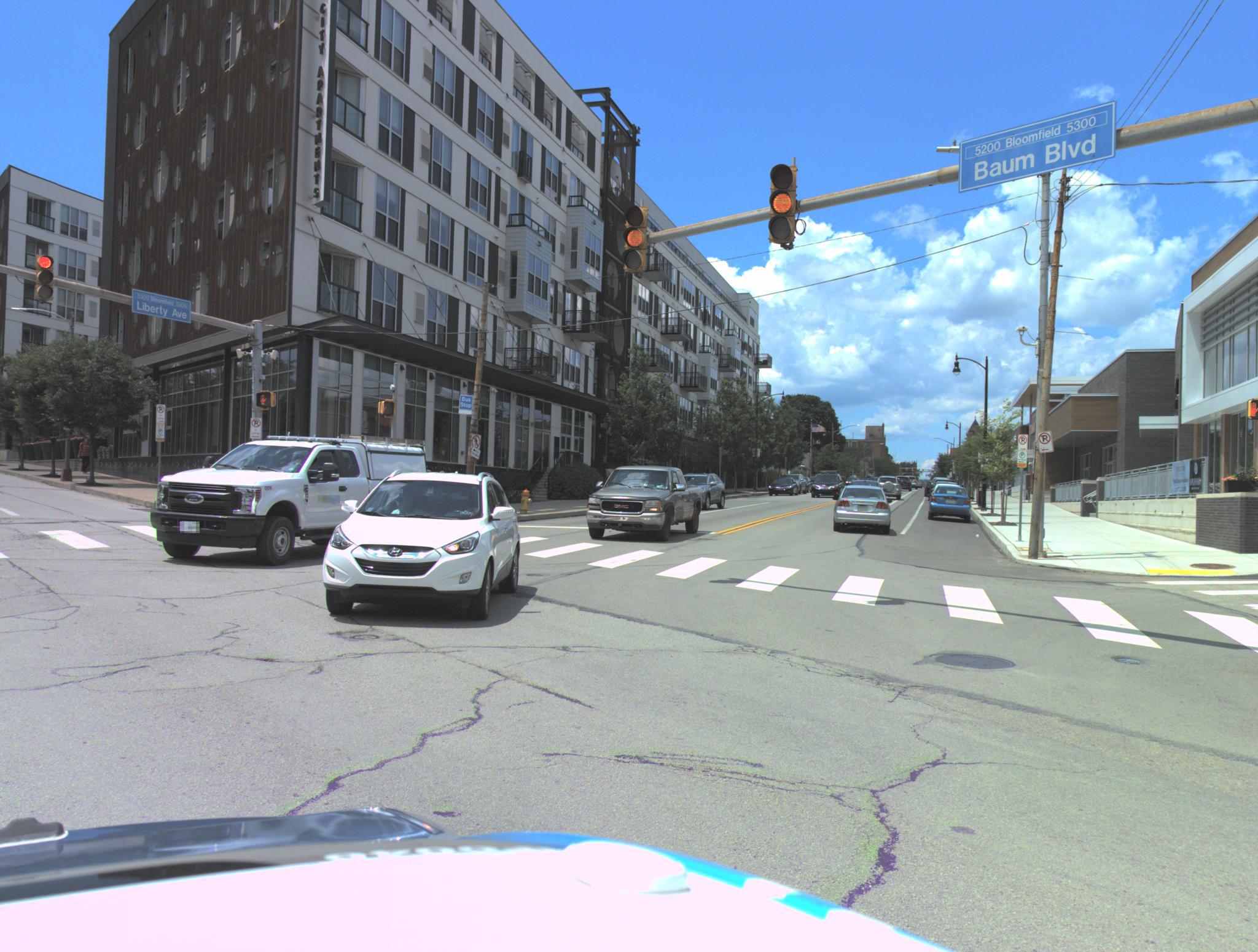} &\includegraphics[width=0.2\linewidth]{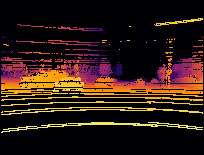} &  \includegraphics[width=0.2\linewidth]{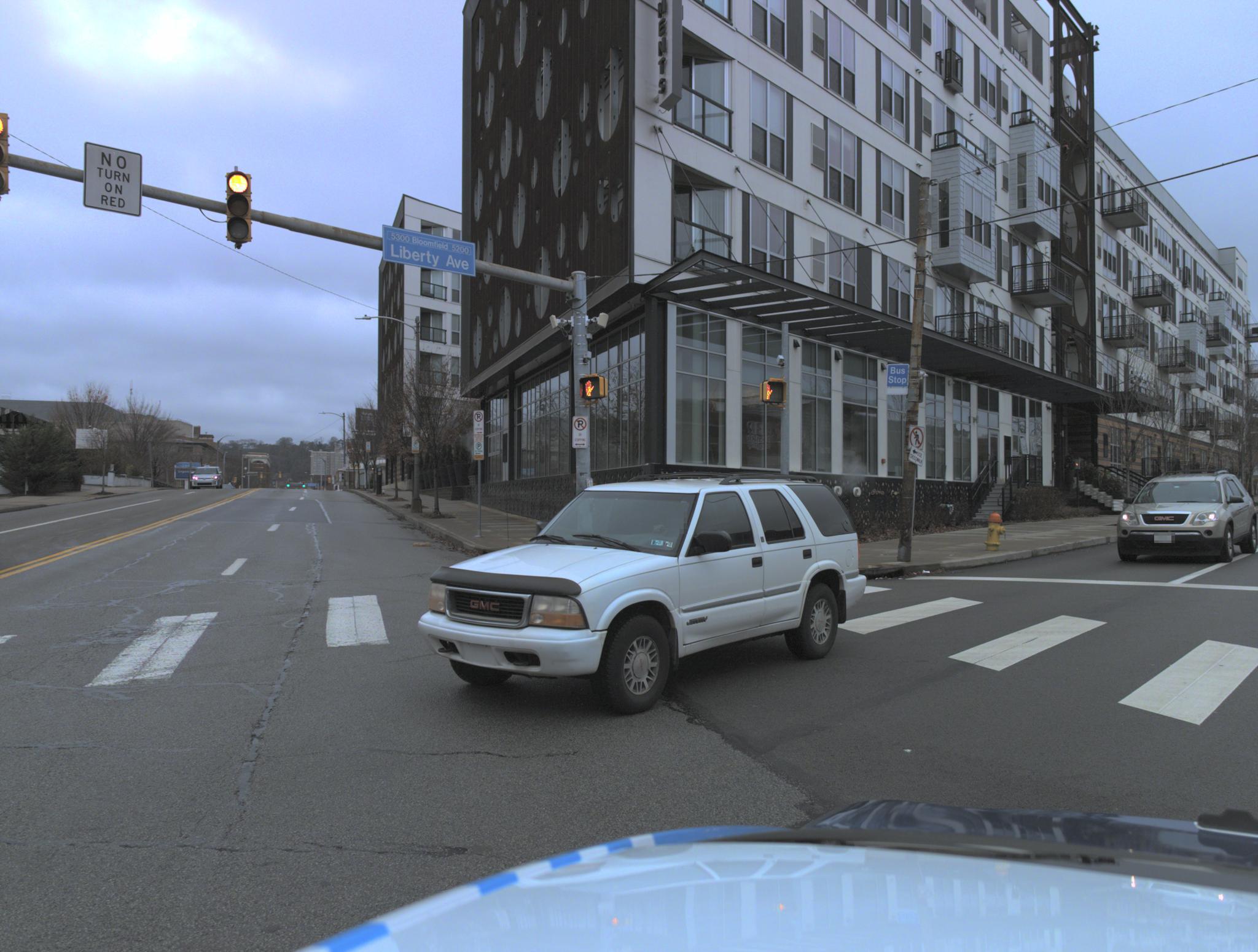} &  \includegraphics[width=0.2\linewidth]{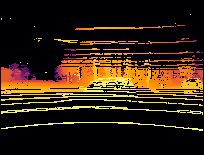} &  \includegraphics[width=0.2\linewidth]{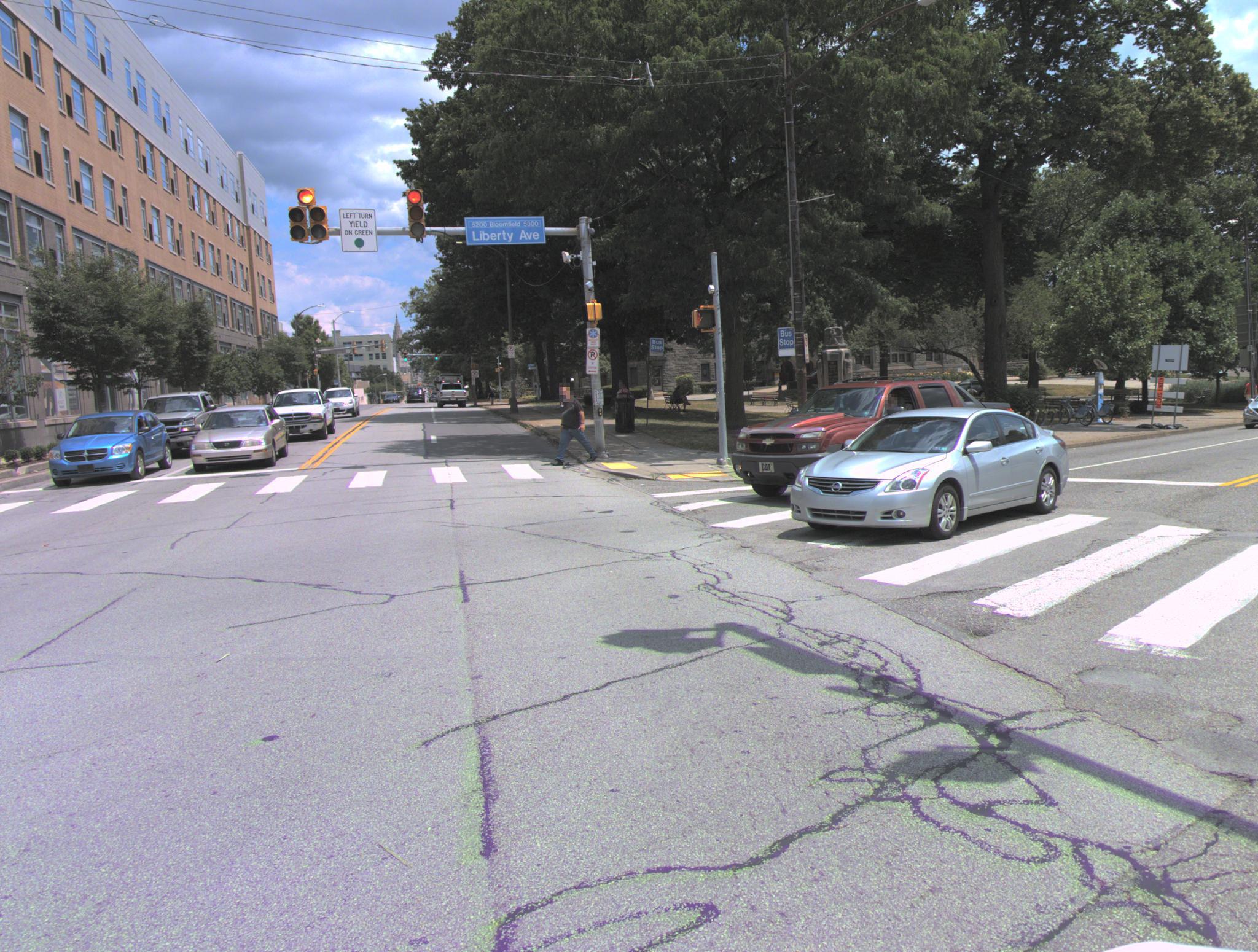} &  \includegraphics[width=0.2\linewidth]{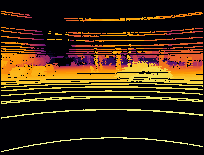} \\
  
  \rotatebox{90}{\hspace{10mm}Ours} & \includegraphics[width=0.2\linewidth]{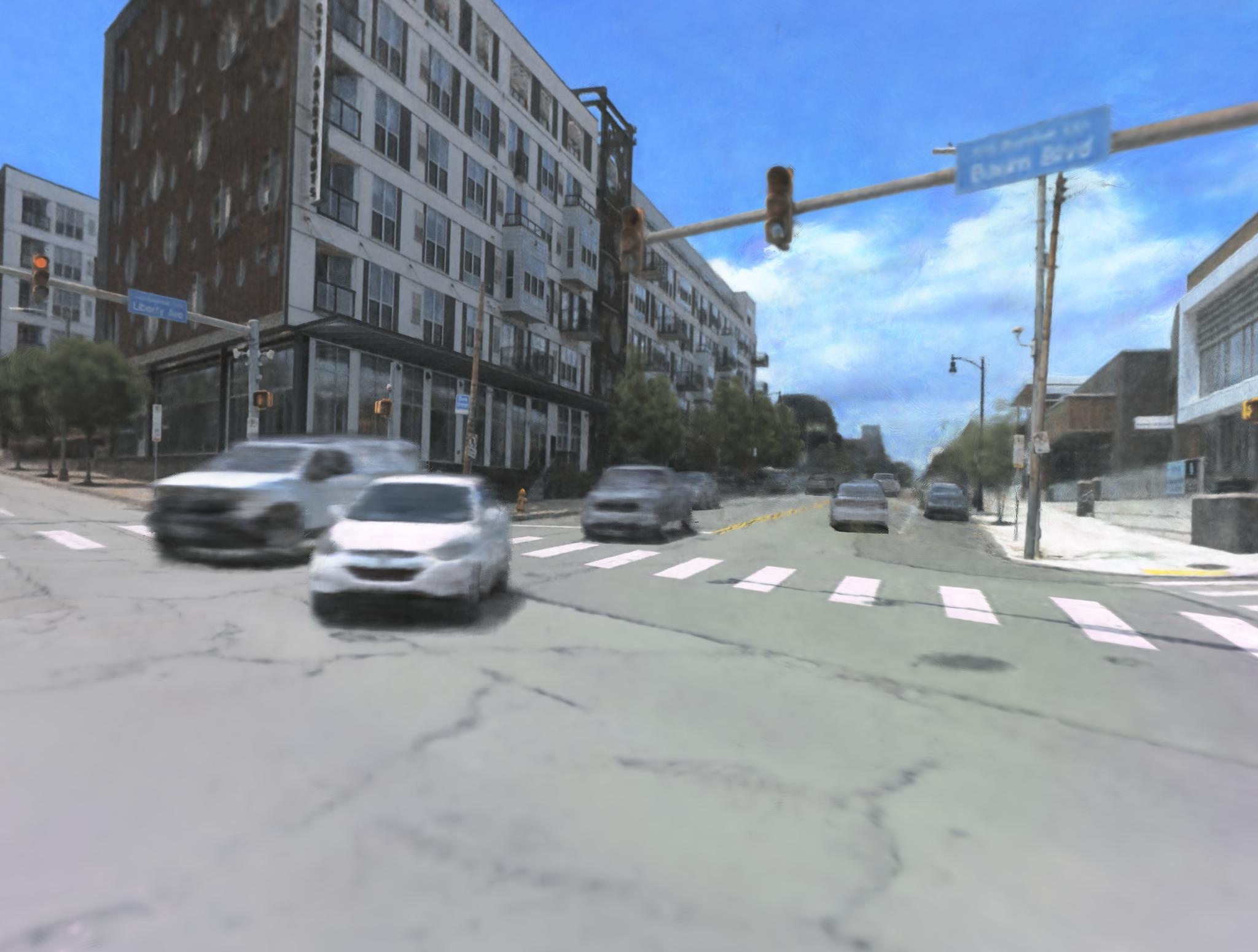} &\includegraphics[width=0.2\linewidth]{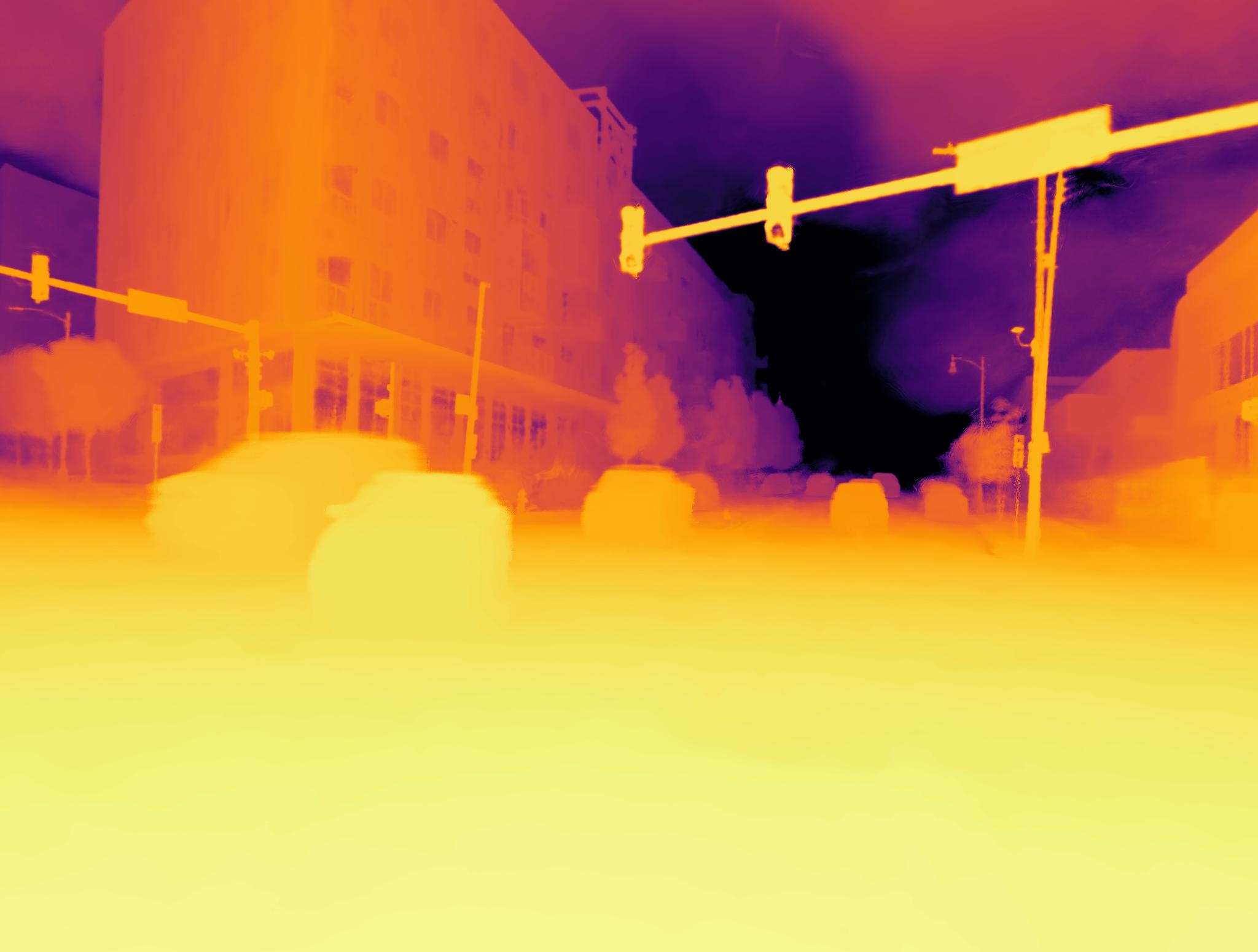} &  \includegraphics[width=0.2\linewidth]{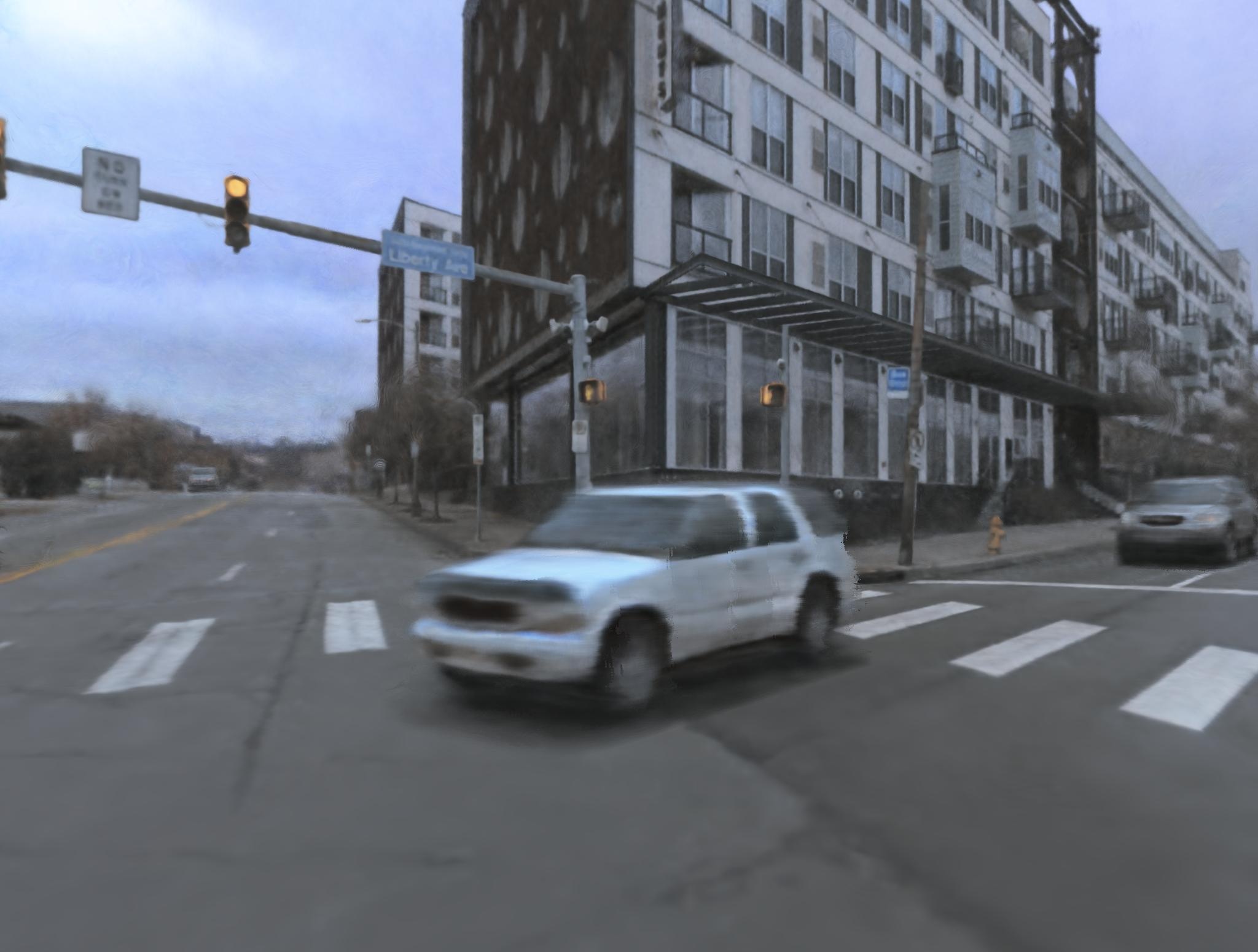} &  \includegraphics[width=0.2\linewidth]{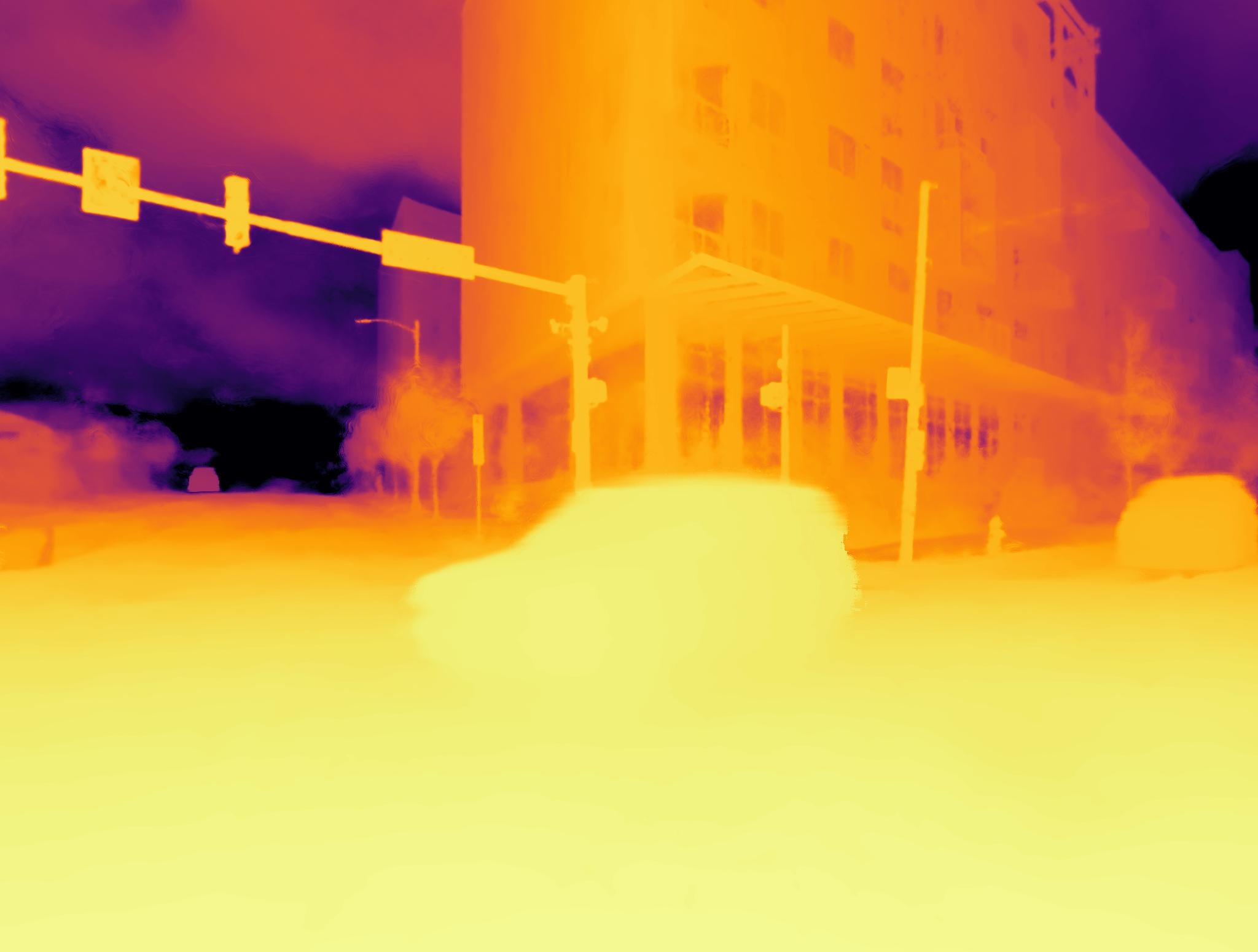} &  \includegraphics[width=0.2\linewidth]{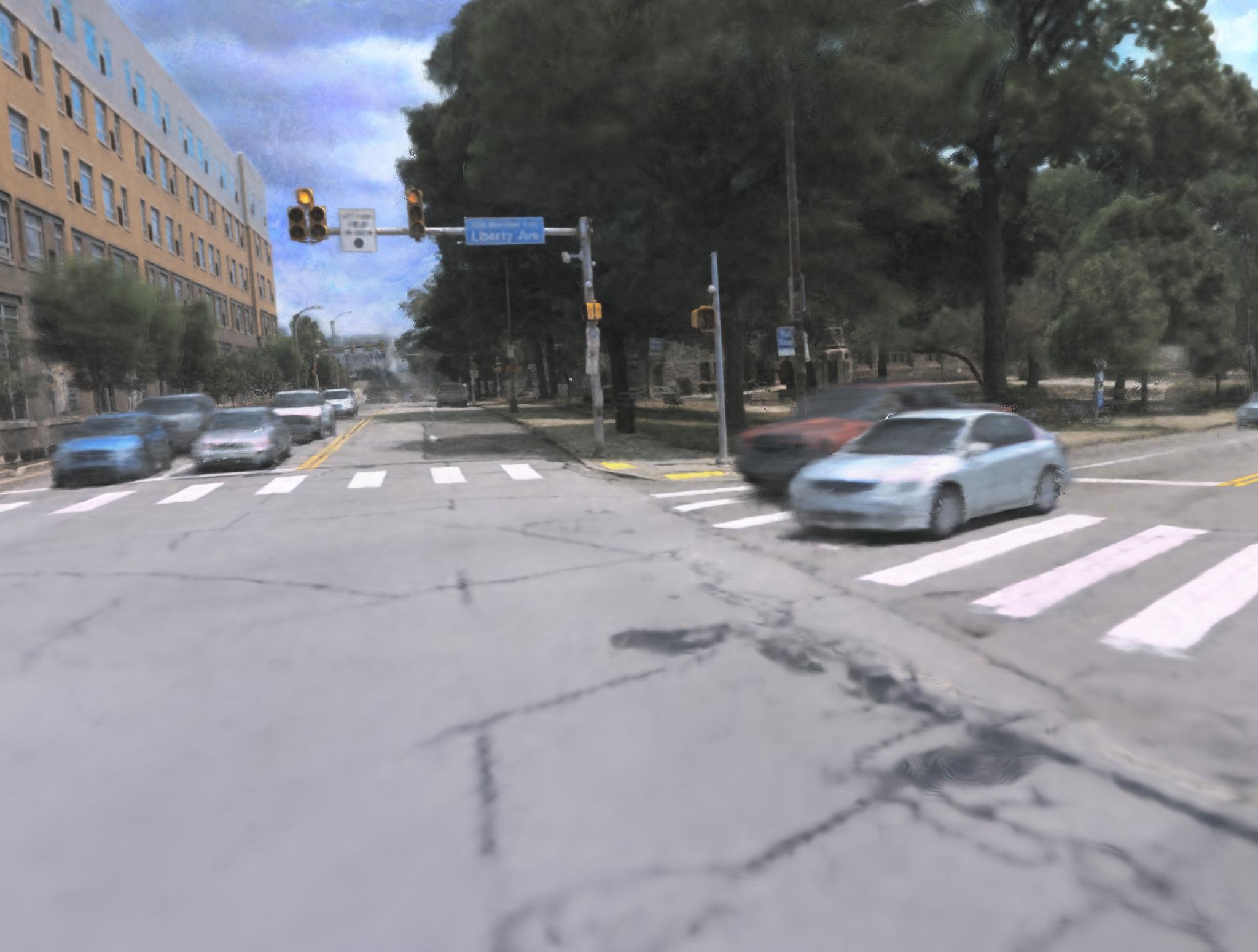} &  \includegraphics[width=0.2\linewidth]{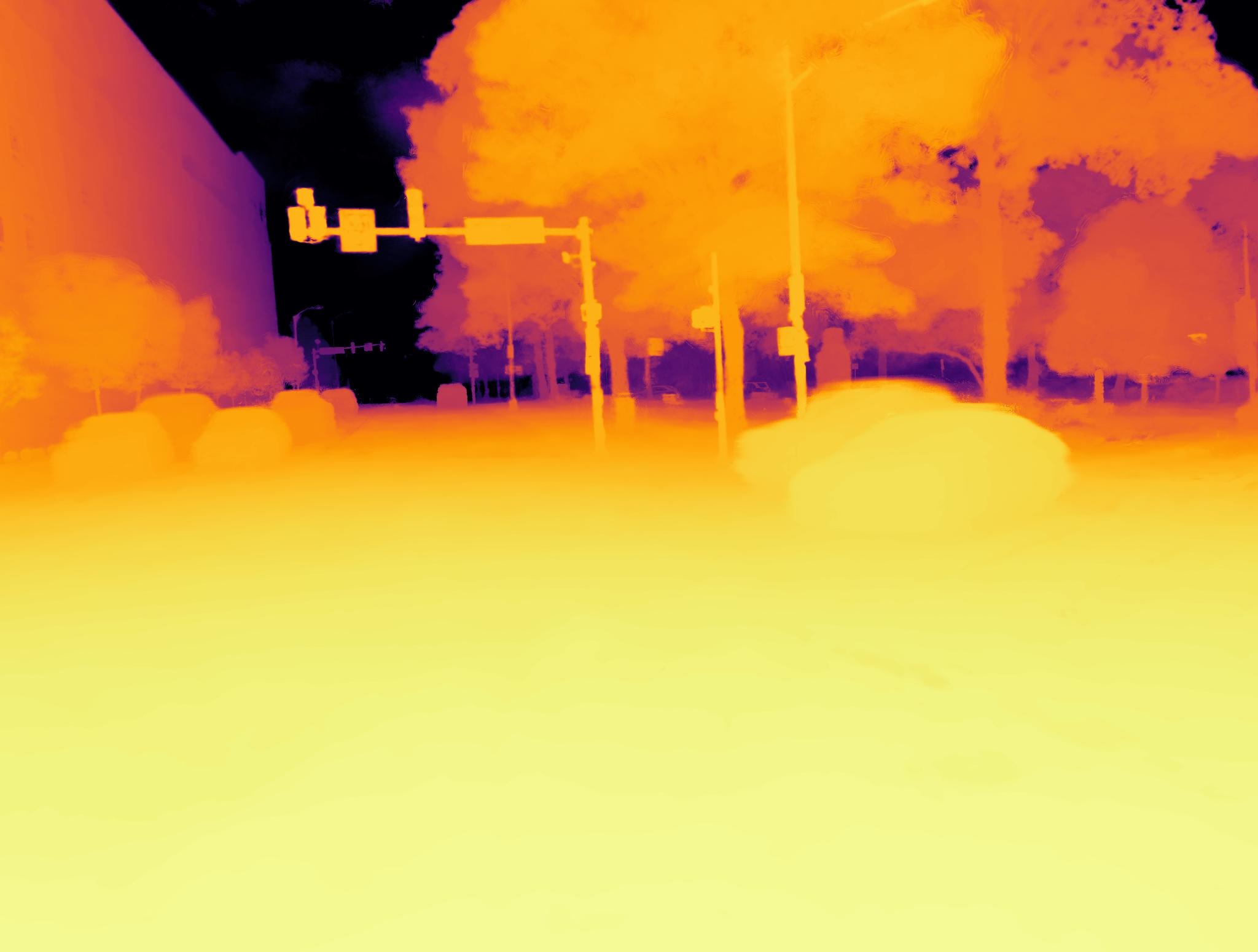} \\

  \rotatebox{90}{\hspace{7mm}SUDS~\cite{turki2023suds}} & \includegraphics[width=0.2\linewidth]{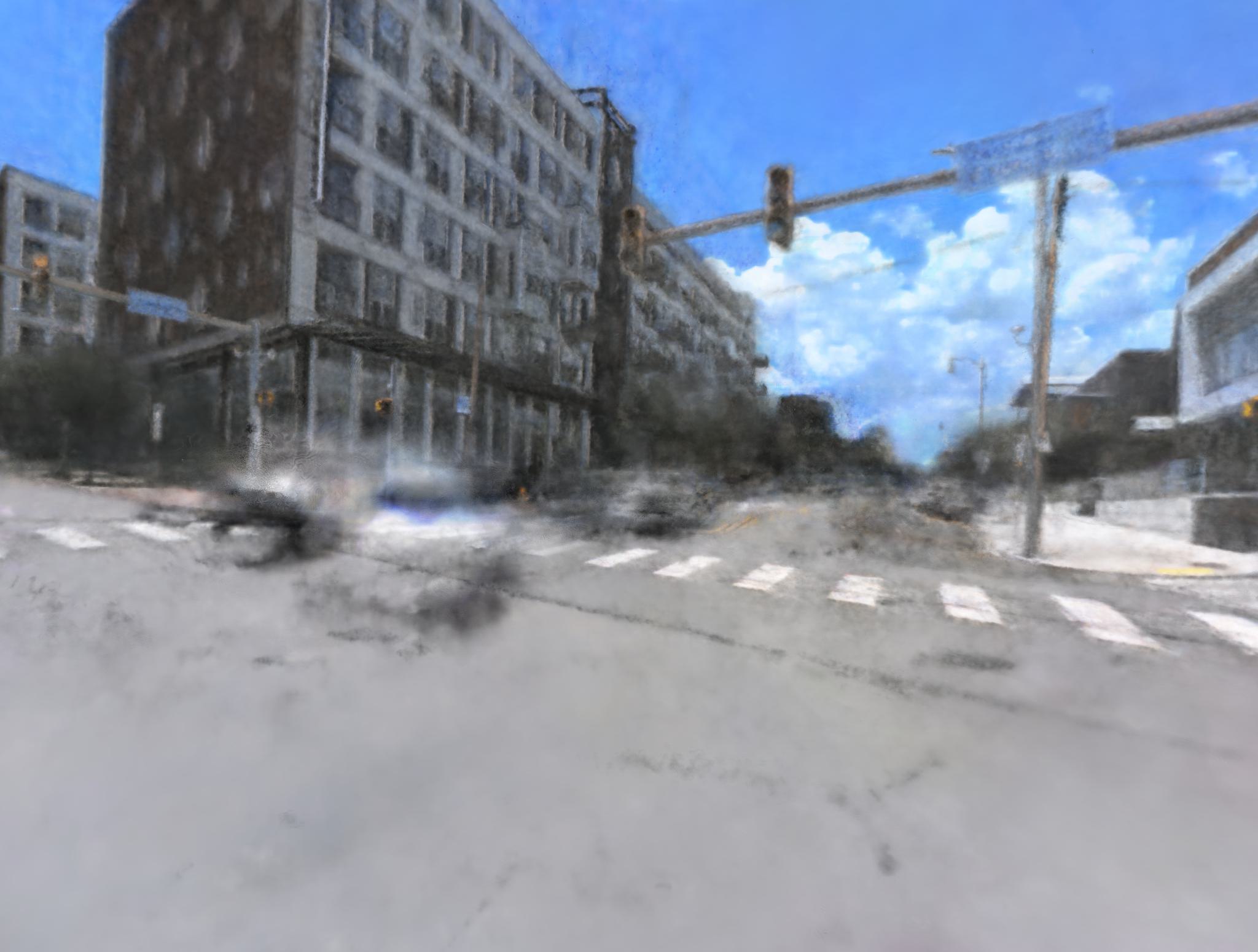} &\includegraphics[width=0.2\linewidth]{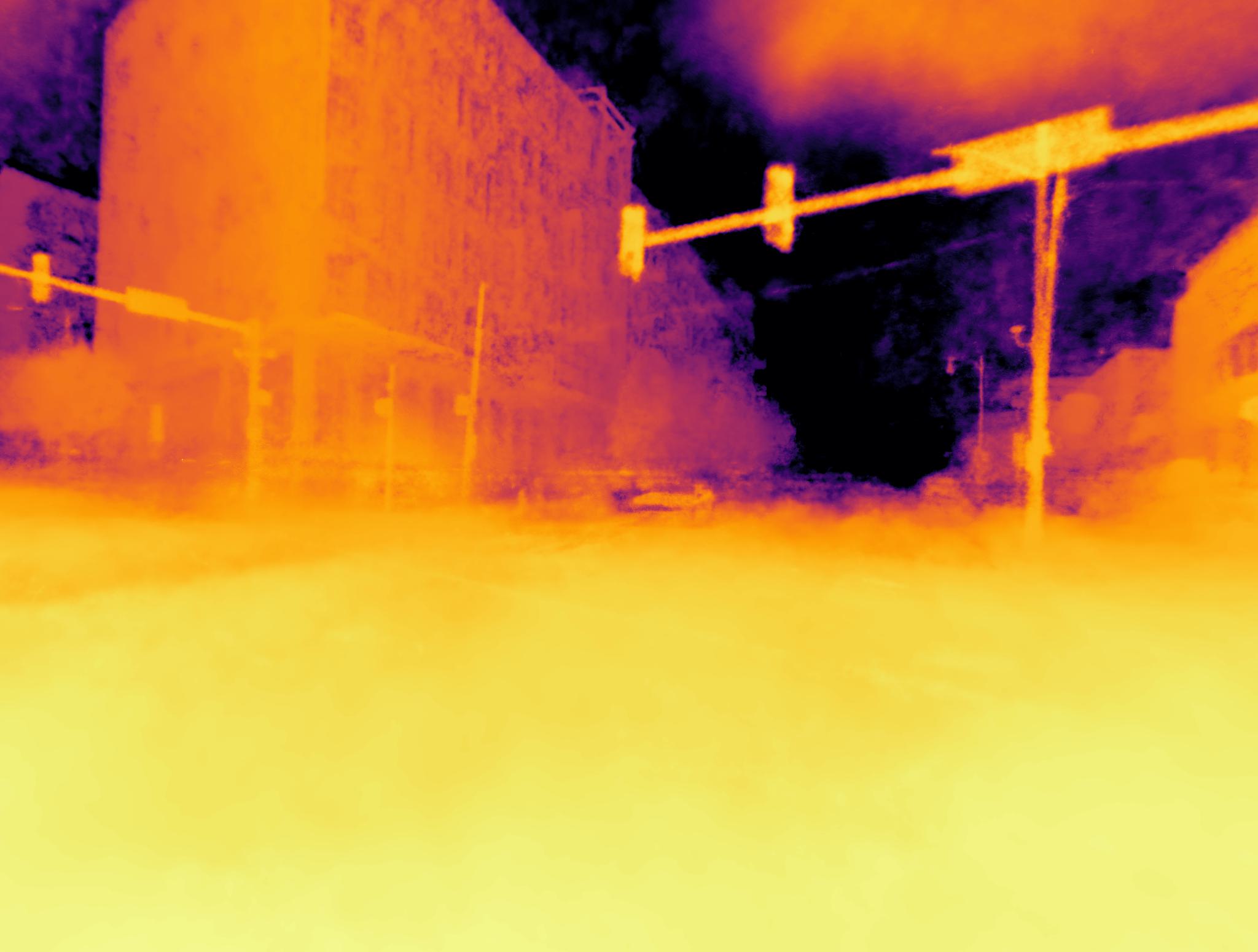} &  \includegraphics[width=0.2\linewidth]{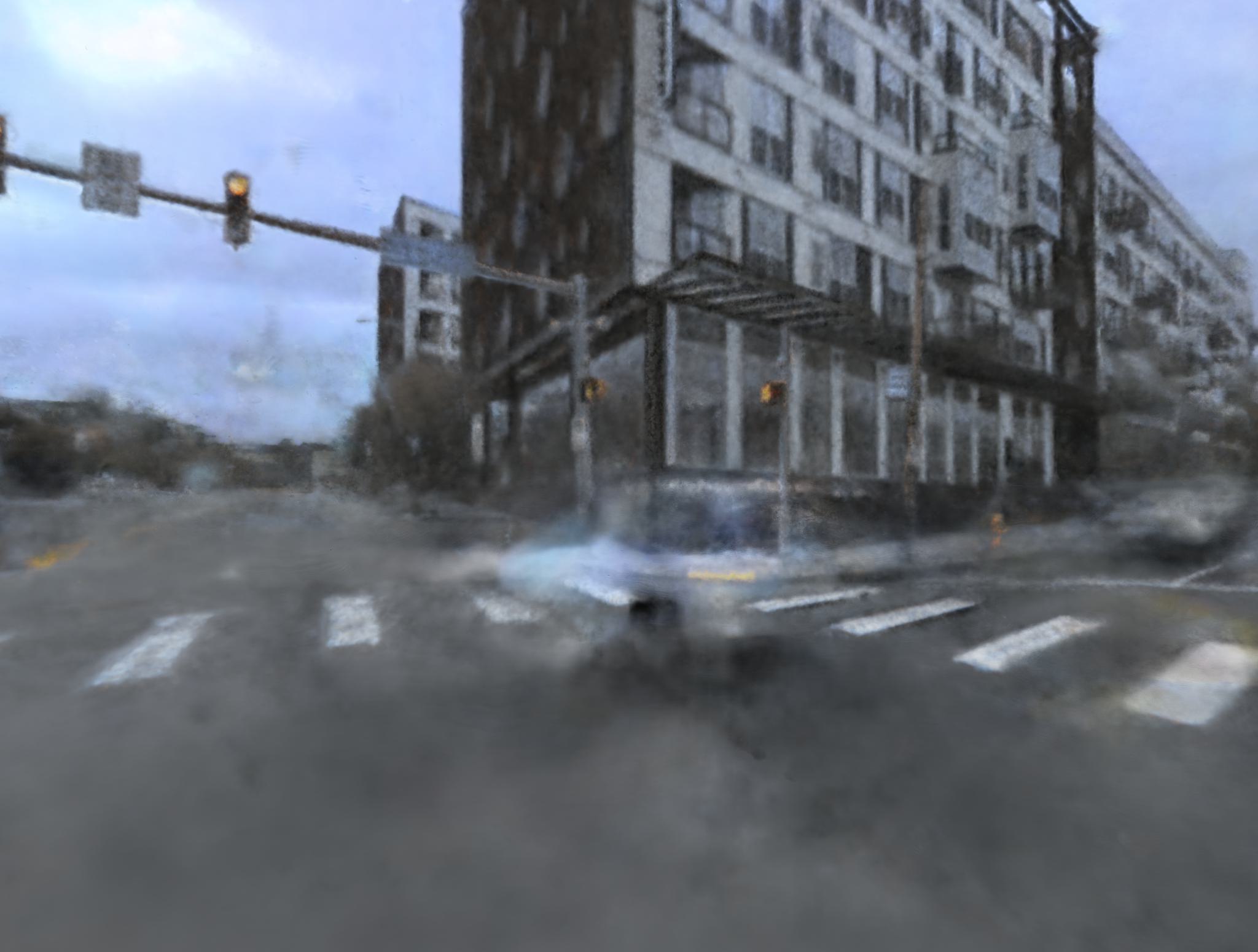} &  \includegraphics[width=0.2\linewidth]{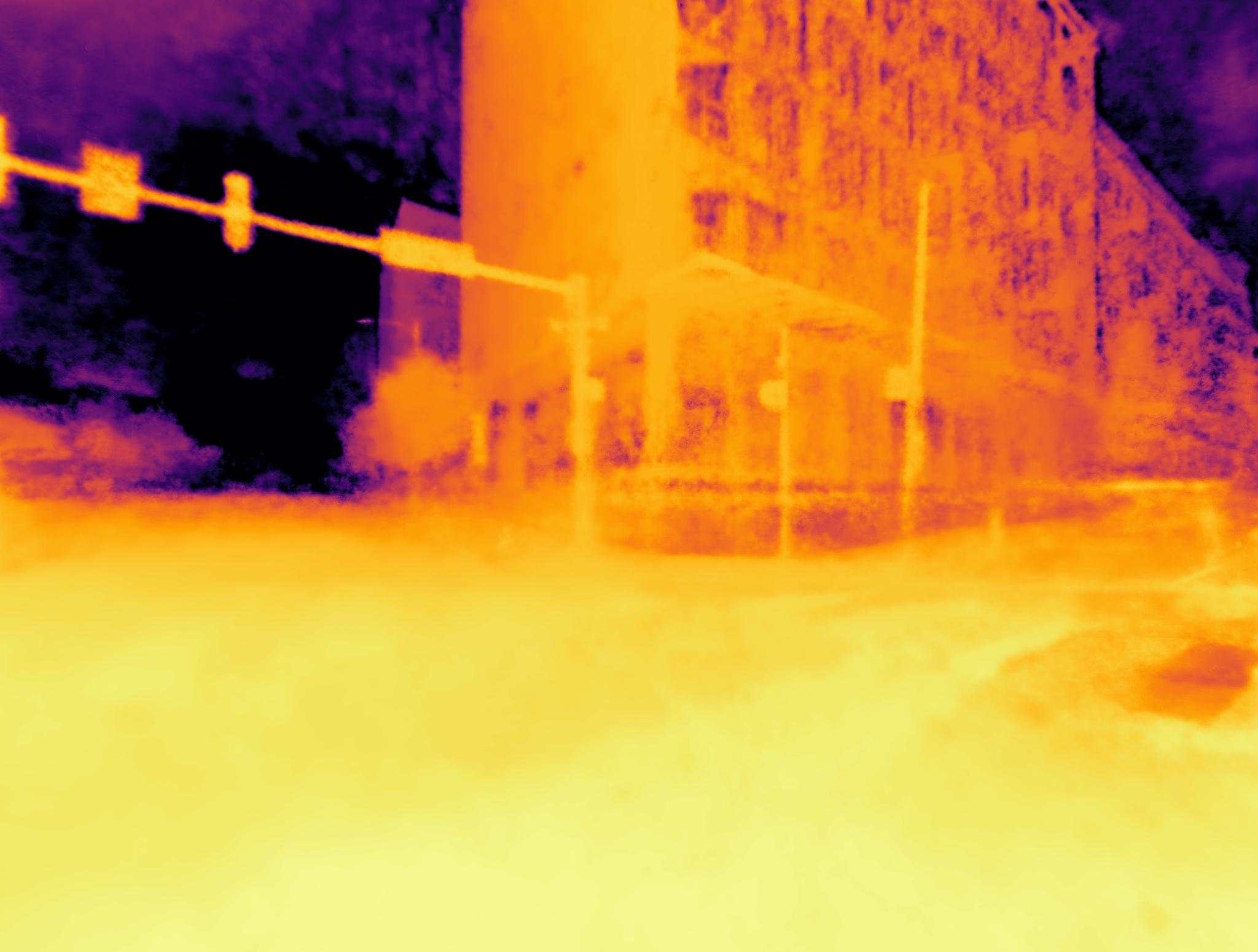} &  \includegraphics[width=0.2\linewidth]{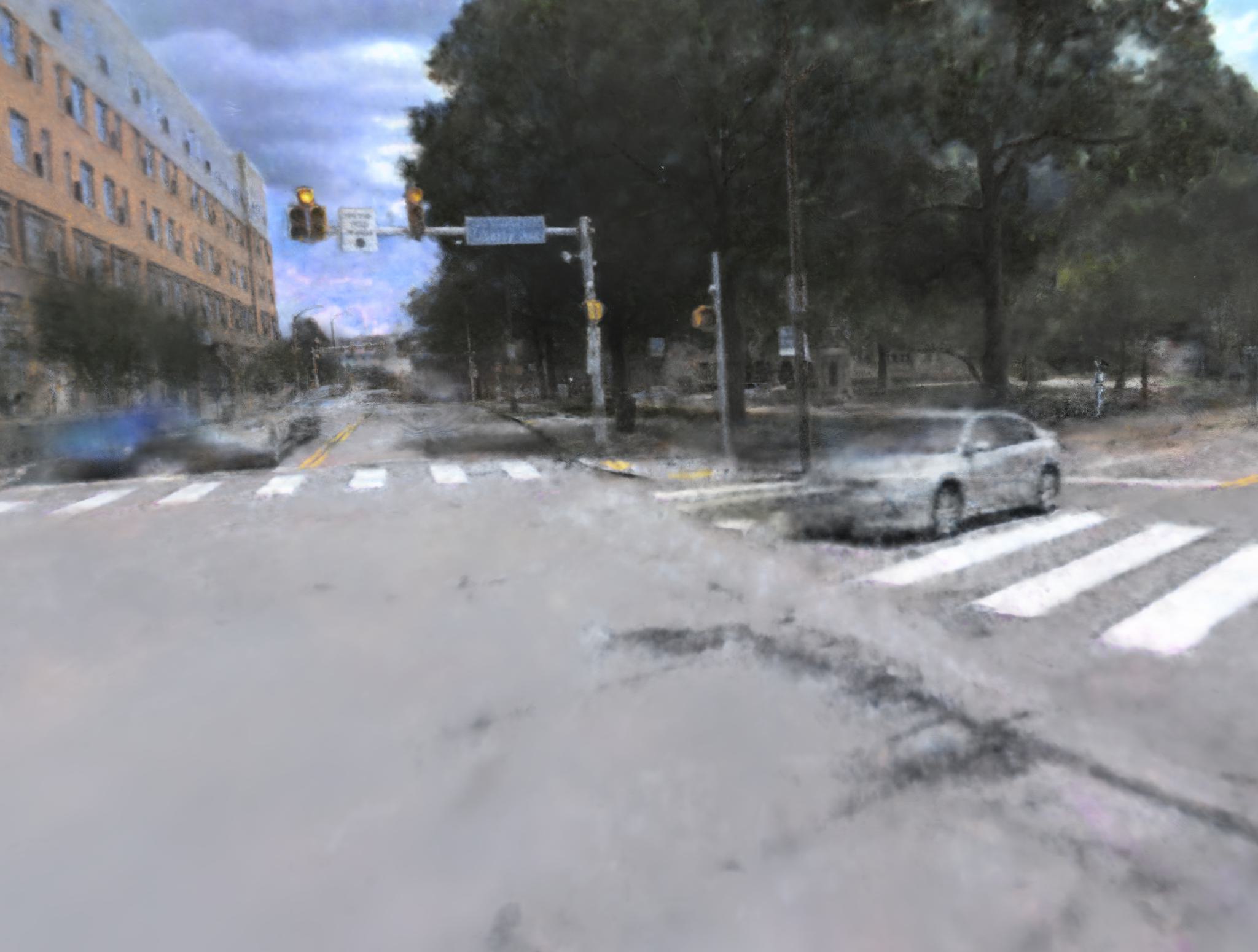} &  \includegraphics[width=0.2\linewidth]{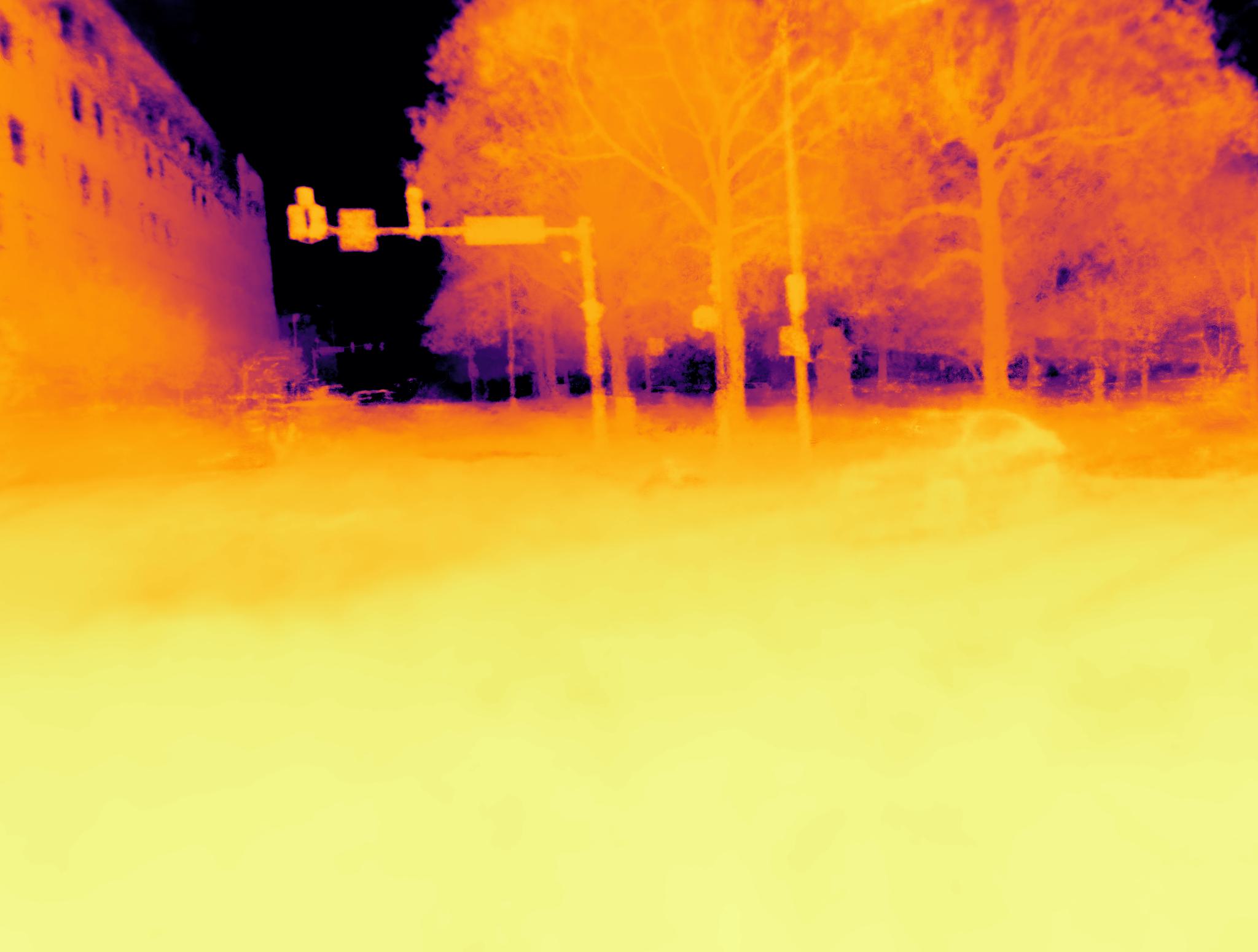} \\

  \rotatebox{90}{\hspace{0.5mm}Nerfacto+Emb.+Time} & \includegraphics[width=0.2\linewidth]{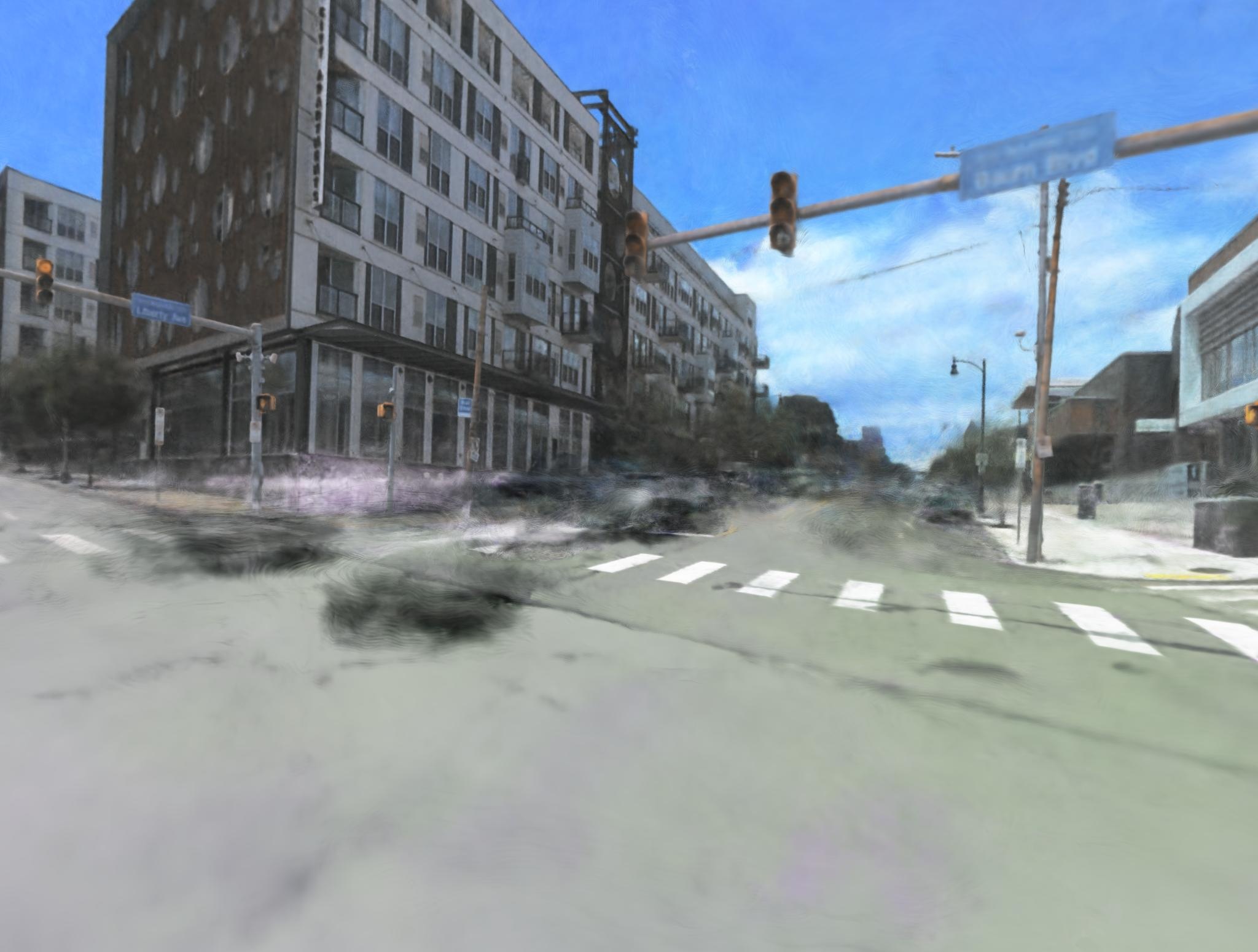} &\includegraphics[width=0.2\linewidth]{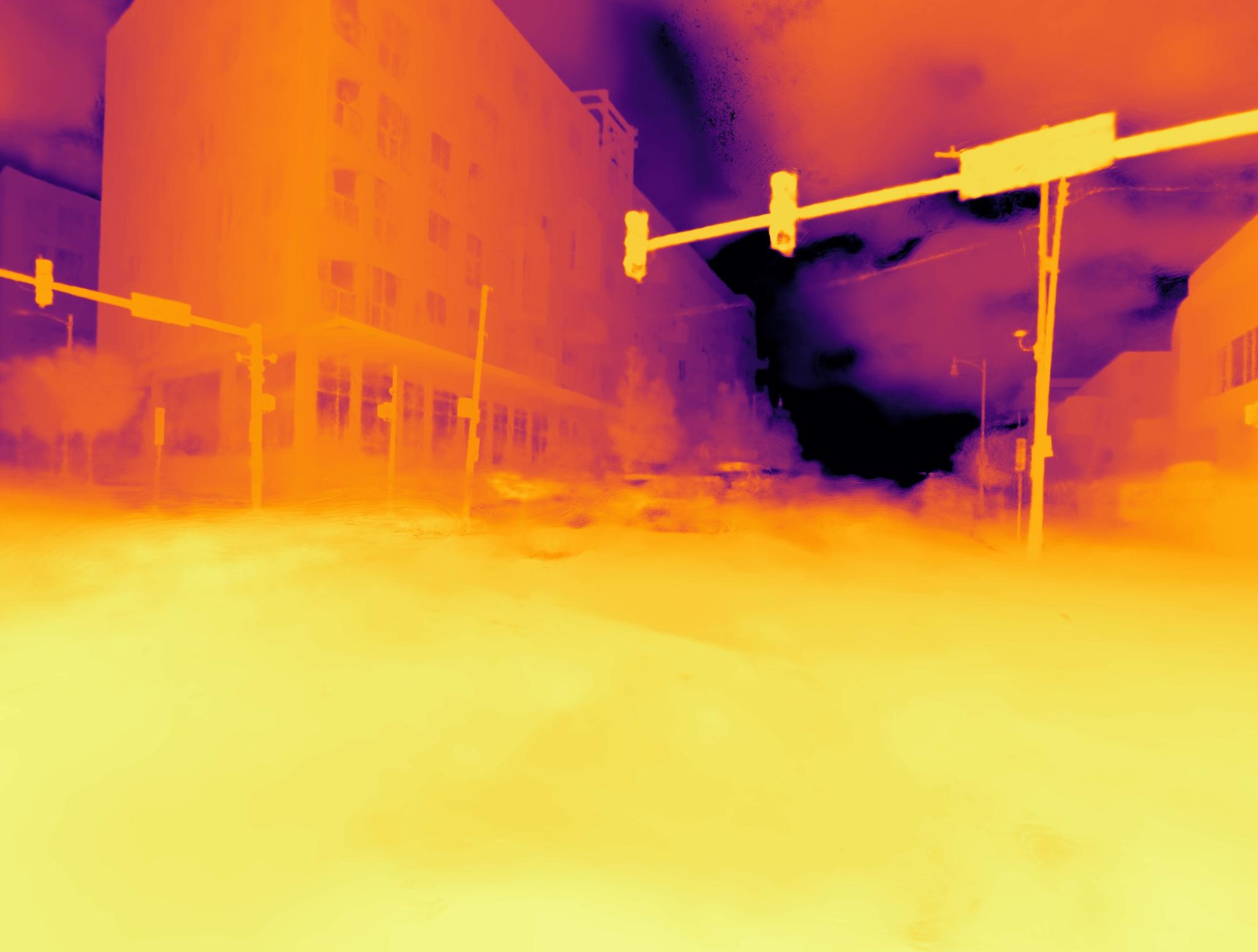} &  \includegraphics[width=0.2\linewidth]{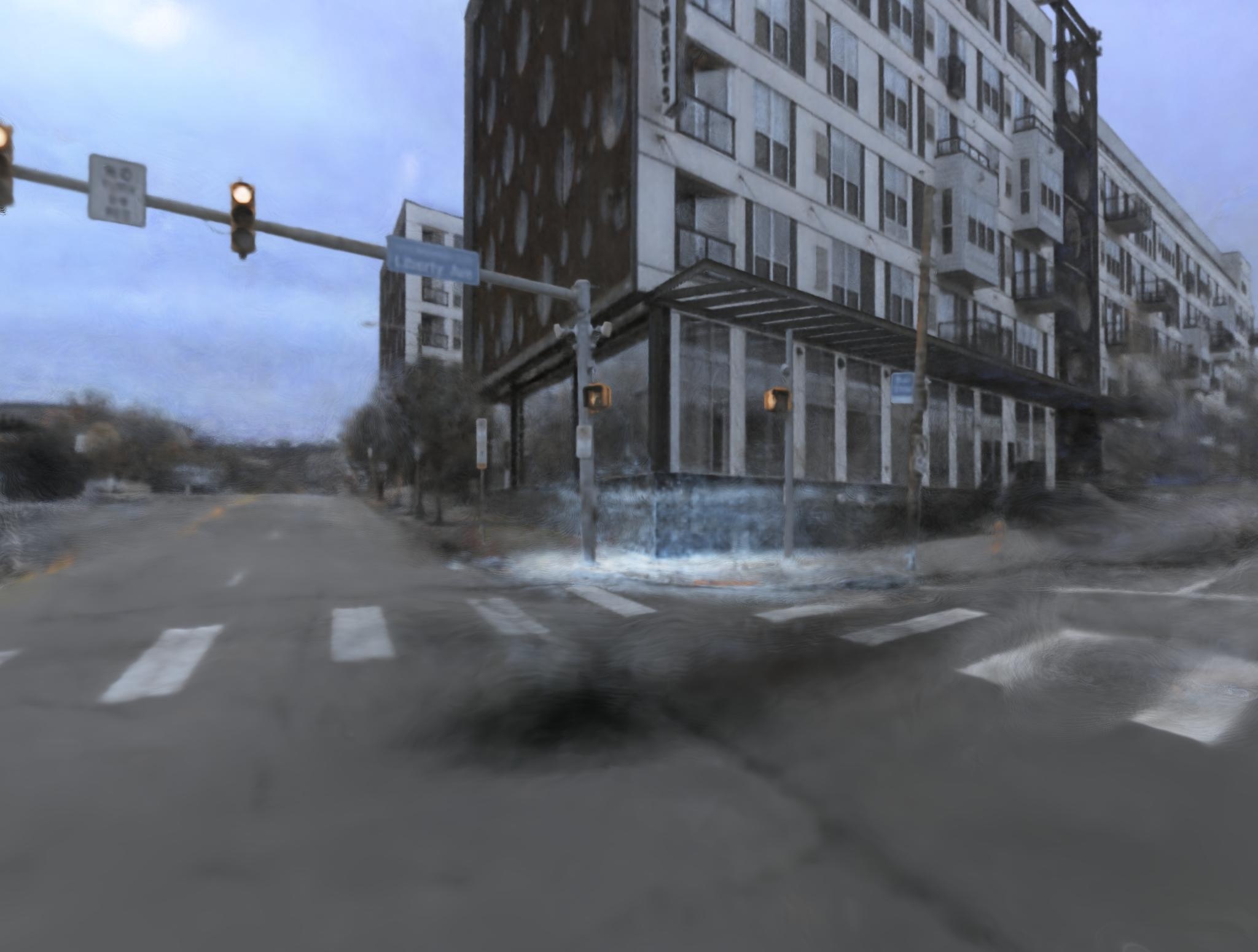} &  \includegraphics[width=0.2\linewidth]{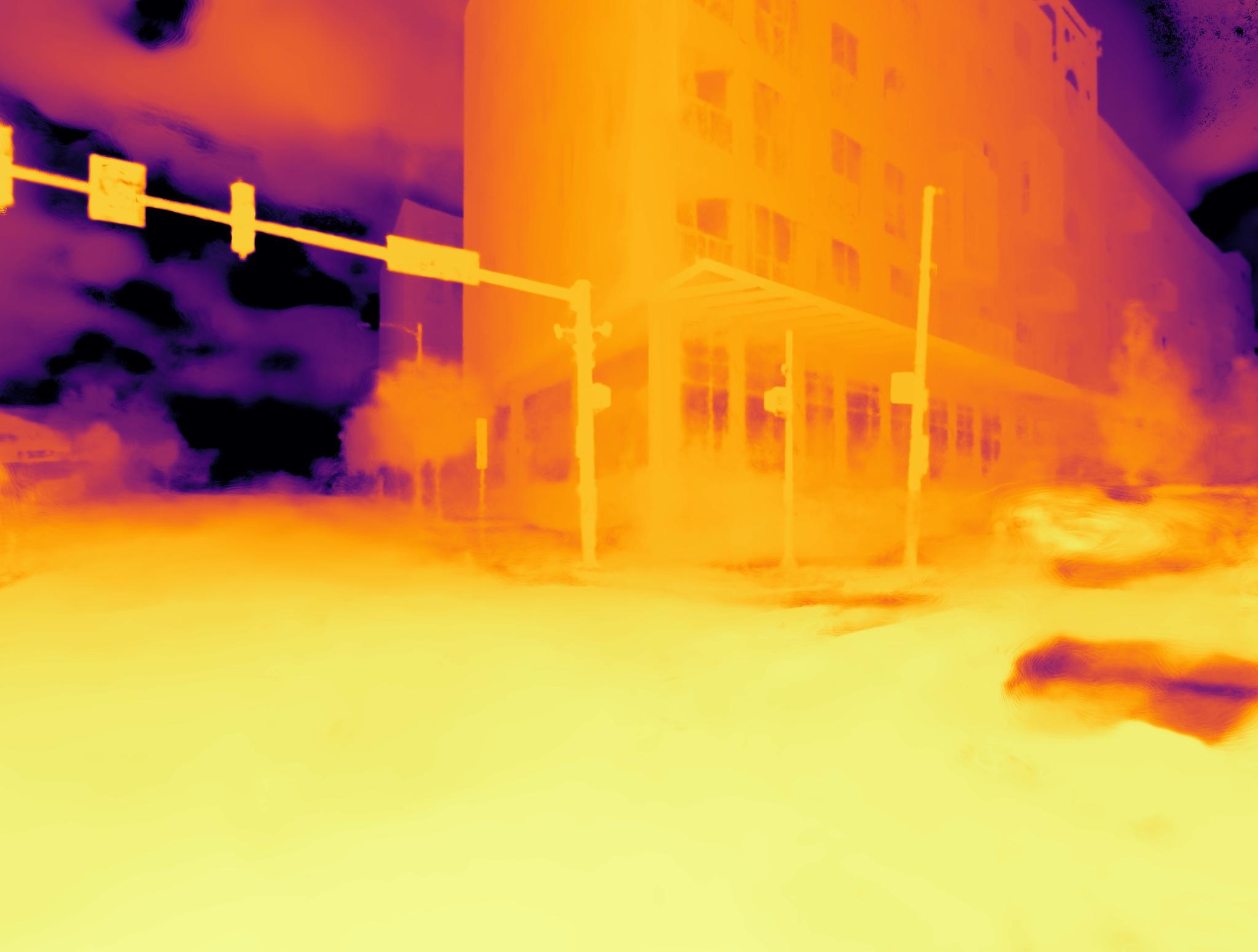} &  \includegraphics[width=0.2\linewidth]{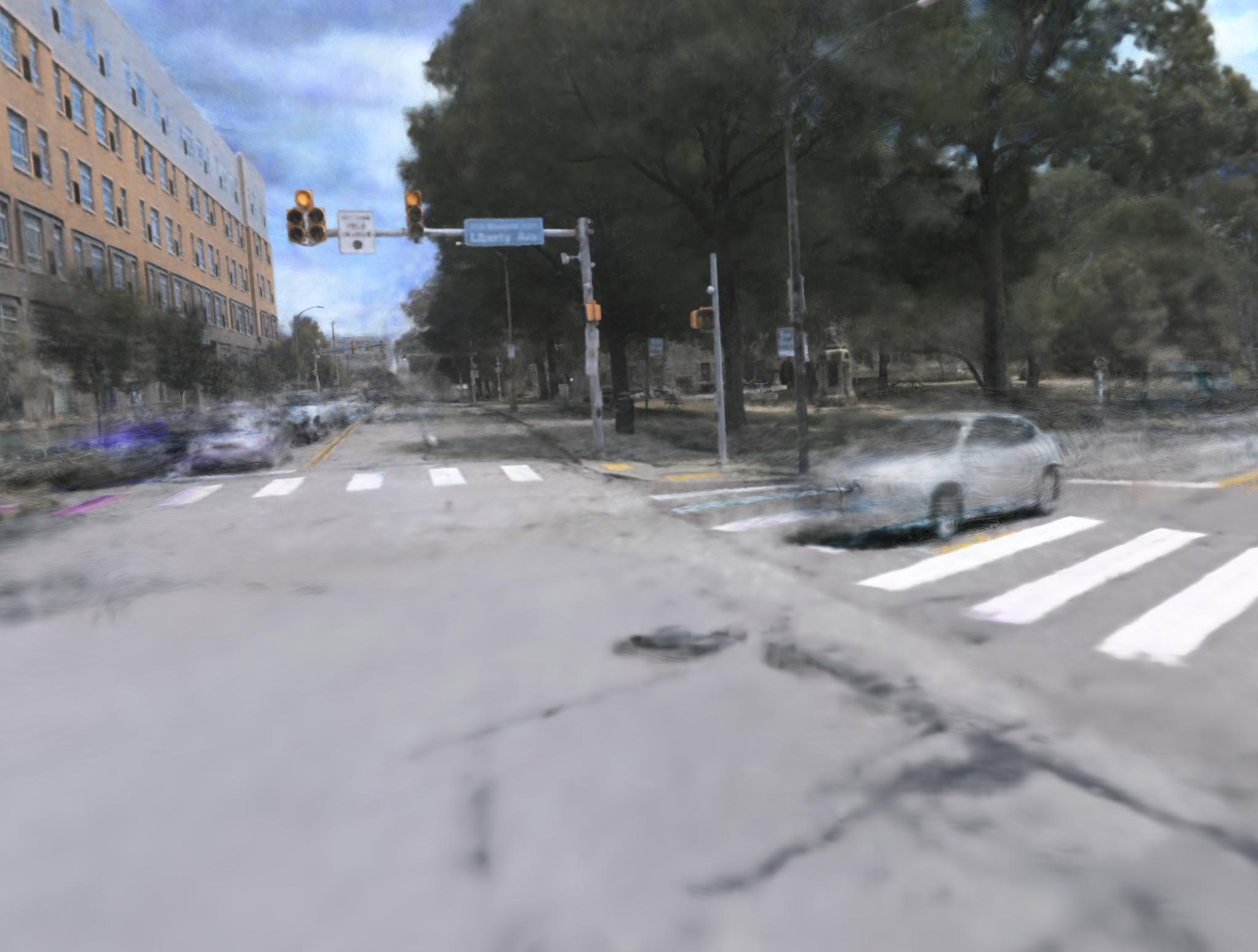} &  \includegraphics[width=0.2\linewidth]{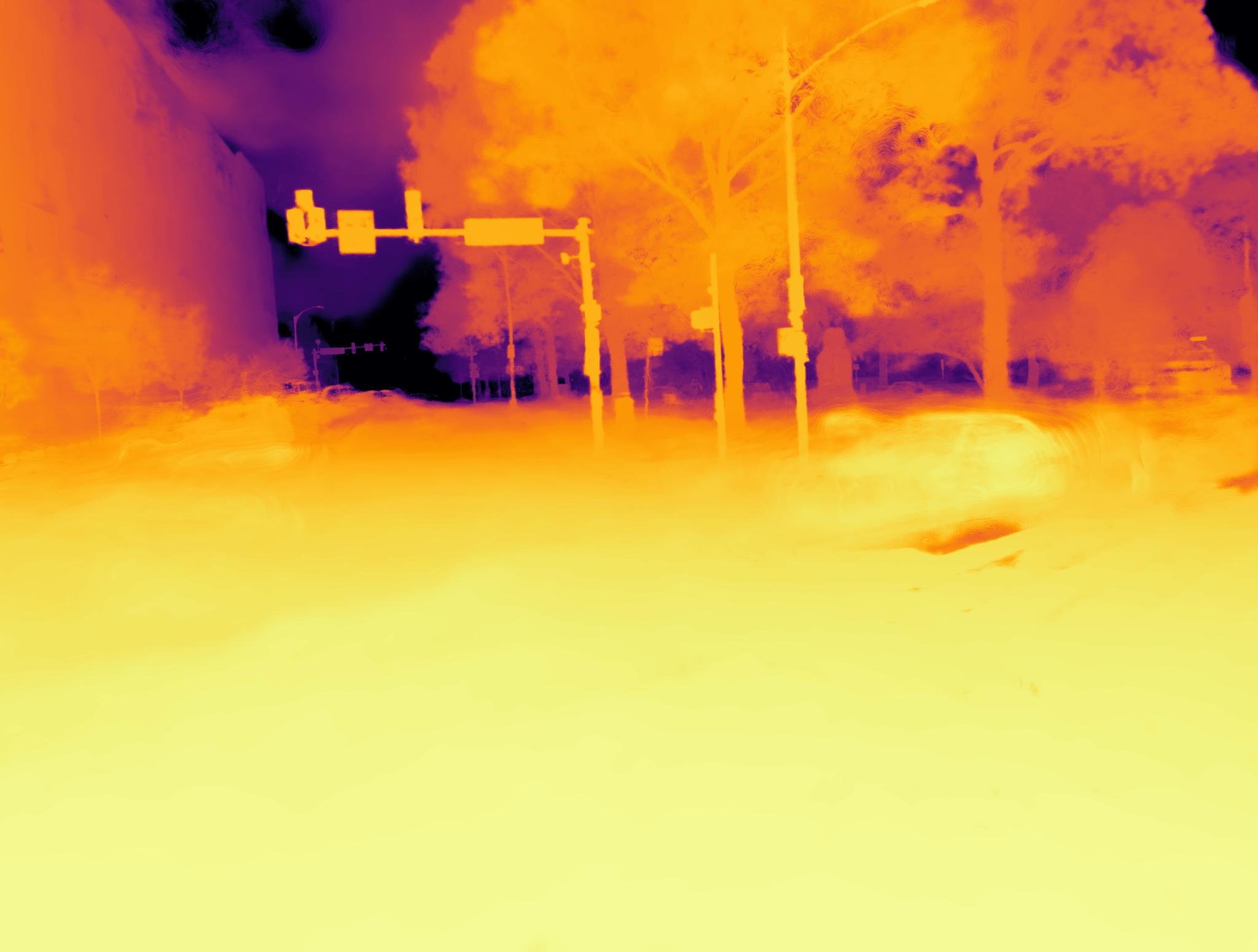} \\

  \rotatebox{90}{\hspace{3mm}Nerfacto + Emb.}  & \includegraphics[width=0.2\linewidth]{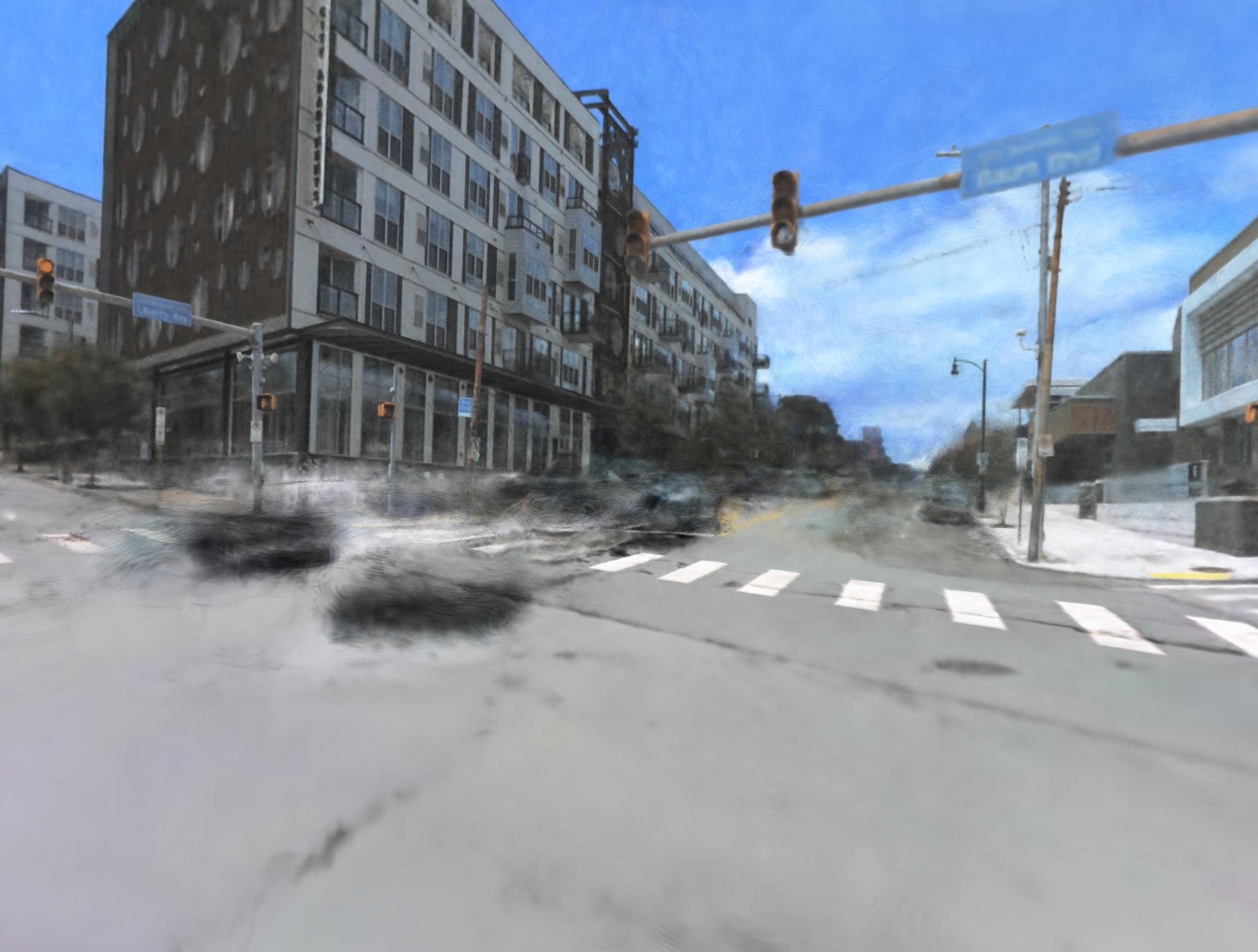} &\includegraphics[width=0.2\linewidth]{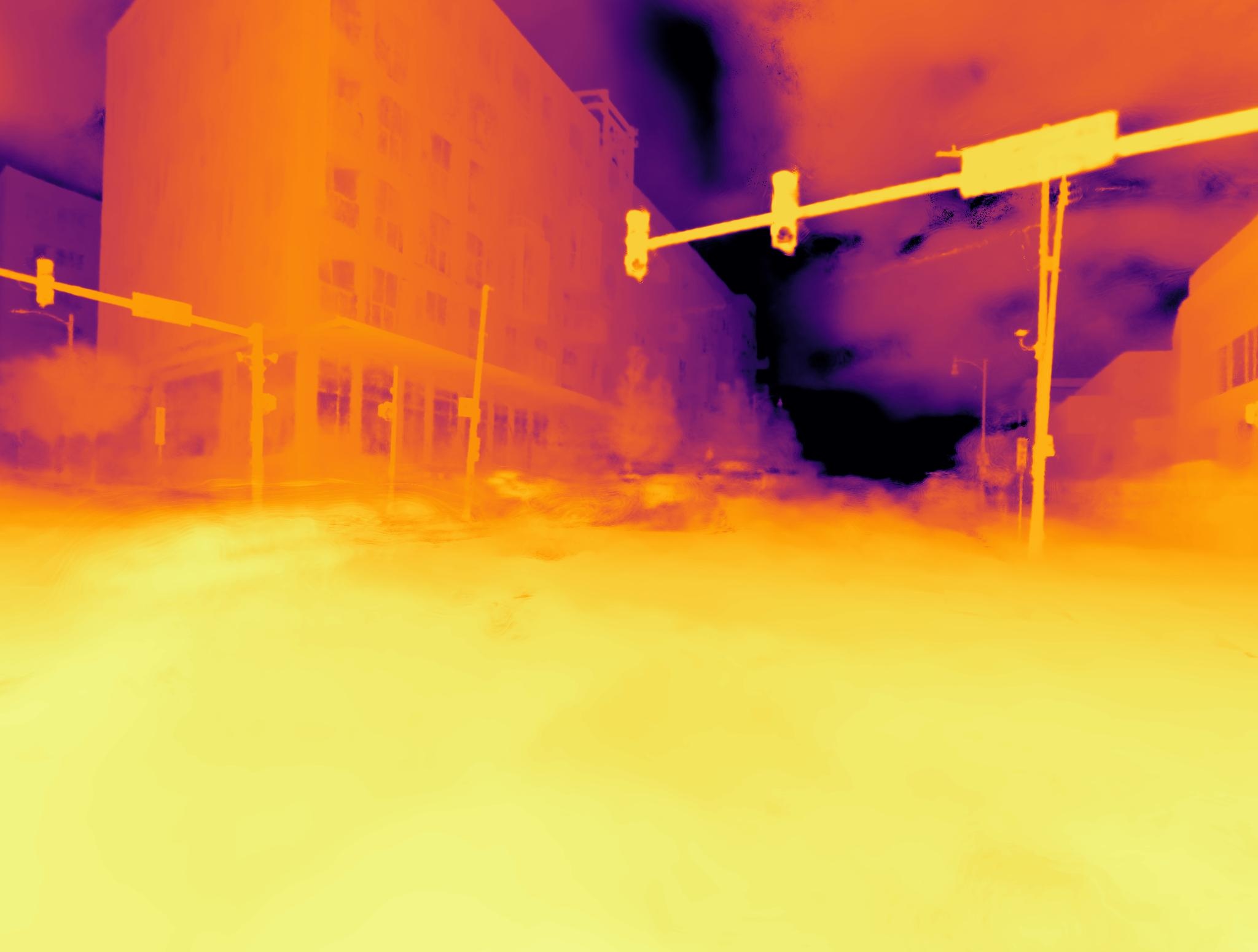} &  \includegraphics[width=0.2\linewidth]{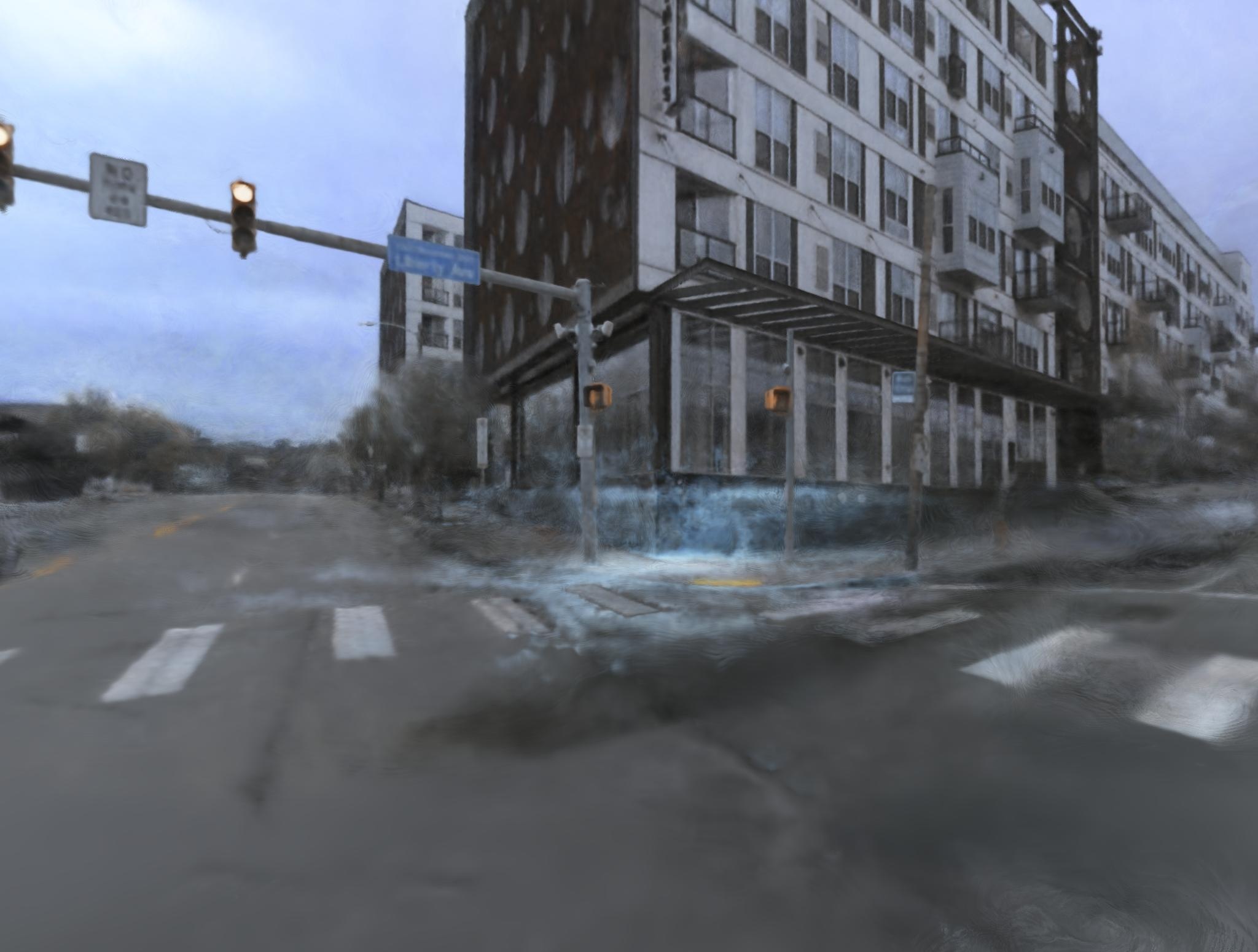} &  \includegraphics[width=0.2\linewidth]{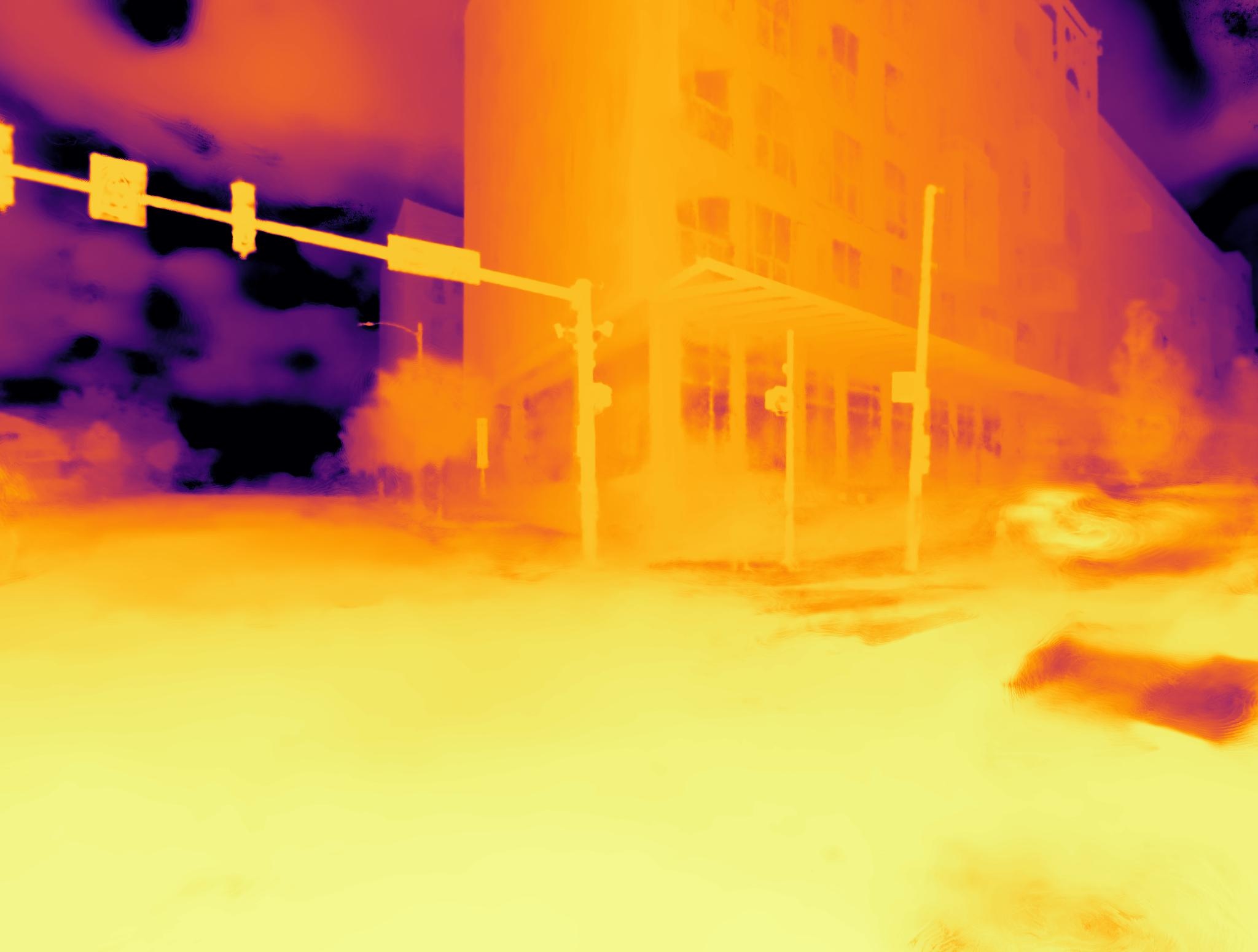} &  \includegraphics[width=0.2\linewidth]{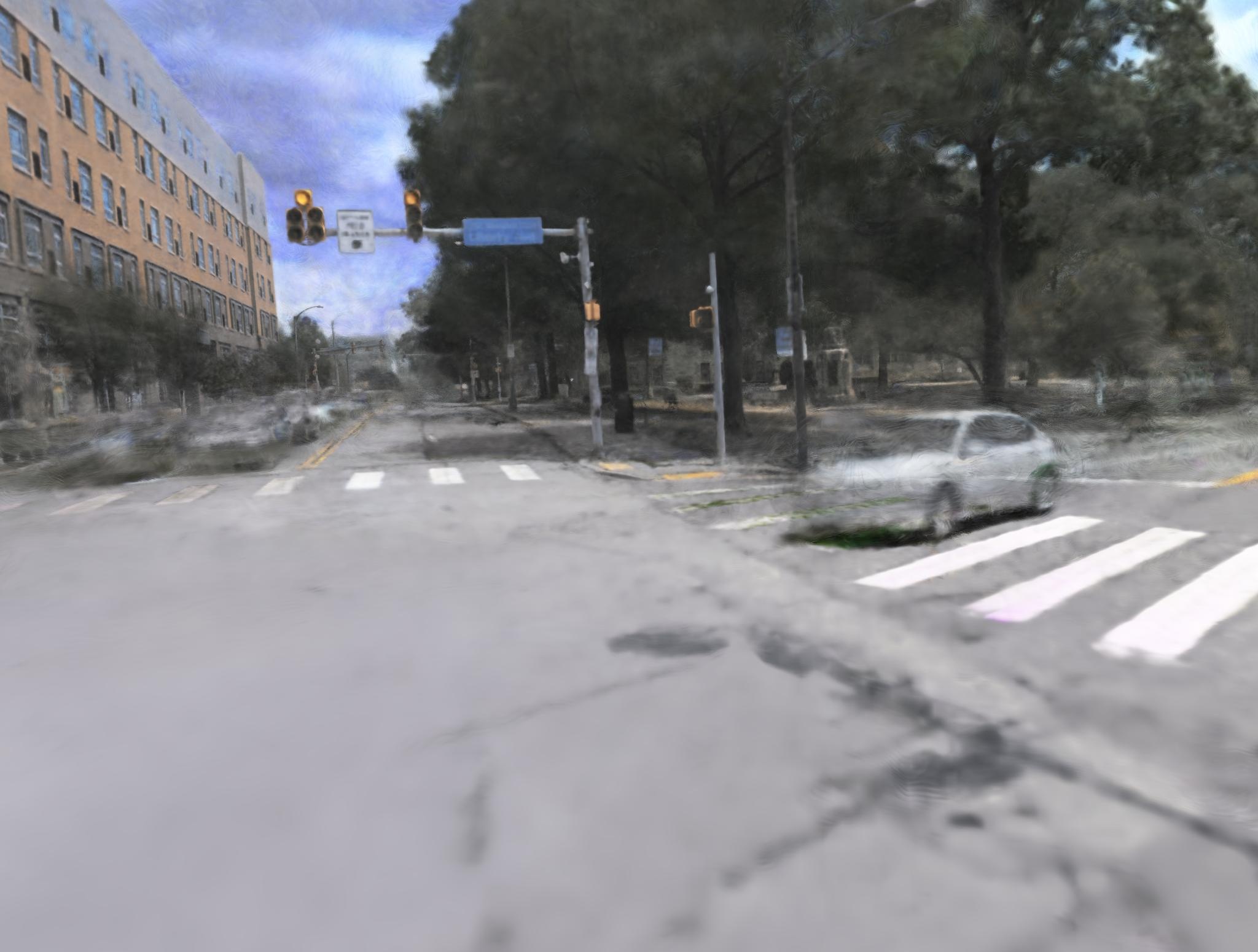} &  \includegraphics[width=0.2\linewidth]{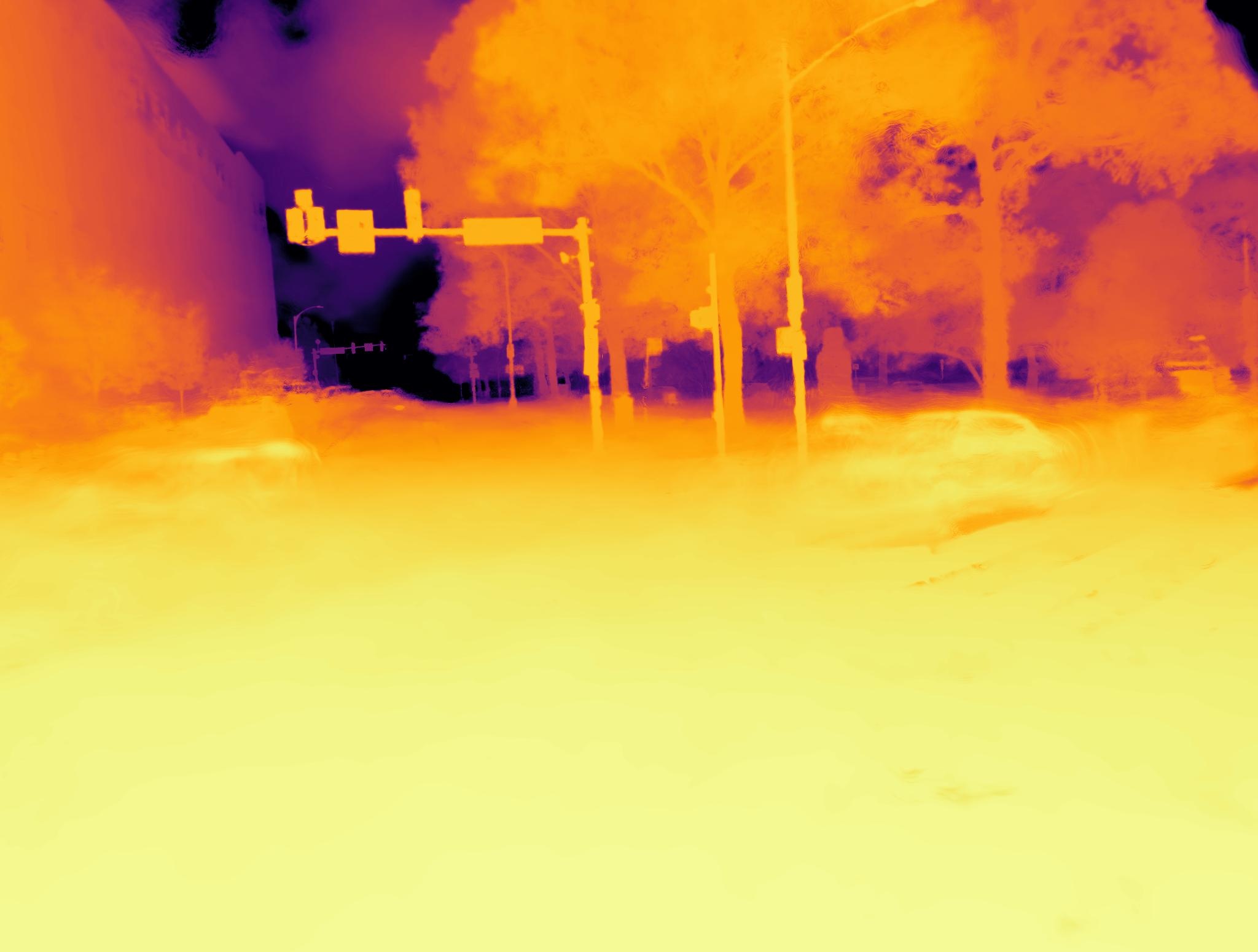} \\

\end{tabular}

}\vspace{-2mm}
\caption{\textbf{Qualitative results on Argoverse 2.} While prior art struggles with dynamic actors in the scene, our representation renders both realistic novel views and plausible depth maps. Further, our method models complex transient geometry subject to, \eg, seasonal changes such as tree leaves. Note that prior art produces depth artifacts and exhibits degraded view quality in those regions (columns 5 and 6).}
\label{fig:av2_qualitative}
\vspace{-2mm}
\end{figure*}

In addition, we show a qualitative comparison in Fig.~\ref{fig:av2_qualitative}. 
We observe major differences in both the rendered images and depth maps. In particular, all other methods struggle to recover the dynamic objects in the scene, while our method produces realistic renderings and accurate depth maps for both static and dynamic areas. We also observe that, thanks to our transient geometry embedding $\transient_\seq\mathcal{F}(t)$, we are able to accurately recover the geometry of the trees and their leaves which notably are not present in every sequence of the area that is reconstructed due to seasonal changes. At the same time, other methods struggle to recover those details. Specifically, other methods exhibit artifacts in the depth maps and degraded rendering quality of the trees left and right of the street in columns five and six of Fig.~\ref{fig:av2_qualitative}, while our method produces accurate color and depth renderings.

\begin{table}
\centering
\scriptsize
\begin{tabular}{cc|ccc}
\toprule
$\appearance_\seq\mathcal{F}(t)$ & $\transient_\seq\mathcal{F}(t)$ & PSNR $\uparrow$ & SSIM $\uparrow$ & LPIPS $\downarrow$ \\ \midrule
- & - & \rd 19.70 & \rd 0.653 & \rd 0.588 \\
\checkmark & - & \nd 22.22 & \nd 0.670 & \nd 0.546 \\
\checkmark & \checkmark & \fs 22.49 & \fs 0.671 & \fs 0.541 \\ \bottomrule
\end{tabular}
\vspace{-2mm}
\caption{\textbf{Ablation study on graph structure.} We show that using the multi-level graph structure of sequences and objects is crucial, \ie omitting sequence vectors $\omega_\seq^t$ results in degraded quality since there is no conditioning on scene-specific appearance. Combining appearance and transient geometry embeddings performs best.}
\label{tab:embedding_ablation}
\vspace{-4mm}
\end{table}

\begin{table}
\centering
\footnotesize
\resizebox{1.0\linewidth}{!}{
\begin{tabular}{@{}l|c|ccc|c@{}}
\toprule
Sampling & Samples per ray & PSNR $\uparrow$ & SSIM $\uparrow$ & LPIPS $\downarrow$ & Rays / sec. \\ \midrule
Uniform~\cite{kundu2022panoptic} & 192 & 25.72 & 0.730 & 0.456 &  28K \\
Uniform~\cite{kundu2022panoptic} & 1024 & \rd 25.84 & \rd 0.734 & \rd 0.449 &  4K \\
Separate~\cite{ost2021neural} & 1024+64+(32+64) & \nd 26.65 & \fs 0.762 & \fs 0.351 & 2.5K \\
\textbf{Ours} & 1024+64 & \fs 27.07 & \nd 0.759 & \nd 0.362 & 30K \\ \bottomrule
\end{tabular}}
\vspace{-2mm}
\caption{\textbf{Ray Sampling schemes.} Our composite ray sampling is about $12\times$ more efficient to train than separate ray sampling~\cite{ost2021neural} with comparable performance. Uniform sampling~\cite{kundu2022panoptic} exhibits lower performance and is slow when sampled more densely. }
\label{tab:ray_sampling}
\vspace{-4mm}
\end{table}

\begin{table}
\centering
\scriptsize
\begin{tabular}{@{}l|ccc}
\toprule
Pose optimization & PSNR $\uparrow$ & SSIM $\uparrow$ & LPIPS $\downarrow$ \\ \midrule
- & \fs 22.49 & \nd 0.671 & \rd 0.541 \\
Naive & \rd 21.28 & \rd 0.663 & \fs 0.519 \\
Hierarchical (\textbf{Ours}) & \nd 22.29 & \fs 0.678 & \nd 0.523\\ \bottomrule
\end{tabular}
\vspace{-2mm}
\caption{\textbf{Pose optimization.} Compared to naively optimizing camera and object poses, optimizing vehicle and object poses hierarchically on the edges of our graph benefits view synthesis quality more consistently, \ie both in SSIM and LPIPS. Note that PSNR is very sensitive to pose drift, and thus does not improve.}
\label{tab:pose_optim}
\vspace{-5mm}
\end{table}

Next, we evaluate our method on established benchmarks, namely KITTI and VKITTI2, in Tab.~\ref{tab:kitti_nvs} following the experimental protocol in~\cite{turki2023suds}.
In particular, we compare novel view synthesis quality at different fractions of training views, \ie testing different levels of view supervision. We run our method and our baselines and observe substantially better view synthesis quality for our method, both compared to our baselines and prior works. All methods exhibit a similar performance drop when the fraction of training views decreases. We illustrate the rendering quality of our method with only 15-20 training views in Fig.~\ref{fig:kitti_qualitative}.
Finally, we also report the results on the task of image reconstruction, \ie reconstruction of images seen during training, following~\cite{ost2021neural, kundu2022panoptic, turki2023suds} in Tab.~\ref{tab:kitti_imrec} on the KITTI dataset. We significantly outperform previous works in both metrics.

\subsection{Ablation studies}
We verify our design through ablation studies. We perform these on the residential area of our benchmark unless otherwise noted. First, we ablate on the multi-level structure of our scene graph. In particular, in Tab.~\ref{tab:embedding_ablation}, we ablate the latent codes of the sequence nodes, reducing to a representation similar to~\cite{ost2021neural} and observe that the quality of the synthesized views drops significantly. In contrast, our scene graph representation achieves better view synthesis quality, \ie achieves almost three points higher PSNR. Further, adding our transient geometry embeddings in addition to sequence-level appearance embeddings improves the view synthesis quality. See supplemental material for a qualitative comparison.
This adds to the fact that through our multi-level graph structure, we show the ability to modulate car appearance through sequence appearance in Fig.~\ref{fig:car_appearance}.

In Tab.~\ref{tab:ray_sampling}, we compare our composite ray sampling scheme to uniformly sampled rays as in~\cite{kundu2022panoptic} and rays sampled separately per node as in~\cite{ost2021neural}. We compare the schemes under the assumption that all other model parameters are equal, implementing them in our method. Ours is about $12\times$ more efficient to train than densely sampling a ray in uniform intervals or separately sampling the ray per node while producing similar results to the latter and being significantly better than the former. Sparsely sampling a ray at uniform intervals is similarly fast as our method, but yields degraded view synthesis quality.
Note that we run this ablation study on only a single sequence of the residential area in our benchmark since densely sampling the rays and separately sampling the rays for all nodes are prohibitively expensive to train in large-scale urban areas.

Finally, in Tab.~\ref{tab:pose_optim}, we compare our hierarchical pose optimization to naive camera pose optimization employed in previous works~\cite{tancik2023nerfstudio, lin2021barf}. While naive camera pose optimization degrades the results significantly in pixel-wise metrics, our hierarchical pose optimization improves the SSIM and maintain a comparable PSNR. Meanwhile, our hierarchical pose optimization exhibits similar gains to naive pose optimization in terms of the perceptual LPIPS metric. This shows that our pose optimization mitigates pose drift while also enabling a more accurate reconstruction. Pose drift usually causes a misalignment between the evaluation viewpoint and scene geometry, which degrades pixel-wise metrics in particular.

\section{Conclusion}
We introduce a novel multi-level scene graph representation for radiance field reconstruction in dynamic urban environments that scales to large geographic areas with more than ten thousand images from dozens of sequences and hundreds of dynamic objects.
We train our representation with an efficient composite ray sampling and rendering scheme and introduce latent variables that enable modeling complex phenomena like varying environmental conditions and transient geometry present across different vehicle captures. We leverage our representation to refine camera and object poses hierarchically using multi-camera constraints.
Finally, we propose a new view synthesis benchmark for dynamic urban driving scenarios.
Our approach yields substantially improved results compared to prior art while allowing for flexible de- and recomposition of the scene.

{
    \small
    \bibliographystyle{ieeenat_fullname}
    \bibliography{main}
}

\clearpage
\appendix

{%
\centerline{\large\bf Appendix}%
\vspace*{12pt}%
\it%
}

\noindent This supplementary material provides details on our method, our experimental setup, and more quantitative and qualitative results and comparisons.
In Sec.~\ref{sec:supp_data}, we provide further details on our benchmark data. 
In Sec.~\ref{sec:supp_method}, we provide further details on our method. 
In Sec.~\ref{sec:supp_exp}, we provide details on our experimental setup, conduct additional experiments, and show more qualitative comparisons.

\section{Data Details}
\label{sec:supp_data}
We describe further details on our benchmark data taken from~\cite{wilson2023argoverse}.
The LiDAR is sampled at 10Hz, and the 3D bounding box annotations are annotated with the LiDAR, \ie they are also provided at 10Hz. The cameras are synchronized with the LiDAR which yields seven images at 10Hz for each sequence. Each camera has a resolution of $1550 \times 2048$ pixels, where all cameras besides the front camera are oriented in landscape mode. Each sequence in Argoverse 2~\cite{wilson2023argoverse} is approximately 15 seconds long. 

Since the original data contains regions where the ego-vehicle is visible in some of the cameras (cf. Fig.~\ref{fig:av2_qualitative} of the main paper), we annotate each camera view with an ego-vehicle mask which we use in all experiments for all methods to constrain the ray sampling process.
We release the full data splits and the sequence alignment transformations with our source code.
\section{Method Details}
\label{sec:supp_method}
In this section, we provide more details on our method.

\parsection{Appearance and transient geometry embeddings.}
To condition our sequence-level appearance and transient geometry matrices $\appearance_\seq$ and $\transient_\seq$ on the time $t$, we use the 1D basis function $\mathcal{F}(t)$ as mentioned in Sec.~\ref{sec:method} of the main paper. We use six as the number of frequencies of $\mathcal{F}(t)$ for both appearance and transient geometry embeddings. The resulting vectors $\omega_\seq^t$ and $\omega_o$ are in $\real^{64}$. For $\omega_\seq^t$, we learn $\appearance$ and $\transient$ per sequence both in $\real^{32 \times 6 \cdot 2 + 1}$, \ie desired latent vector size by output dimension of $\mathcal{F}(t)$. For $\omega_o$, we learn separate geometry and appearance codes per object both in $\real^{32}$. 

\parsection{Transient density $\sigma_\transient$ and color $\colvec_\transient$.}
In Eq.~\ref{eq:transient_rf} of the paper, we define the output of the transient geometry branch which is used to calculate the final static color. We blend the transient color $\colvec_\transient$ with the predicted color $\colvec_\static$ in Eq.~\ref{eq:static_rf} weighted by the densities $\sigma_\static$ and $\sigma_\transient$ analogous to Eq.~\ref{eq:dc_mixture}:
\begin{equation}
    \sigma_\static =  \sigma_\static + \sigma_\transient \, , \,
    \colvec_\static = \frac{\sigma_\static}{\sigma_\static + \sigma_\transient}\colvec_\static + \frac{\sigma_\transient}{\sigma_\static + \sigma_\transient}\colvec_\transient  .
\end{equation}

\parsection{Proposal network $\sigma_\text{prop}$.}
We align with~\cite{tancik2023nerfstudio} and use two separate proposal networks, one for each proposal sampling iteration. These proposal networks and our final static radiance field $\static$ have increasing hash table sizes, acting as a coarse-to-fine representation of the scene geometry. In contrast to previous works~\cite{tancik2023nerfstudio, barron2022mip}, we condition the proposal networks on the sequence-specific geometry codes to account for varying transient geometry across sequences in the proposal sampling stage.

\parsection{Dynamic object radiance field $\dynamic$.}
For our dynamic object radiance field $\dynamic$, we use separate shape and appearance latent vectors that condition the radiance field. In particular, we use a shape code at the network input that we concatenate with the input coordinate $\loc$ and further an appearance code that we concatenate with the direction $\dir$ at the bottleneck after density prediction. We concatenate sequence and object appearance latent vectors to propagate sequence appearance to the individual objects.

\parsection{Space contraction.} As mentioned in Sec.~\ref{sec:method} of the main paper, we follow~\cite{barron2022mip,tancik2023nerfstudio} and contract the unbounded scene space into a unit cube. In particular, we use the following function for space contraction:
\begin{equation*}
    \chi(\loc) = \begin{cases} \loc,  & ||\loc||_\infty \leq 1 \\ ( 2 - \frac{1}{||\loc||_\infty} ) \frac{\loc}{||\loc||_\infty}, & ||\loc||_\infty > 1  \end{cases}.
\end{equation*}


\parsection{Limitations.}
While our method sets a new state-of-the-art for radiance field reconstruction in dynamic urban environments under varying environmental conditions, the extremely challenging nature of the problem persists and further research in this area is needed. For example, we can much better represent highly dynamic, rigid objects such as cars, vans, trucks, and buses. Still, objects with highly intricate motions such as pedestrians or cyclists continue to be a challenge.
Another limitation stems from the inherent problem of insufficient view coverage. For areas that were not clearly visible to the ego-car, we find that the rendering quality is significantly lower. This is particularly pronounced for dynamic objects since they are only present in a single sequence. However, we note that this problem is attenuated by the initialization of radiance field $\dynamic$ with a semantic prior.
Overall, views farther away from the training trajectories would constitute an interesting addition to the evaluation setup. Yet, utilizing (partial) hold-out sequences is not suitable for our task as these would contain distinct transient geometry and dynamic objects, and possibly appearance unknown to the model. Thus, a different capturing setup would be required which is outside the scope of our work but is an interesting area for future research.

Finally, while our method improves over naive pose optimization, we note that this is a challenging problem and that large pose errors are hard to correct during reconstruction. We thus tackled this issue by pre-aligning the sequences with our offline ICP procedure. We hope that our proposed benchmark can spark further research that addresses these issues.

\section{Additional Experiments}
\label{sec:supp_exp}

\def \cropcloudl {1024px}
\def \cropcloudb {870px}
\def \cropcloudr {0px}
\def \cropcloudt {200px}

\def \boxl {0.17}
\def \boxb {0.8}
\def \boxr {0.4}
\def \boxt {0.1}

\begin{figure}[t]
\centering
\footnotesize
\setlength{\tabcolsep}{1pt}
\resizebox{1.0\linewidth}{!}{
\begin{tabular}{@{}cc@{}}
No embed.  & App. only \\

    \begin{tikzpicture}
        \node[anchor=south west,inner sep=1] (image) at (0,0) {\includegraphics[width=0.5\linewidth,trim={{\cropcloudl} {\cropcloudb} {\cropcloudr} {\cropcloudt}},clip]{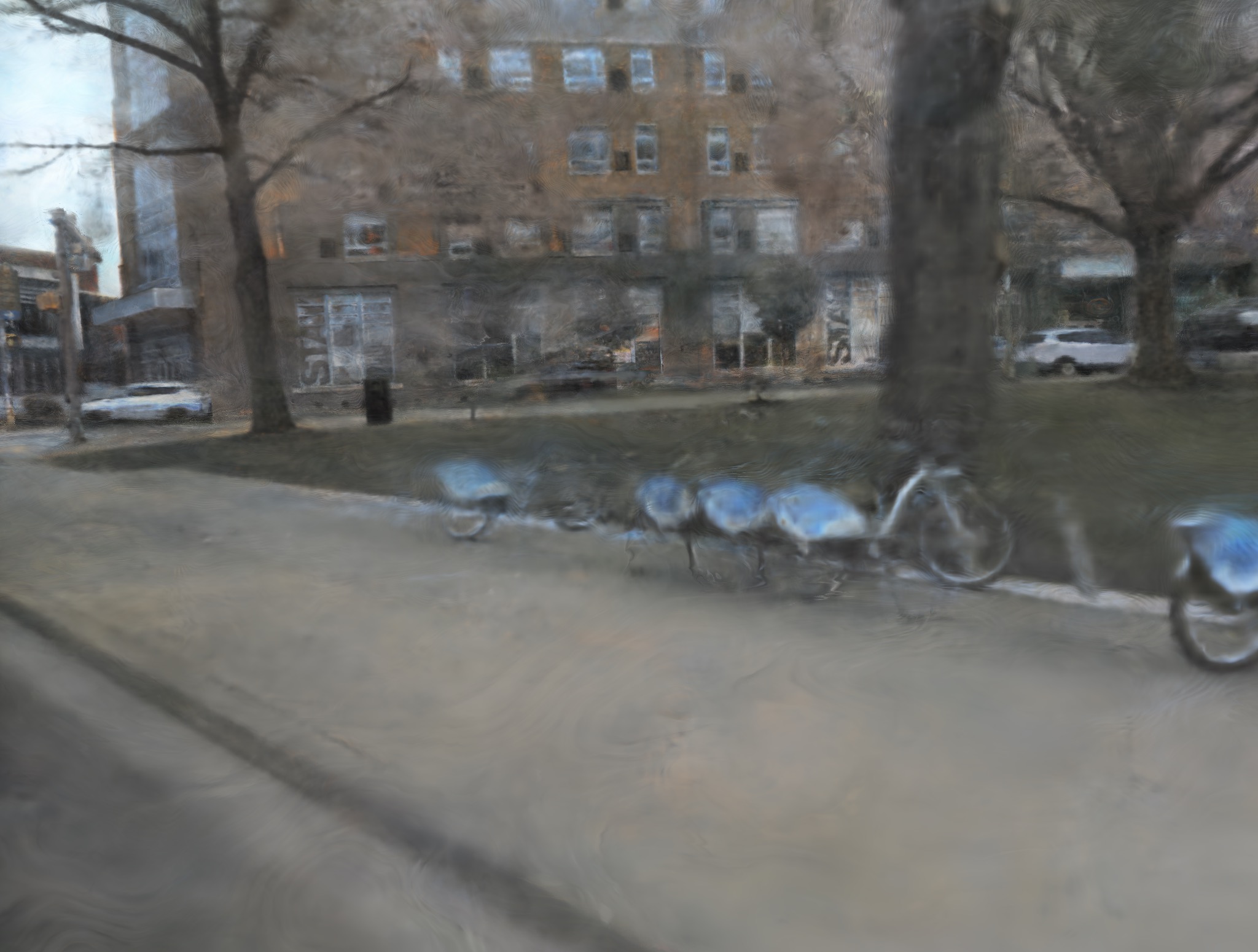}};
        \begin{scope}[x={(image.south east)},y={(image.north west)}]
            \draw[red,thick] ({\boxl},{\boxt}) rectangle ({\boxr},{\boxb});
        \end{scope}
      \end{tikzpicture}
    &
    \begin{tikzpicture}
        \node[anchor=south west,inner sep=1] (image) at (0,0) {\includegraphics[width=0.5\linewidth,trim={{\cropcloudl} {\cropcloudb} {\cropcloudr} {\cropcloudt}},clip]{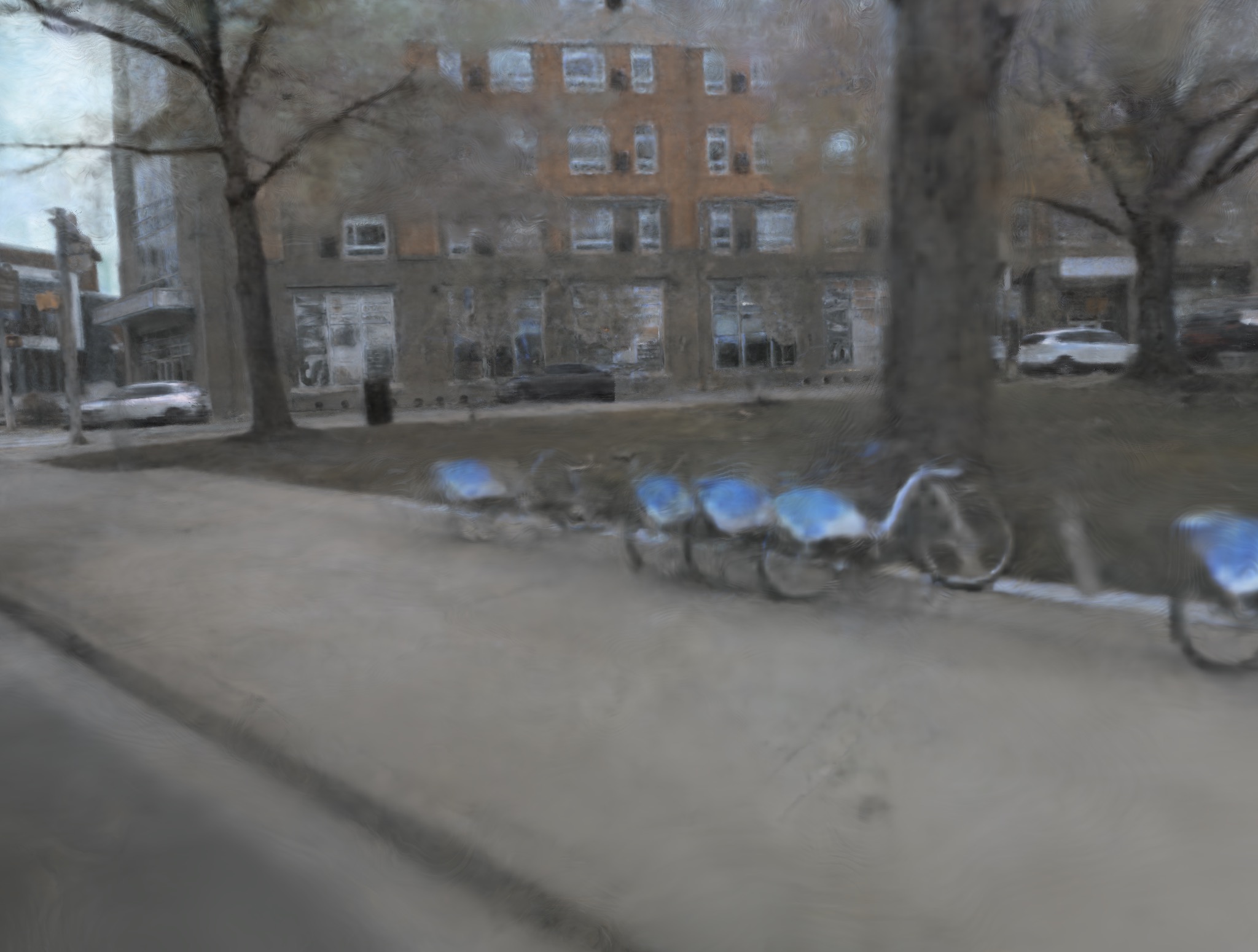}};
        \begin{scope}[x={(image.south east)},y={(image.north west)}]
            \draw[red,thick] ({\boxl},{\boxt}) rectangle ({\boxr},{\boxb});
        \end{scope}
      \end{tikzpicture}
\\
Full & GT \\
    \begin{tikzpicture}
        \node[anchor=south west,inner sep=1] (image) at (0,0) {\includegraphics[width=0.5\linewidth,trim={{\cropcloudl} {\cropcloudb} {\cropcloudr} {\cropcloudt}},clip]{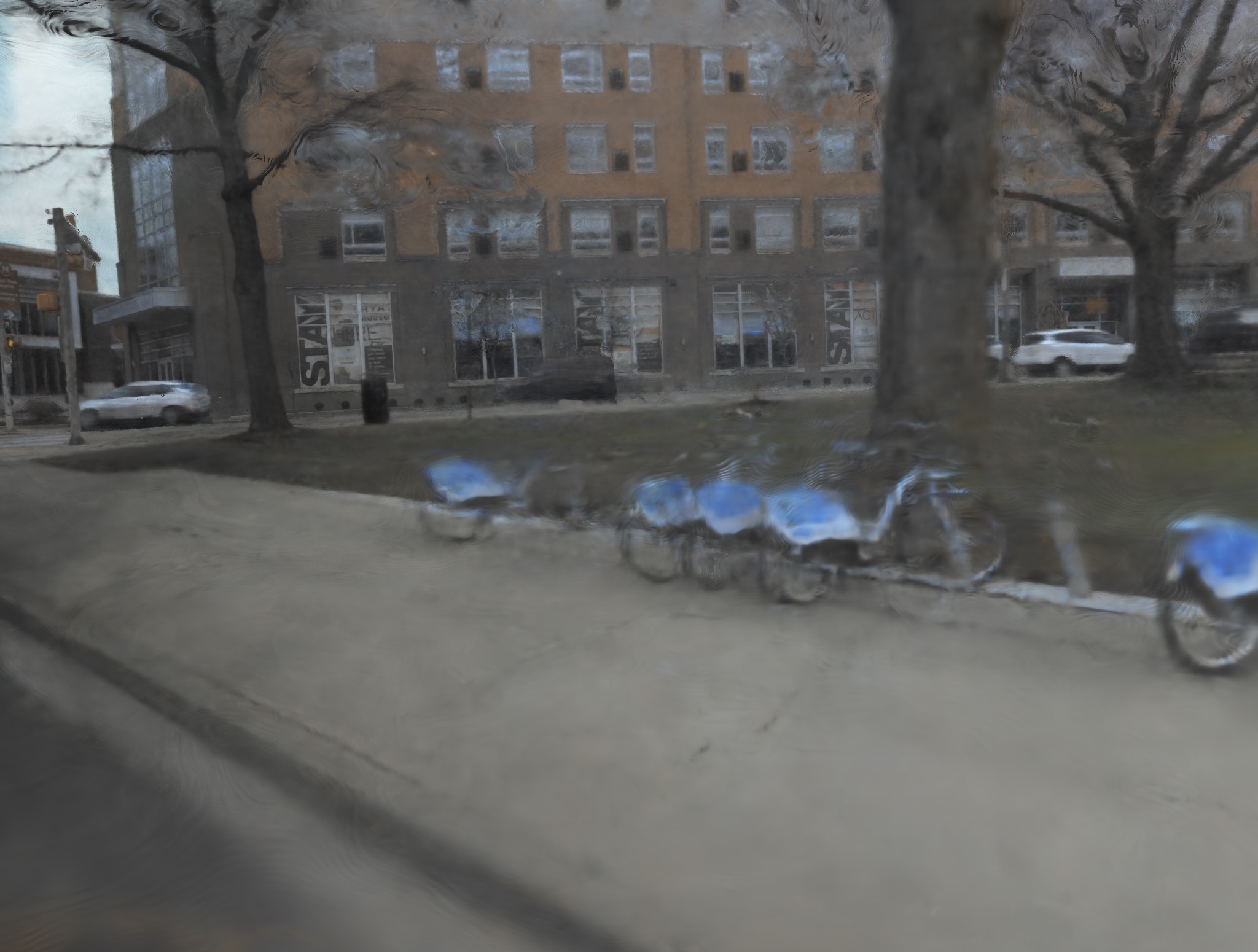}};
        \begin{scope}[x={(image.south east)},y={(image.north west)}]
            \draw[red,thick] ({\boxl},{\boxt}) rectangle ({\boxr},{\boxb});
        \end{scope}
      \end{tikzpicture}

    &
    \begin{tikzpicture}
        \node[anchor=south west,inner sep=1] (image) at (0,0) {\includegraphics[width=0.5\linewidth,trim={{\cropcloudl} {\cropcloudb} {\cropcloudr} {\cropcloudt}},clip]{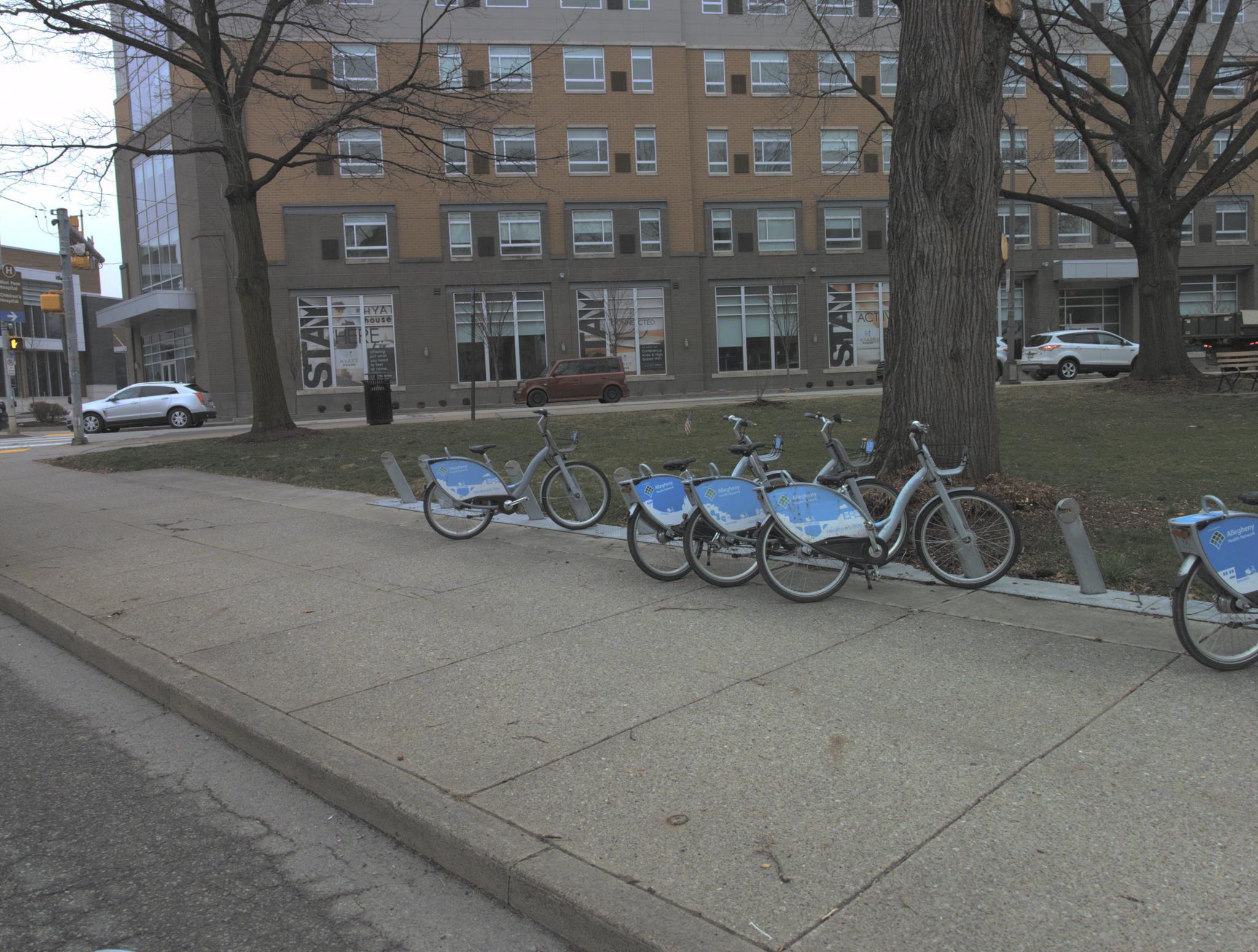}};
        \begin{scope}[x={(image.south east)},y={(image.north west)}]
            \draw[red,thick] ({\boxl},{\boxt}) rectangle ({\boxr},{\boxb});
        \end{scope}
      \end{tikzpicture}
\\
\end{tabular}}
\caption{\textbf{Qualitative comparison of graph structure.} We show a qualitative illustration of our ablation study in Tab.~\ref{tab:embedding_ablation}. In particular, we show the results of our method without any sequence-dependent latent vectors $\omega_\seq^t$, with only the appearance vectors $\appearance_\seq \mathcal{F}(t)$ and of our full method.}
\label{fig:embedding_comparison_supp}
\end{figure}


\begin{table}
\centering
\scriptsize
\begin{tabular}{l|c|ccc}
\toprule
Split & 3D Box Type & PSNR $\uparrow$ & SSIM $\uparrow$ & LPIPS $\downarrow$ \\ \midrule
\multirow{2}{*}{Single Seq.} & GT & 27.07 & 0.759 & 0.362 \\
& Prediction & 26.71 & 0.756 & 0.365 \\ \midrule
\multirow{2}{*}{Residential} & GT & 22.29 & 0.678 & 0.523 \\
& Prediction & 21.28 & 0.667 & 0.538 \\
\bottomrule
\end{tabular}
\caption{\textbf{Ablation on 3D bounding boxes.} We show results on a single sequence of the residential area in our benchmark and the full residential area. We train our method with either the provided 3D bounding box annotations or predictions acquired from an off-the-shelf 3D tracker.}
\label{tab:pred_boxes}
\end{table}

\def \cropcll {1024px}
\def \cropclb {760px}
\def \cropclr {0px}
\def \cropclt {930px}

\begin{figure}[t]
\centering
\footnotesize
\setlength{\tabcolsep}{1pt}
\resizebox{1.0\linewidth}{!}{
\begin{tabular}{@{}c|cc@{}}
GT  & Render w. GT  & Render w. Pred. \\
\includegraphics[width=0.33\linewidth,trim={{\cropcll} {\cropclb} {\cropclr} {\cropclt}},clip]{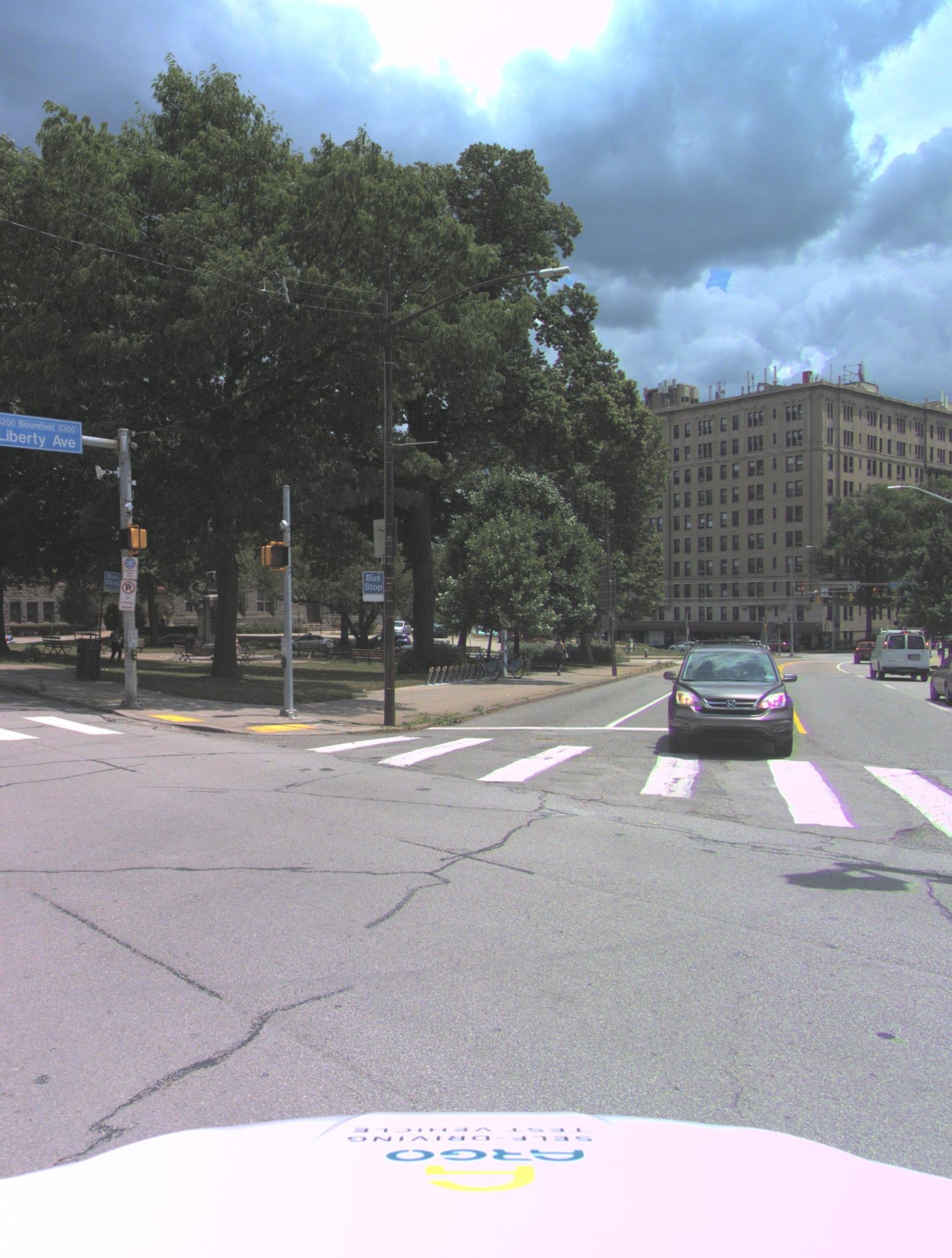} & \includegraphics[width=0.33\linewidth,trim={{\cropcll} {\cropclb} {\cropclr} {\cropclt}},clip]{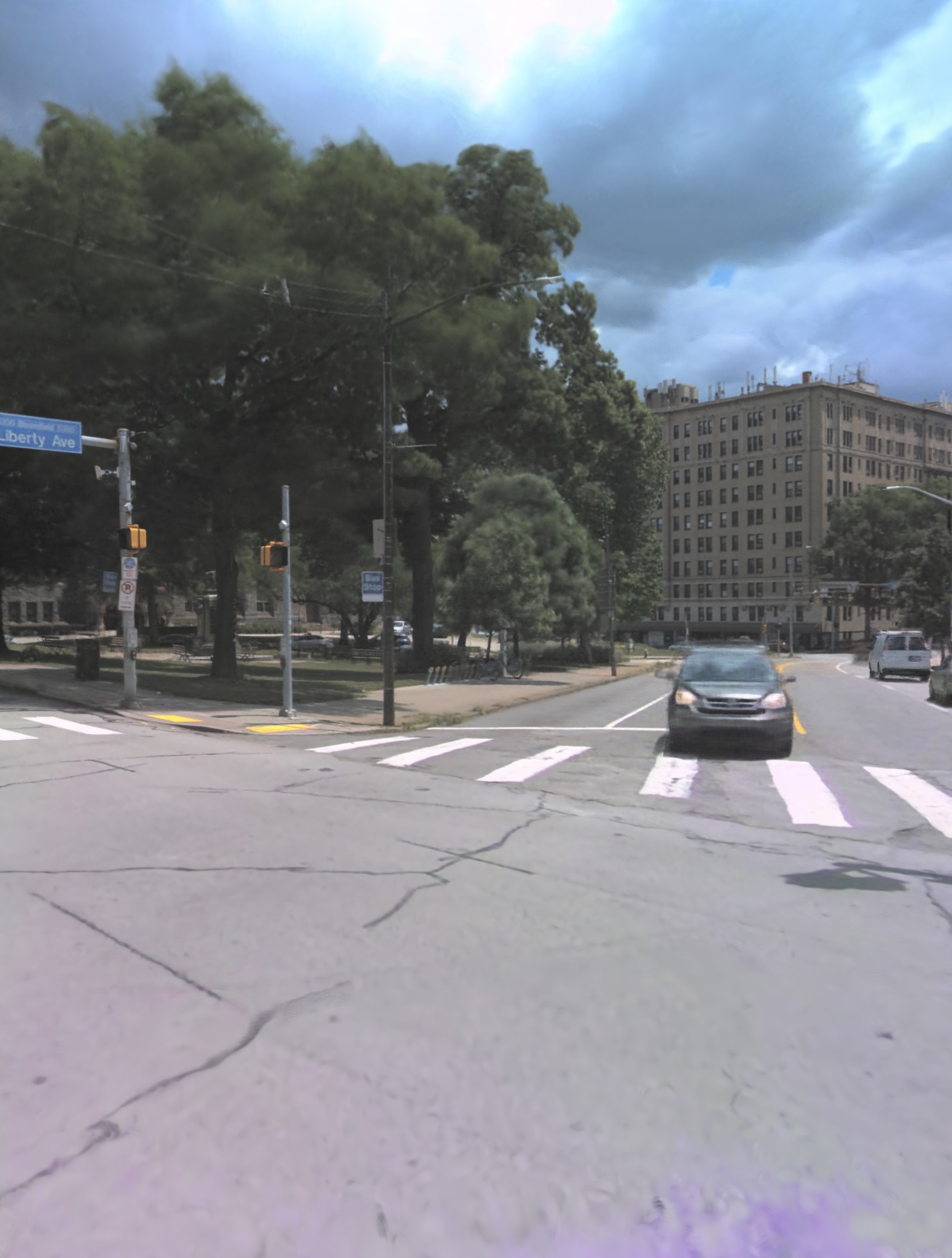}  & \includegraphics[width=0.33\linewidth,trim={{\cropcll} {\cropclb} {\cropclr} {\cropclt}},clip]{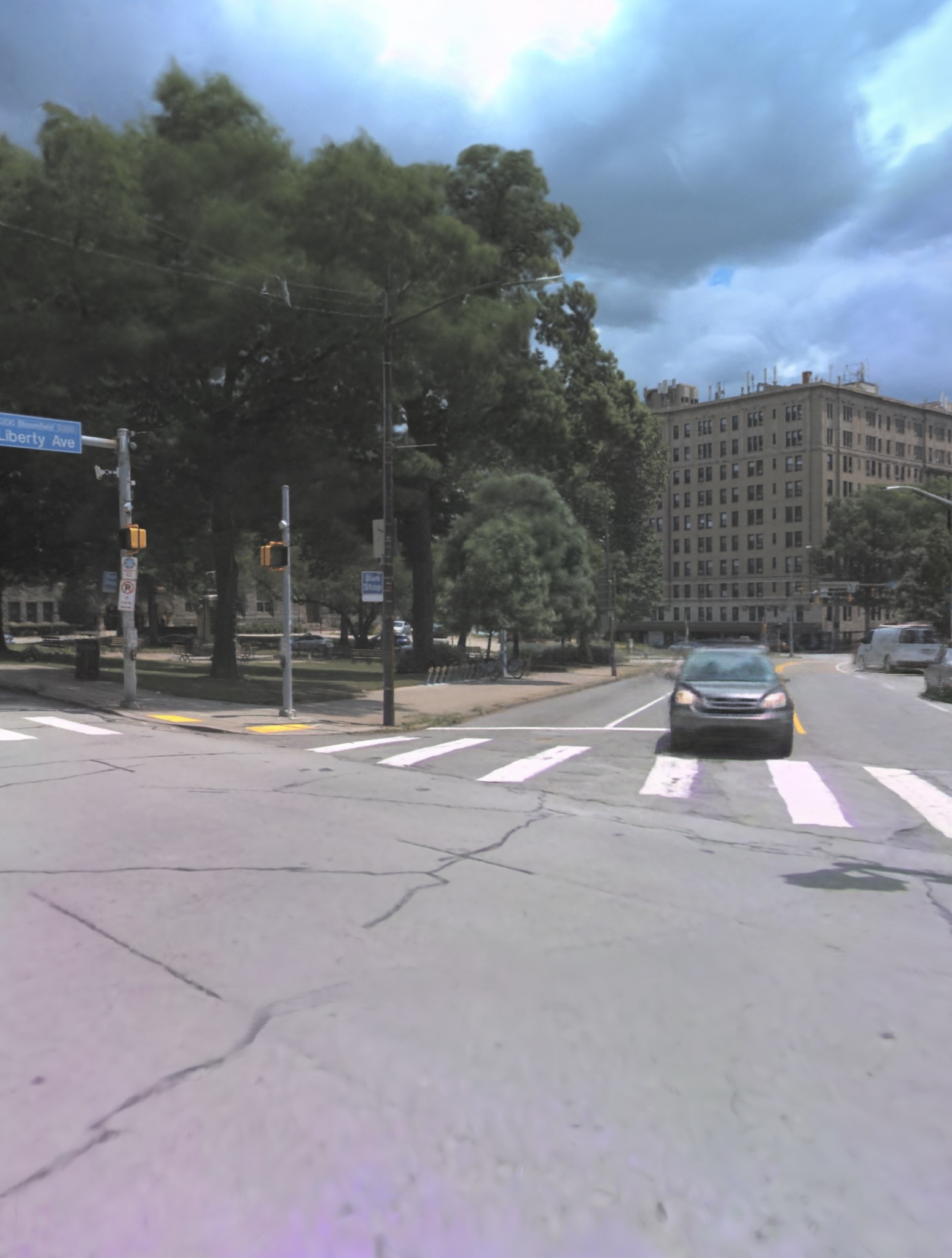} \\
\end{tabular}}
\caption{\textbf{Failure case of predicted 3D bounding boxes.} While the car in the foreground is well reconstructed with both ground truth and predicted 3D bounding boxes, the van in the background is rendered with incorrect orientation and is slightly too big.}
\label{fig:box_comparison_supp}
\end{figure}

\begin{figure}[t]
\centering
\footnotesize
\setlength{\tabcolsep}{1pt}
\resizebox{1.0\linewidth}{!}{
\begin{tabular}{@{}cc|c@{}}
&Training view & Testing view \\ \midrule
\rotatebox{90}{\hspace{2mm}SUDS} & \includegraphics[width=0.45\linewidth]{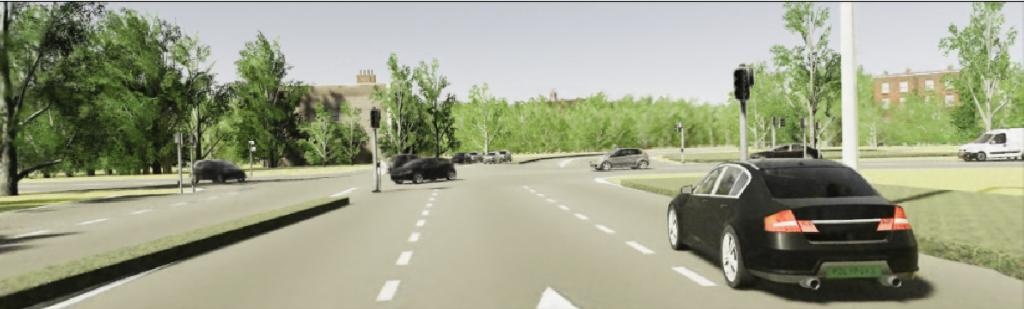} & \includegraphics[width=0.45\linewidth]{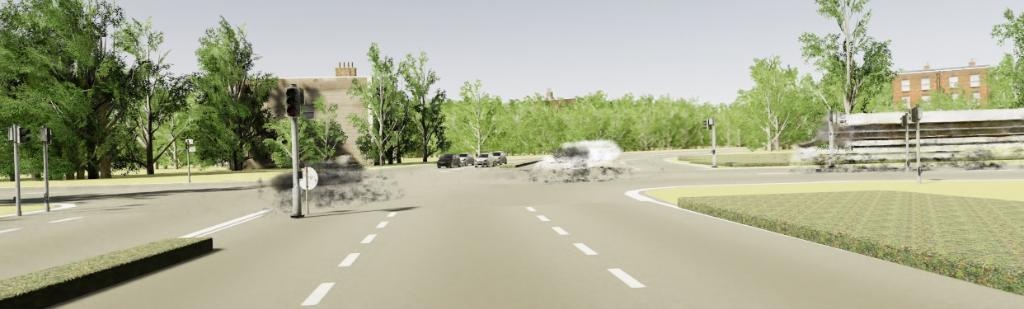}  \\
\rotatebox{90}{\hspace{3mm}Ours} & \includegraphics[width=0.45\linewidth]{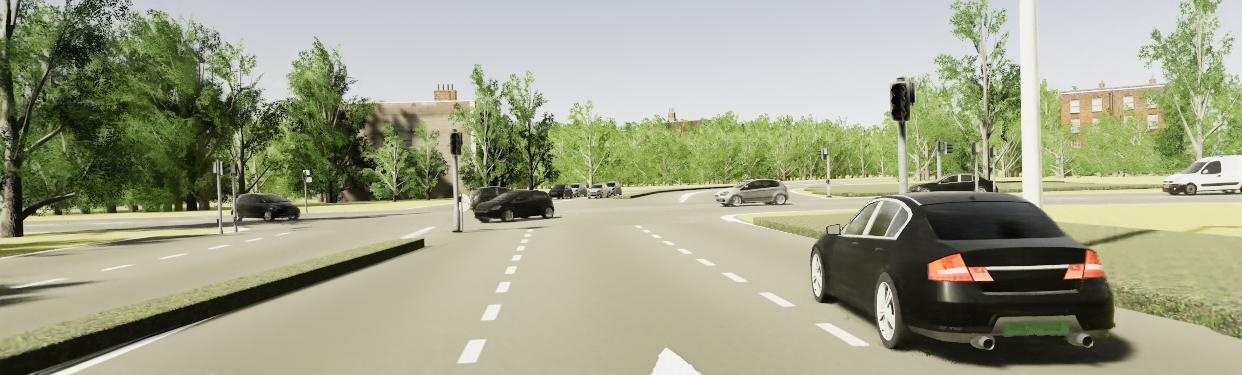} & \includegraphics[width=0.45\linewidth]{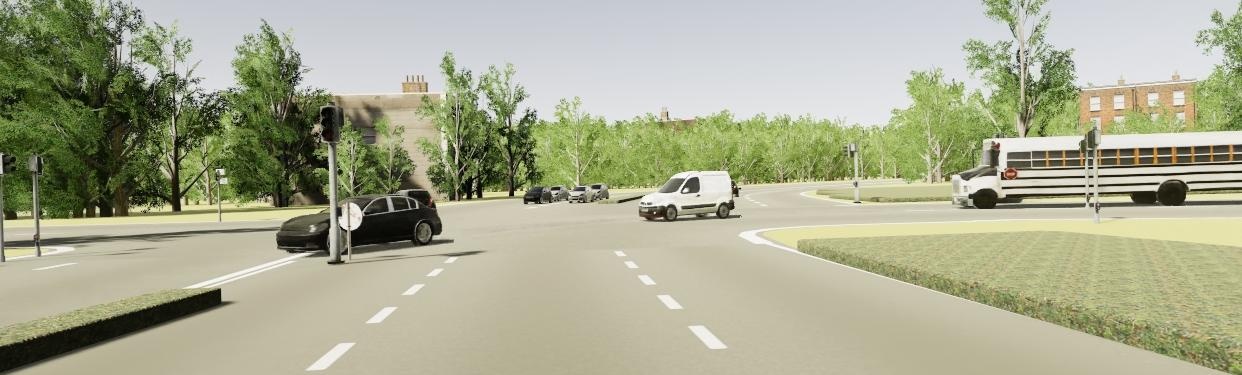}  \\
\rotatebox{90}{\hspace{4mm}GT} & \includegraphics[width=0.45\linewidth]{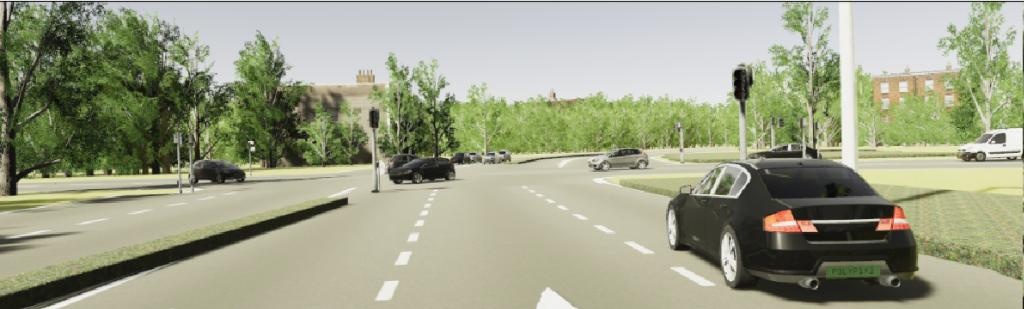} &  \includegraphics[width=0.45\linewidth]{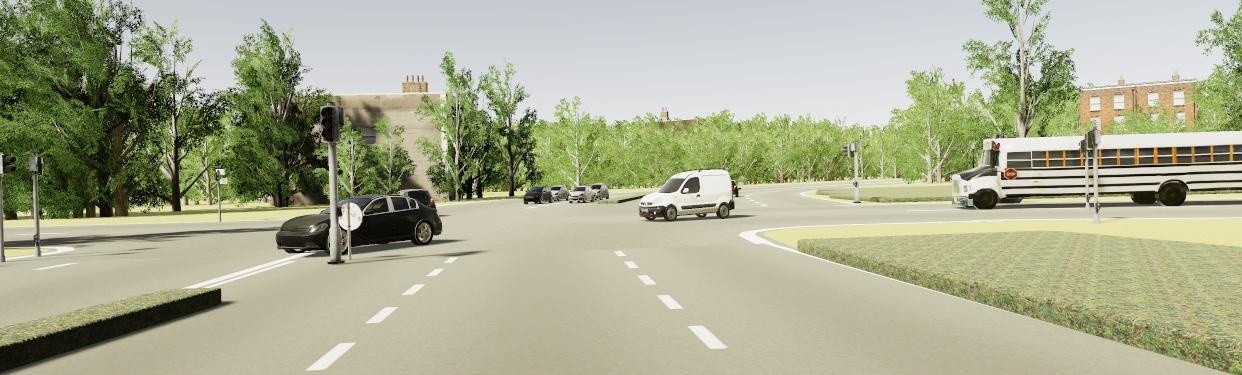} \\
\end{tabular}}
\caption{\textbf{Qualitative comparison on VKITTI2.} We observe that SUDS~\cite{turki2023suds} can render realistic training views, but cannot properly recover the dynamic actors in the testing views. In contrast, our method shows no quality difference in rendering training and novel testing views.}
\label{fig:vkitti_comparison_supp}
\end{figure}

We discuss further details on our experimental setup, additional experiments, and qualitative results. 

\parsection{Implementation details.}
We compute the scene bounds from the LiDAR point cloud with a maximum distance of 80 meters per sweep, \ie we filter each point cloud so that only points less than 80 meters from the ego-vehicle remain, and use the ego-vehicle poses to register all point clouds in the global world frame. With this global, world-frame point cloud we compute the scene bounds.
We use the following loss weights for Eq.~\ref{eq:loss} of the main paper: $\lambda_\text{dist} = 0.002$, $\lambda_\text{prop} = 1.0$, $\lambda_\text{dep} = 0.05$, and $\lambda_\text{entr} = 0.0001$. For $\loss_\text{dep}$, we use the LiDAR measurements as ground truth. We only take depth measurements at the time of the camera sensor recording to ensure dynamic objects receive valid depth supervision.
Following previous works~\cite{lin2021barf, tancik2023nerfstudio}, we optimize the evaluation camera poses during the validation phase to compensate for pose errors introduced by drifting geometry through optimized training poses that would otherwise contaminate the view synthesis quality measurement. For this, we use $\loss_\text{rgb}$ only.
For the depth map visualizations shown in our qualitative results, we use linear scaling with a maximum depth of 82.5 meters and a minimum depth of 1 meter.
Rendering a $1920\times1080$ image takes $\sim\!16.4$ seconds. The training speed is 30K rays per second.

\parsection{Baselines.}
We run SUDS~\cite{turki2023suds} on our benchmark using their official code release. Since it requires several additional inputs such as LiDAR depth and optical flow predictions, we compute the optical flow of all sequences with RAFT~\cite{teed2020raft} following the experimental setup of SUDS~\cite{turki2023suds} on their City-1M benchmark. We align all other auxiliary inputs such as depth with our method. We deactivate the DINO feature distillation branch that is used in SUDS for semantic reconstruction.

\parsection{Influence of graph structure on view quality.}
In Fig.~\ref{fig:embedding_comparison_supp}, we show a qualitative comparison of our method without the node latent codes in our graph, \ie $\appearance_\seq\mathcal{F}(t)$ and $\transient_\seq\mathcal{F}(t)$, our method with only appearance codes $\appearance_\seq\mathcal{F}(t)$, and our full method.

We observe that without any conditioning on the sequence $\seq$, there are strong artifacts, \eg the smaller tree highlighted in red is being rendered with leaves and the wall of the building behind it has a significantly different color than in the ground truth image. With the appearance embedding only, the color problem is alleviated, but the texture of the wall is highly distorted since there is no way for the model to distinguish between sequences where the tree leaves are present and the wall is occluded and where the wall is visible. In contrast, our full method recovers a faithful rendering of the tree, the wall and also the windows above.

\parsection{3D bounding box predictions.}
While we follow previous works~\cite{ost2021neural,yang2023unisim} and use provided 3D bounding boxes in our experiments, we also report results using off-the-shelf algorithms to predict the 3D bounding boxes of dynamic objects. In particular, we use an off-the-shelf LiDAR-based 3D object tracker~\cite{yin2021center,pang2021simpletrack} to generate 3D bounding box tracks. We use those tracks instead of the provided 3D bounding box annotations. Note that neither the 3D object detector nor the tracking algorithm is adjusted or fine-tuned for the Argoverse 2~\cite{wilson2023argoverse} dataset. We take the officially provided models trained on the nuScenes dataset~\cite{caesar2020nuscenes}. This dataset notably has different LiDAR sensor properties than Argoverse 2.

In Tab.~\ref{tab:pred_boxes}, we compare the results of our method with annotated 3D bounding boxes and with the predicted bounding boxes. 
We train our method both on a single sequence, \ie the same sequence as in Tab.~\ref{tab:ray_sampling} of the main paper, and the full residential area in our benchmark. When trained on a single sequence, we observe that the difference in view quality is marginal. Trained on the full residential split of our benchmark, the gap becomes slightly larger while the performance is still competitive. We analyze this more closely in Fig.~\ref{fig:box_comparison_supp}, where we observe some failure cases when predicted boxes are inaccurate, \eg when the orientation of an object is not correctly predicted and can also not be recovered through our pose optimization. Still, the synthesized views look realistic.
Overall, this shows that our approach can be scaled to large vehicle fleet data without the need for manual data annotation simply through employing off-the-shelf LiDAR 3D tracking algorithms without much loss in realism.

\parsection{Analysis of KITTI and VKITTI2 results.}
We observe a large gap between previous state-of-the-art methods and our method in terms of novel view synthesis quality in Tab.~\ref{tab:kitti_nvs}. At the same time, our image reconstruction quality is superior, but the gap is significantly smaller (cf. Tab.~\ref{tab:kitti_imrec}). Motivated by this observation, we retrain SUDS~\cite{turki2023suds} on the VKITTI2 dataset and visualize its renderings for example training and testing views alongside the results of our method in Fig.~\ref{fig:vkitti_comparison_supp}. We observe that indeed the reconstruction quality of the training views is comparable for SUDS and our method, but SUDS fails to recover dynamic objects in the testing view properly, while our method produces high-quality renderings also for novel views. This shows that our scene graph-based, high-level decomposition excels at representing scenes with highly dynamic objects while previous work struggles with this.

\parsection{Ablation of $\loss_\text{dep}$, $\loss_\text{entr}$.}
In Tab.~\ref{tab:loss_ablation}, we observe that while depth and entropy losses have a limited effect on evaluation view quality, the depth loss helps in improving geometry accuracy (AbsRel). Intuitively, this improves view quality farther from the training trajectory. The entropy loss encourages static and dynamic radiance separation, improving object renderings and scene decomposition. We illustrate these effects in Fig.~\ref{fig:loss_comp}.

\begin{figure}[t]
\centering
\footnotesize
\setlength{\tabcolsep}{1pt}
\resizebox{1.0\linewidth}{!}{
\begin{tabular}{@{}c|c|c|c|c|c@{}}
\includegraphics[width=0.25\linewidth]{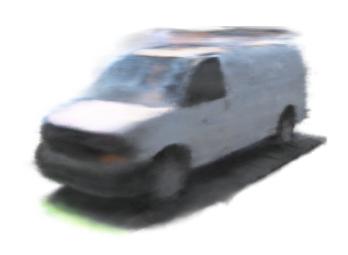} &
\includegraphics[width=0.25\linewidth]{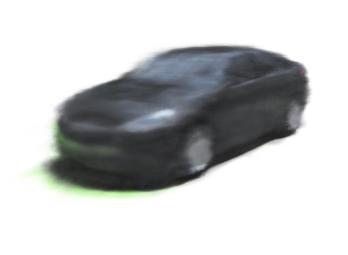} &
\includegraphics[width=0.25\linewidth]{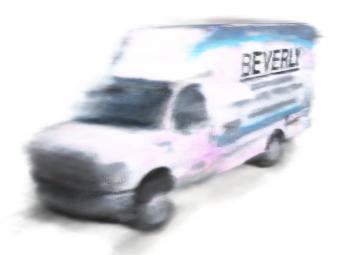} &
\includegraphics[width=0.25\linewidth]{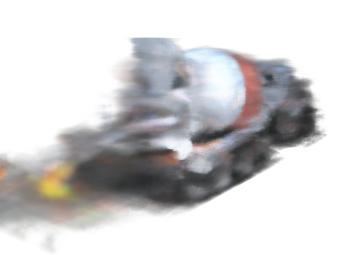} &
\includegraphics[width=0.25\linewidth]{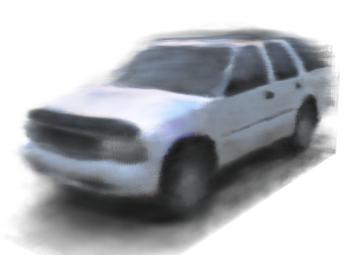} &
\includegraphics[width=0.25\linewidth]{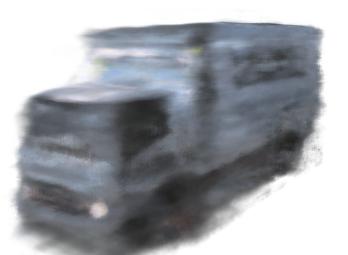} \\
\includegraphics[width=0.25\linewidth]{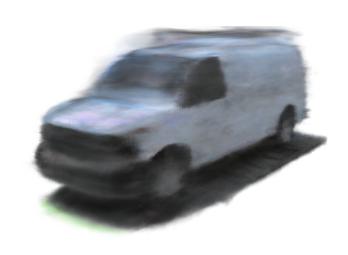} &
\includegraphics[width=0.25\linewidth]{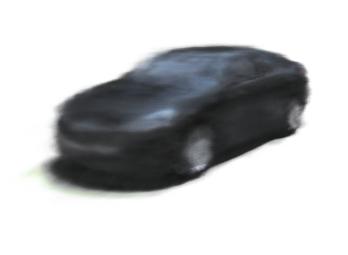} &
\includegraphics[width=0.25\linewidth]{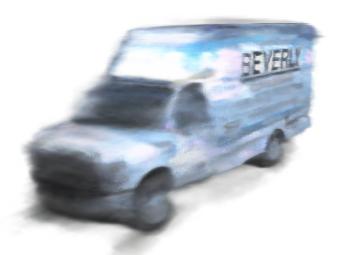} &
\includegraphics[width=0.25\linewidth]{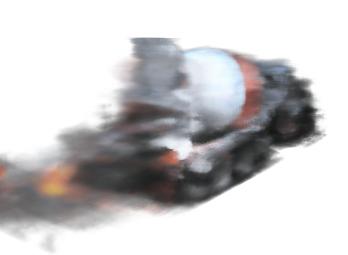} &
\includegraphics[width=0.25\linewidth]{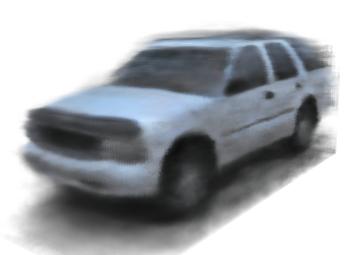} &
\includegraphics[width=0.25\linewidth]{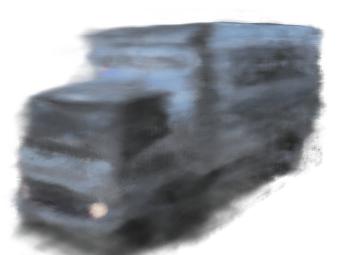} \\
\end{tabular}}
\caption{\textbf{Object renderings.} We illustrate object instances in sunny conditions (top) and cloudy conditions (bottom). }
\label{fig:car_appearance_supp}
\end{figure}

\parsection{Additional qualitative results of object-centric renderings.}
In Fig.~\ref{fig:car_appearance_supp}, we show additional object-centric renderings conditioned on different scene appearances. In particular, we depict objects with more intricate textures, showing the ability of $\dynamic$ to generalize to a wide variety of object instances. Note that the instance reconstruction quality varies with the observed training views (cf. Fig.~\ref{fig:car_appearance_supp} right). Yet, we note that our method models objects much better than existing works.

\begin{table}[t]
\centering
\scriptsize
\begin{tabular}{@{}ccc|cccc@{}}
\toprule
$\loss_\text{rgb}$ & $\loss_\text{dep}$ & $\loss_\text{entr}$ & PSNR $\uparrow$ & SSIM $\uparrow$ & LPIPS $\downarrow$ & AbsRel $\downarrow$ \\ \midrule
\checkmark & - & - & 22.22 & 0.675 & 0.524 & 0.321\\
\checkmark & \checkmark & - & 22.23 & 0.677 & \textbf{0.523} & 0.219 \\
\checkmark & - & \checkmark & 22.20 & 0.676 & \textbf{0.523} & 0.333 \\
\checkmark & \checkmark & \checkmark & \textbf{22.29} & \textbf{0.678} & \textbf{0.523} & \textbf{0.218} \\  \bottomrule
\end{tabular}
\caption{\textbf{Loss term ablation.} We report both the view and depth quality of our model when ablating different loss terms.}
\label{tab:loss_ablation}
\end{table}

\def \cropcloudl {1150px}
\def \cropcloudb {500px}
\def \cropcloudr {500px}
\def \cropcloudt {400px}

\def \boxl {0.01}
\def \boxb {0.15}
\def \boxr {0.45}
\def \boxt {0.7}

\def \gboxl {0.8}
\def \gboxb {0.01}
\def \gboxr {0.99}
\def \gboxt {0.95}

\begin{figure}[t]
\centering
\footnotesize
\setlength{\tabcolsep}{1pt}
\resizebox{1.0\linewidth}{!}{
\begin{tabular}{@{}ccc@{}}
$\loss_\text{rgb}$ & $\loss_\text{rgb}$, $\loss_\text{dep}$ & $\loss_\text{rgb}$, $\loss_\text{dep}, \loss_\text{entr}$ \\
    \begin{tikzpicture}
        \node[anchor=south west,inner sep=1] (image) at (0,0) {\includegraphics[width=0.5\linewidth,trim={{\cropcloudl} {\cropcloudb} {\cropcloudr} {\cropcloudt}},clip]{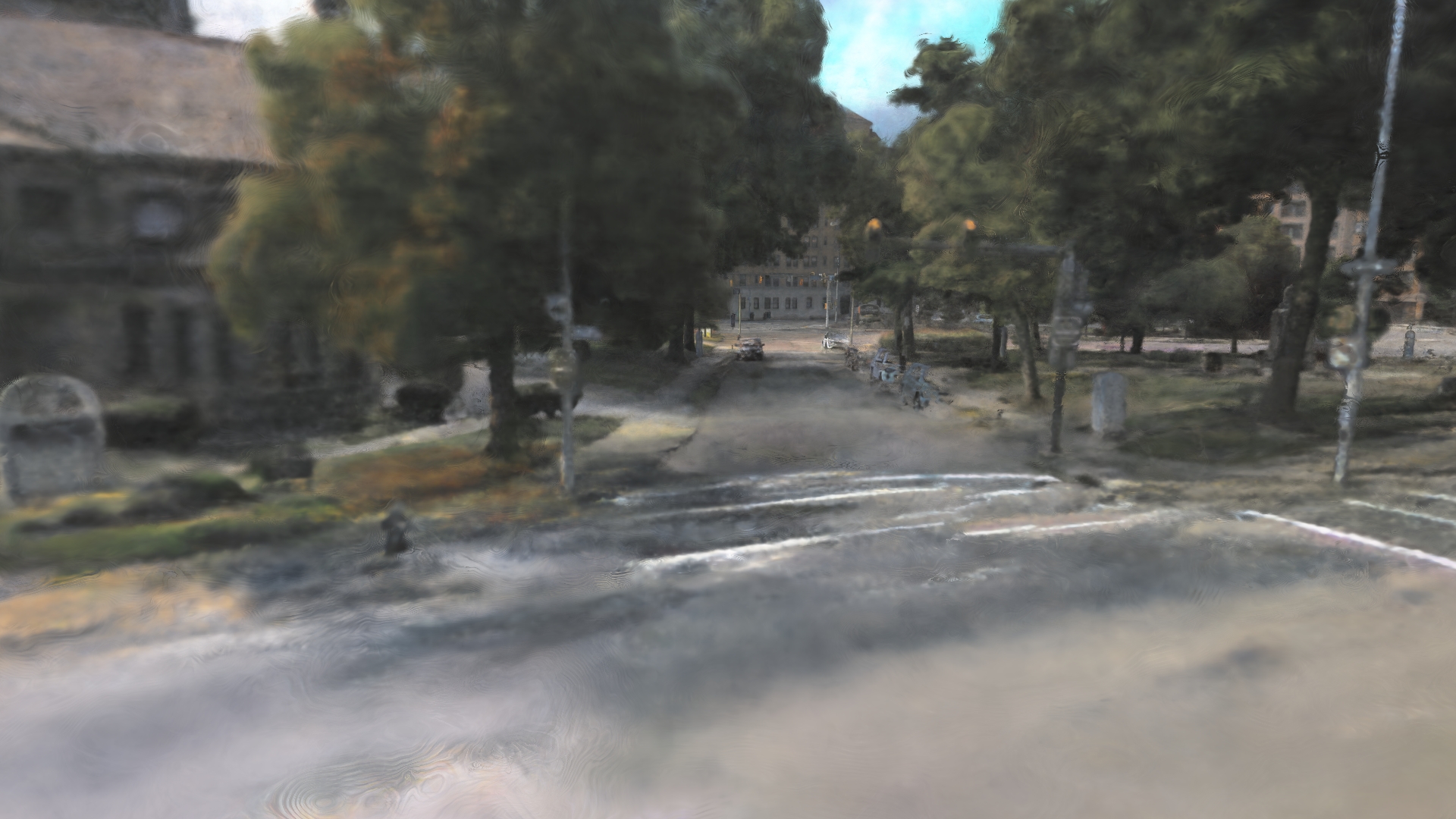}};
        \begin{scope}[x={(image.south east)},y={(image.north west)}]
            \draw[red,thick] ({\boxl},{\boxt}) rectangle ({\boxr},{\boxb});
        \end{scope}
        \begin{scope}[x={(image.south east)},y={(image.north west)}]
            \draw[green,thick] ({\gboxl},{\gboxt}) rectangle ({\gboxr},{\gboxb});
        \end{scope}
      \end{tikzpicture}
    &
    \begin{tikzpicture}
        \node[anchor=south west,inner sep=1] (image) at (0,0) {\includegraphics[width=0.5\linewidth,trim={{\cropcloudl} {\cropcloudb} {\cropcloudr} {\cropcloudt}},clip]{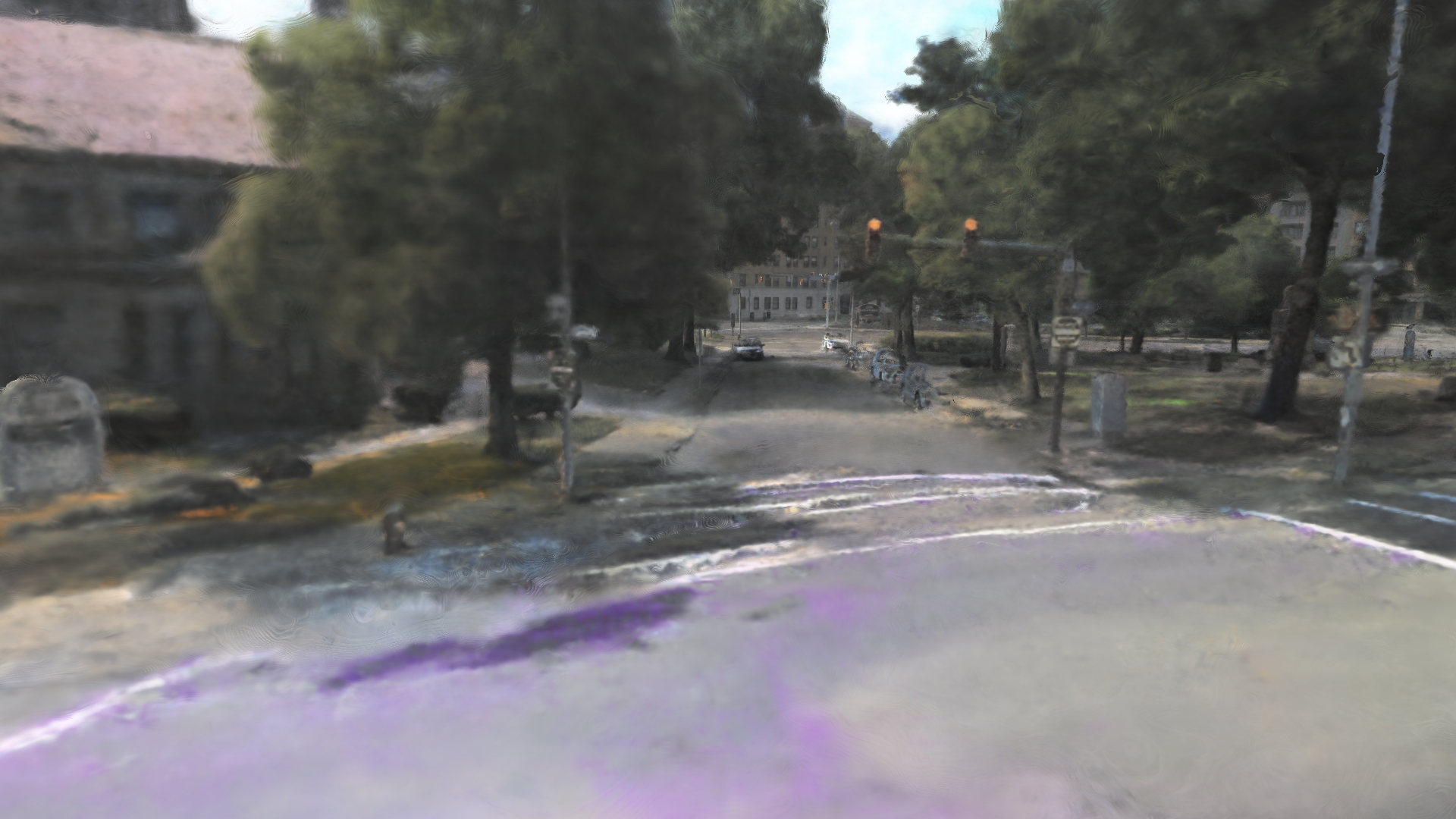}};
        \begin{scope}[x={(image.south east)},y={(image.north west)}]
            \draw[red,thick] ({\boxl},{\boxt}) rectangle ({\boxr},{\boxb});
        \end{scope}
        \begin{scope}[x={(image.south east)},y={(image.north west)}]
            \draw[green,thick] ({\gboxl},{\gboxt}) rectangle ({\gboxr},{\gboxb});
        \end{scope}
      \end{tikzpicture}
    &
    \begin{tikzpicture}
        \node[anchor=south west,inner sep=1] (image) at (0,0) {\includegraphics[width=0.5\linewidth,trim={{\cropcloudl} {\cropcloudb} {\cropcloudr} {\cropcloudt}},clip]{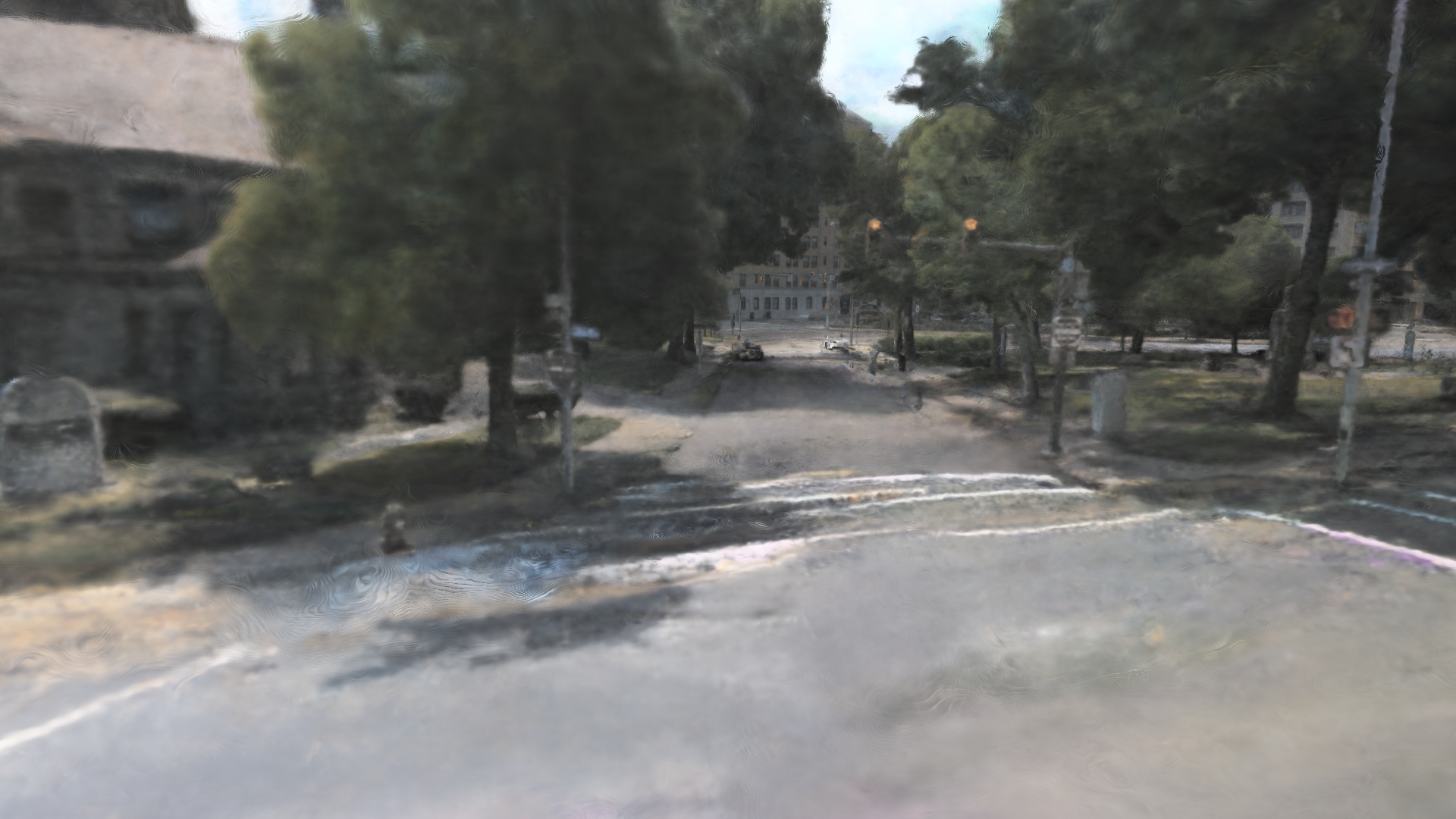}};
        \begin{scope}[x={(image.south east)},y={(image.north west)}]
            \draw[red,thick] ({\boxl},{\boxt}) rectangle ({\boxr},{\boxb});
        \end{scope}
        \begin{scope}[x={(image.south east)},y={(image.north west)}]
            \draw[green,thick] ({\gboxl},{\gboxt}) rectangle ({\gboxr},{\gboxb});
        \end{scope}
      \end{tikzpicture}
\\
\end{tabular}}
\caption{\textbf{Free viewpoint renderings.} Without $\loss_\text{entr}$, the separation between dynamic and static content is ambiguous (red). Without $\loss_\text{dep}$, the traffic pole exhibits artifacts (green).}
\label{fig:loss_comp}
\end{figure}


\begin{figure*}[t]
\centering
\footnotesize
\setlength{\tabcolsep}{1pt}
\resizebox{1.0\linewidth}{!}{
\begin{tabular}{@{}cccc|c@{}}
\toprule
 Nerfacto + Emb. & Nerfacto + Emb. + Time & SUDS~\cite{turki2023suds} & Ours & Ground Truth \\
\midrule
\includegraphics[width=0.2\linewidth]{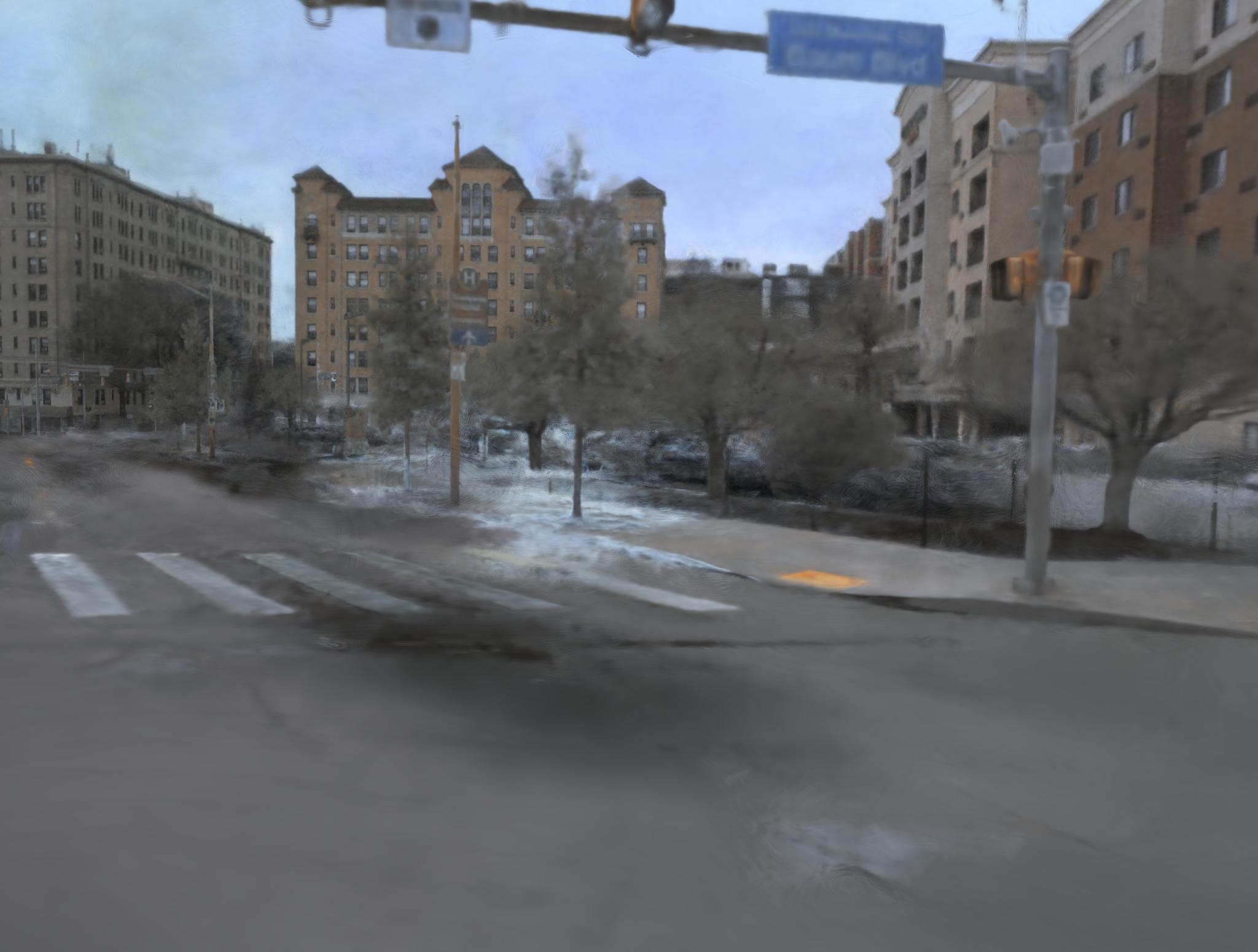} &
\includegraphics[width=0.2\linewidth]{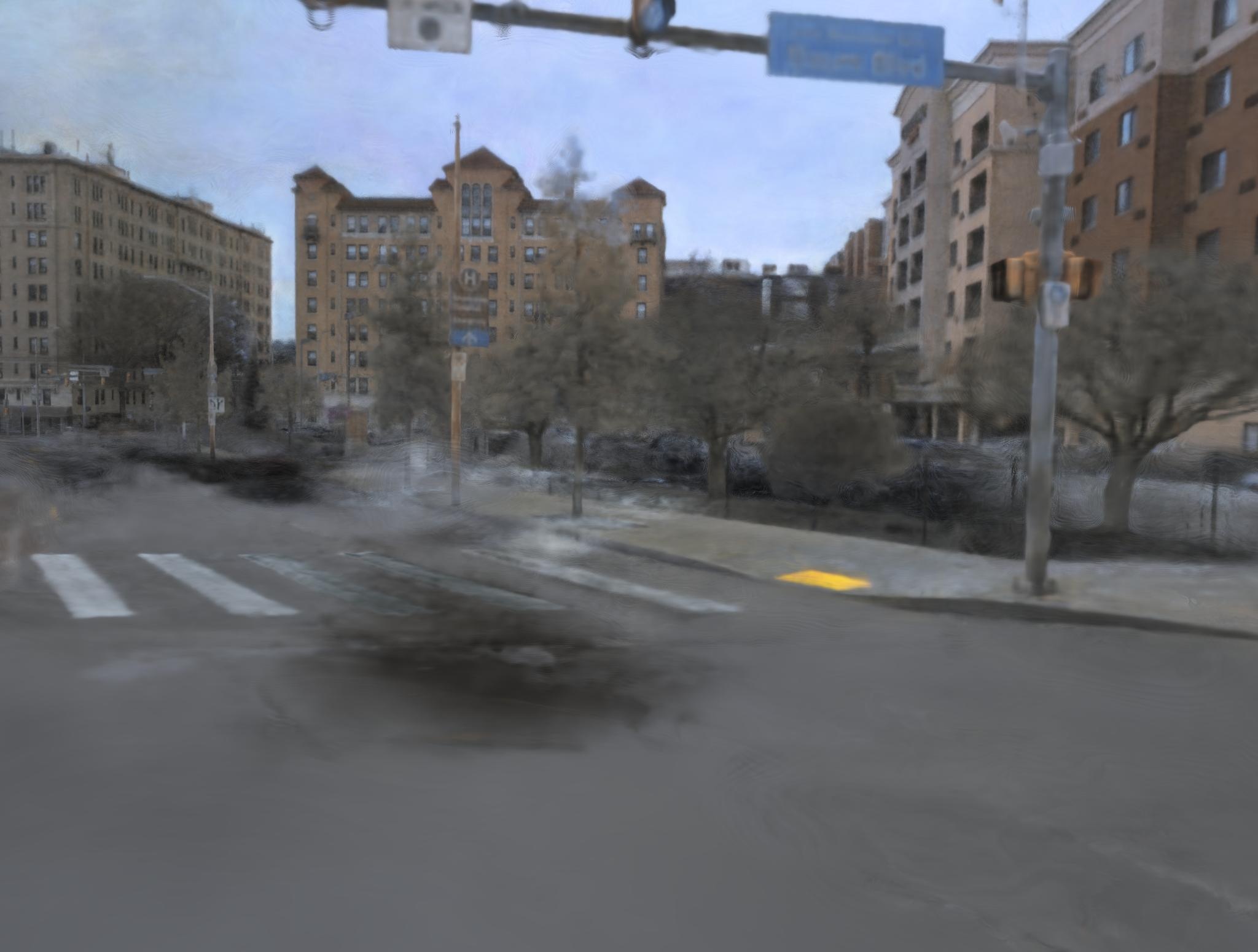} &
\includegraphics[width=0.2\linewidth]{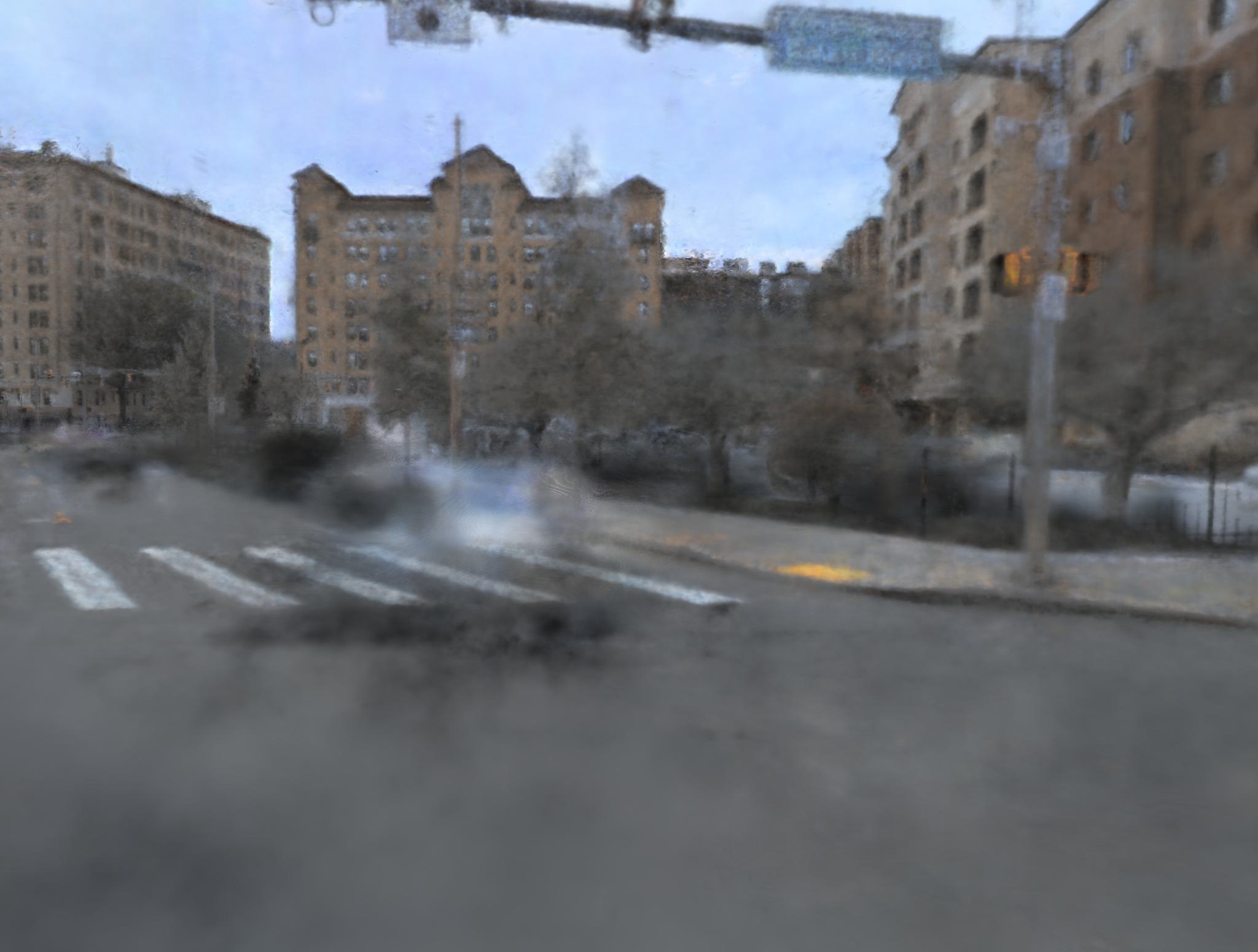} &
\includegraphics[width=0.2\linewidth]{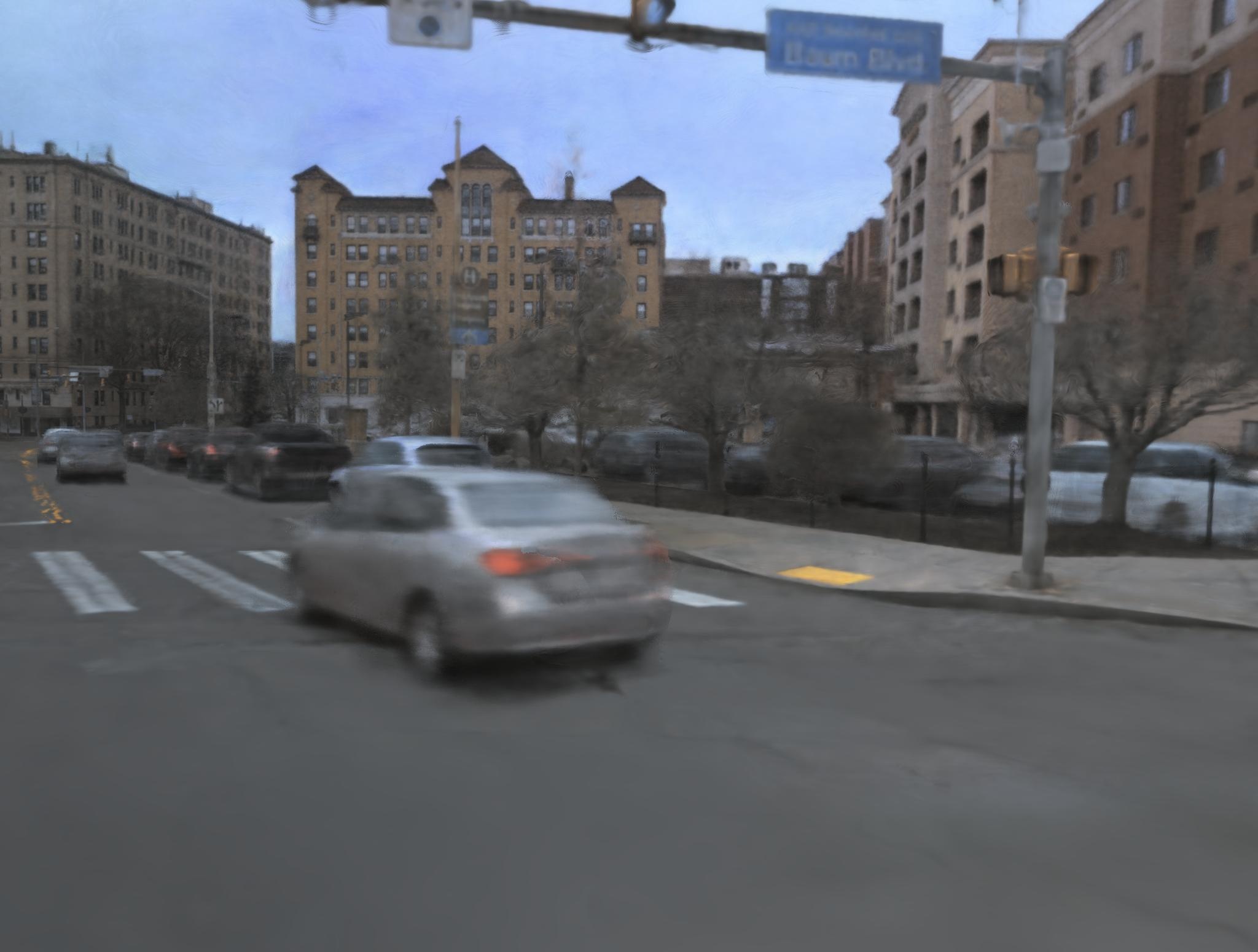} &
\includegraphics[width=0.2\linewidth]{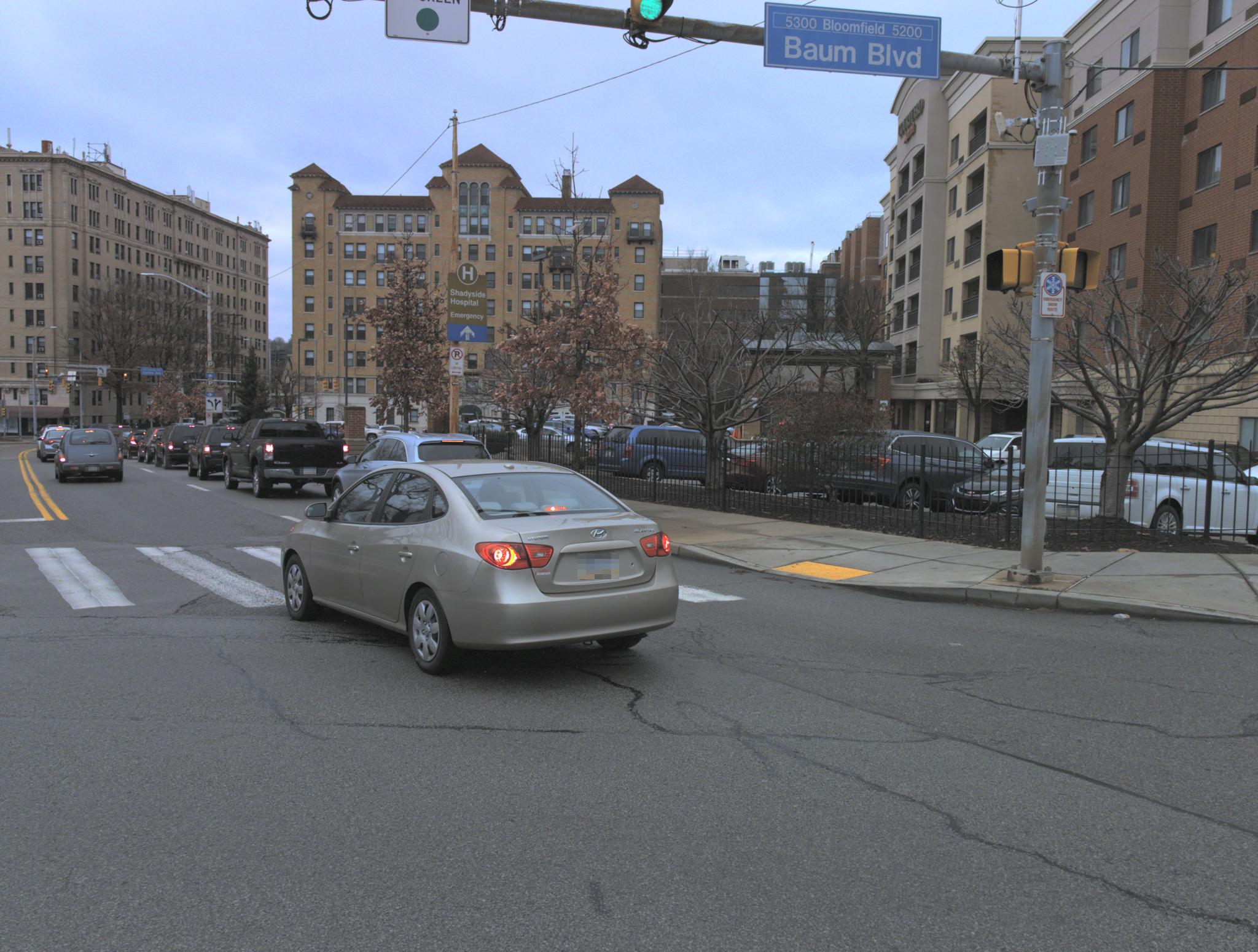} \\
\includegraphics[width=0.2\linewidth]{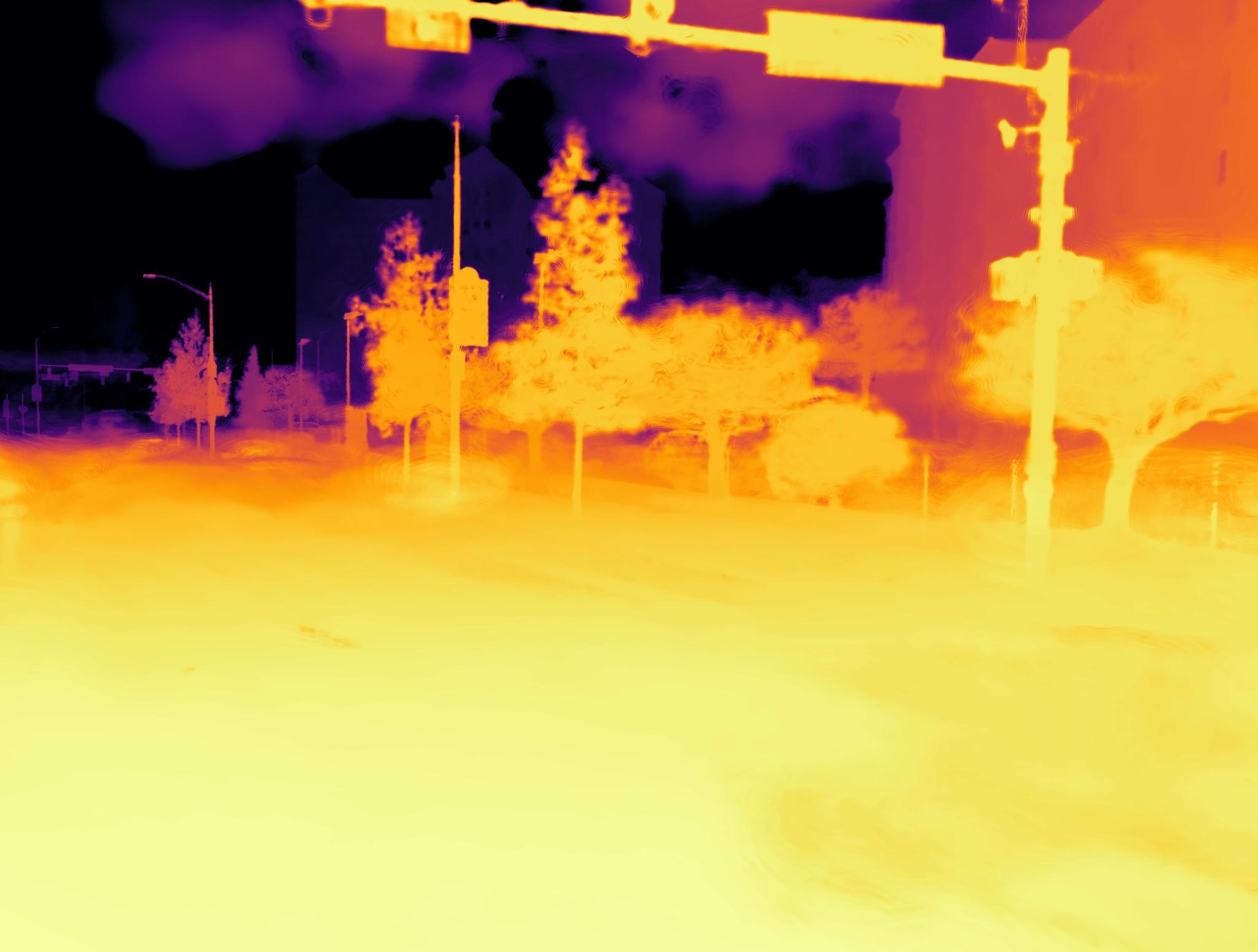} &
\includegraphics[width=0.2\linewidth]{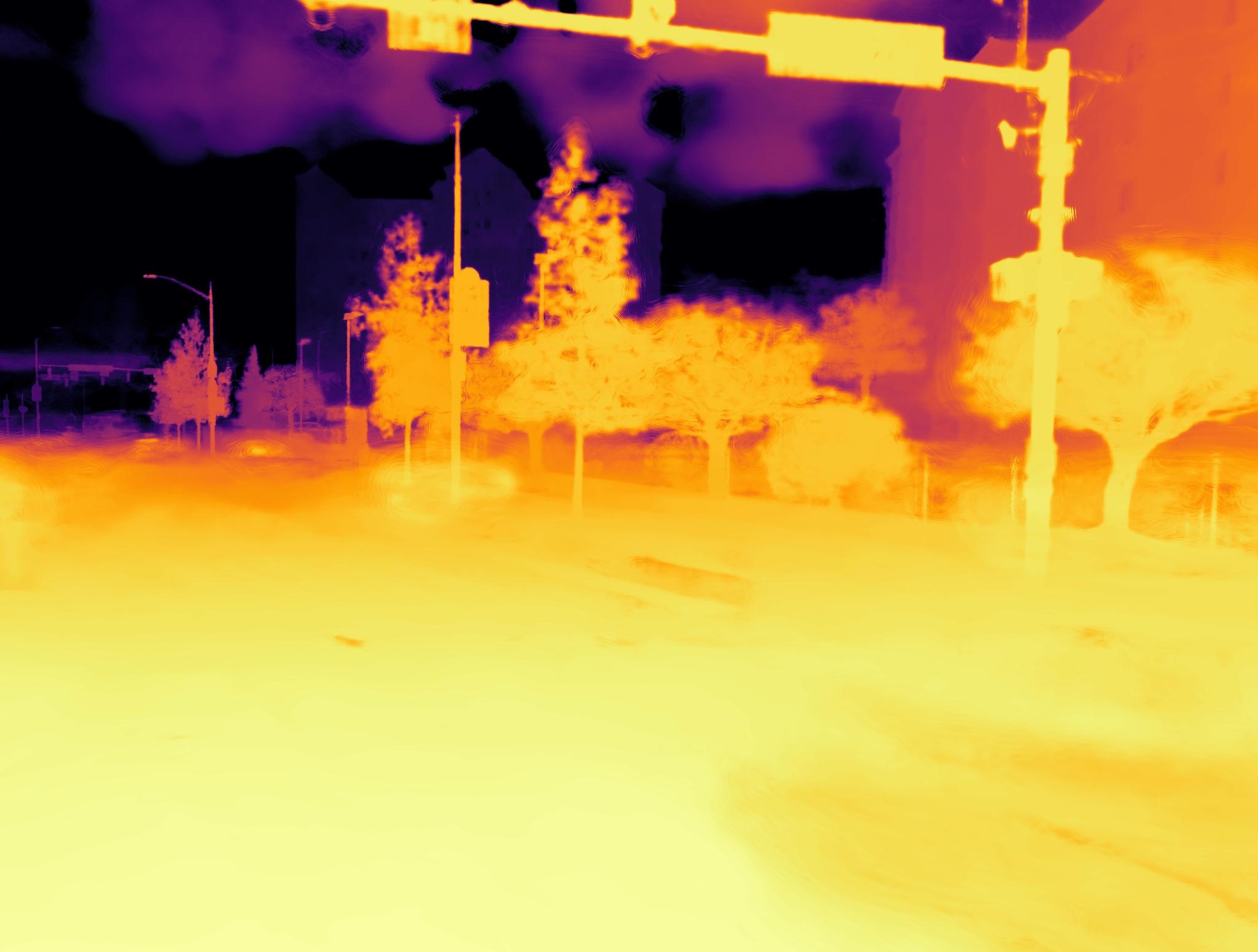} &
\includegraphics[width=0.2\linewidth]{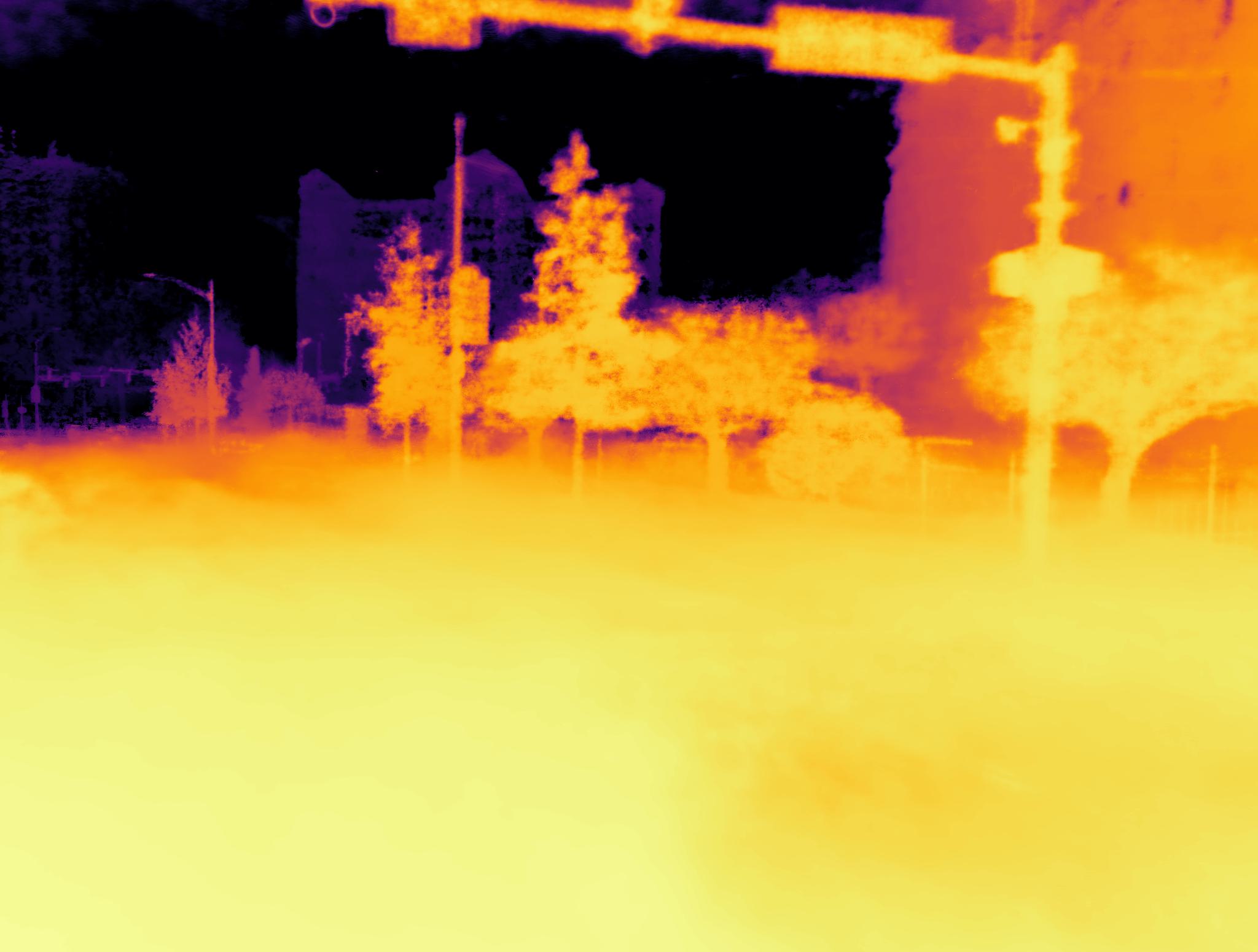} &
\includegraphics[width=0.2\linewidth]{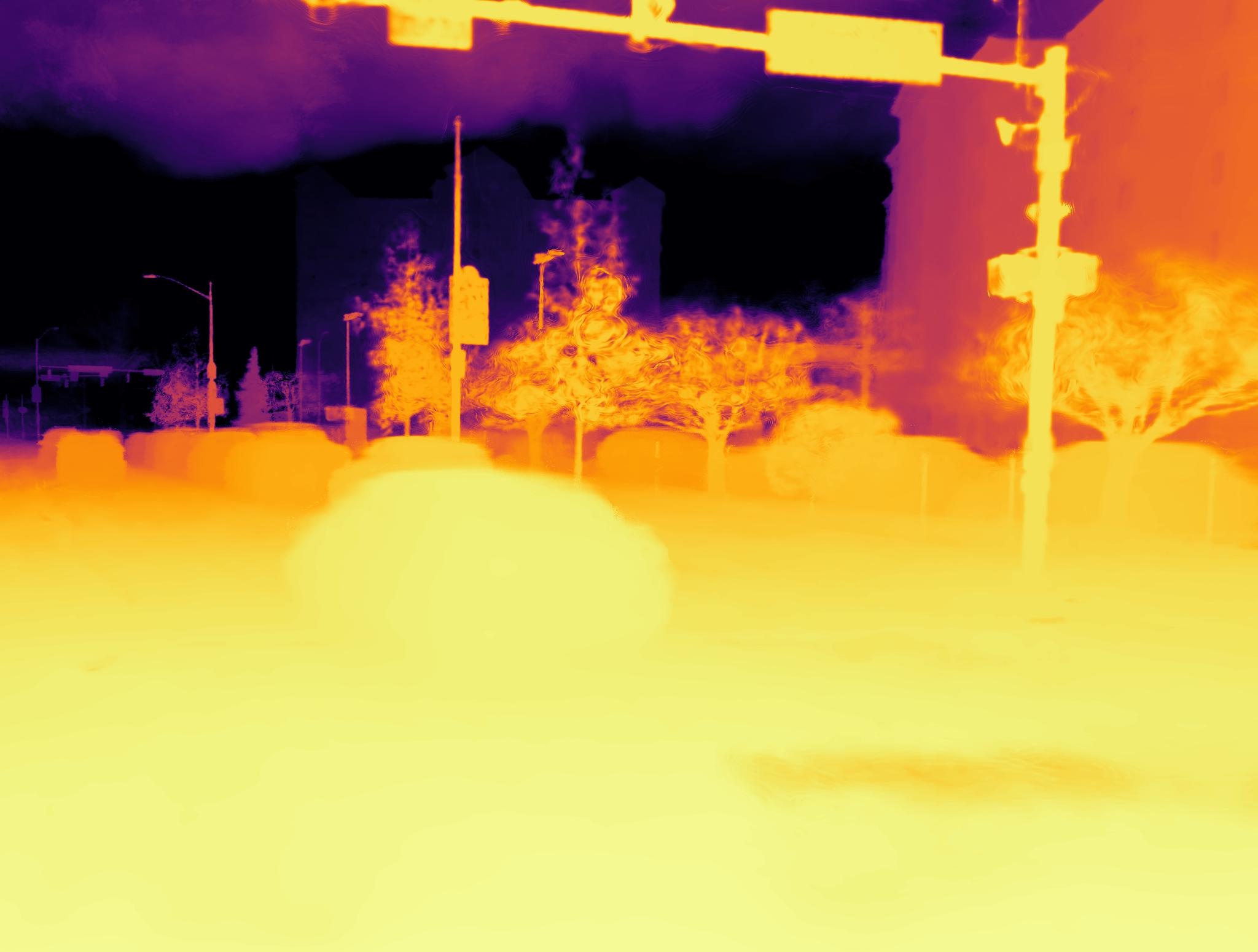} &
\includegraphics[width=0.2\linewidth]{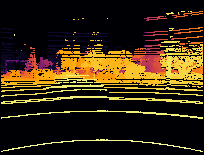} \\

\includegraphics[width=0.2\linewidth]{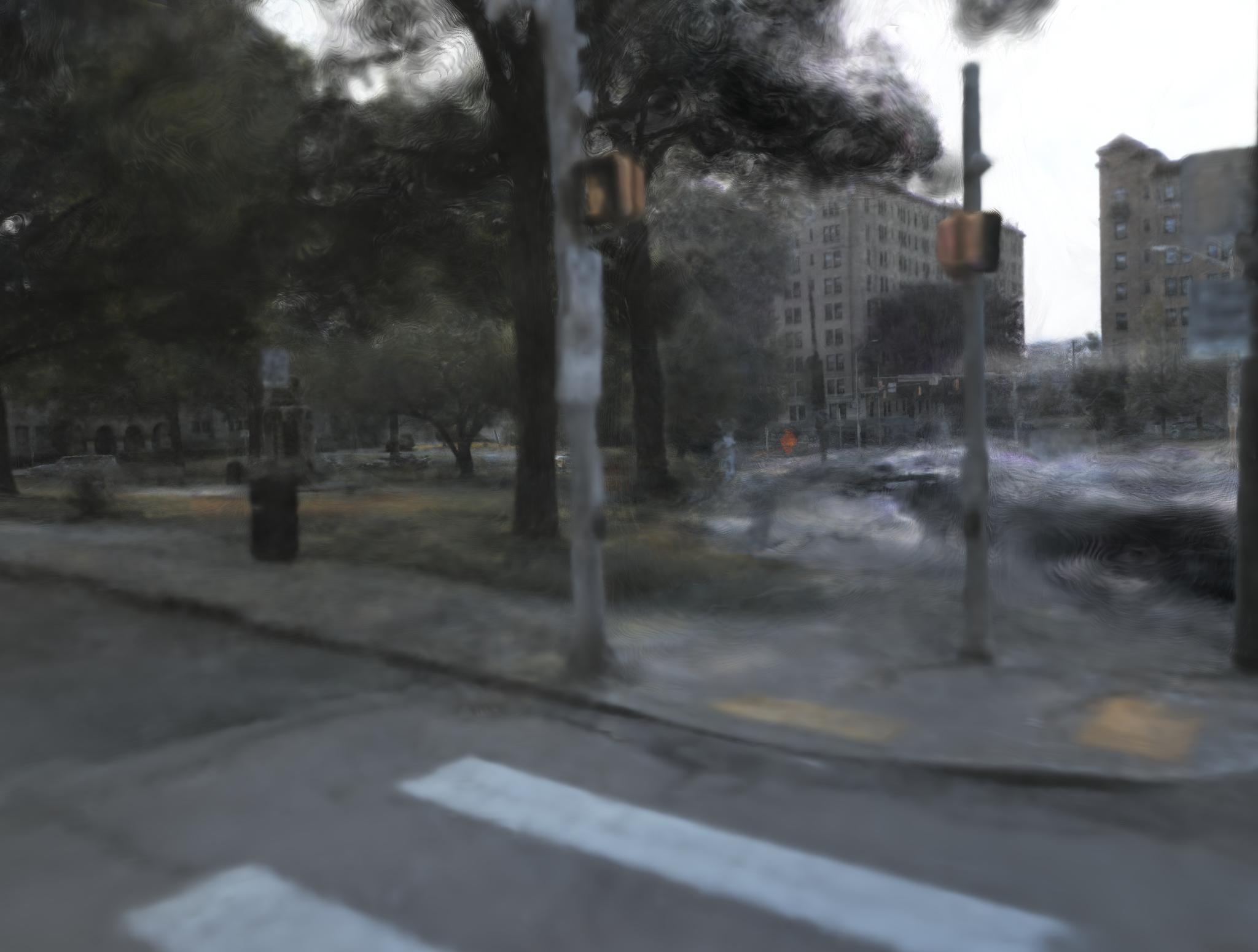} &
\includegraphics[width=0.2\linewidth]{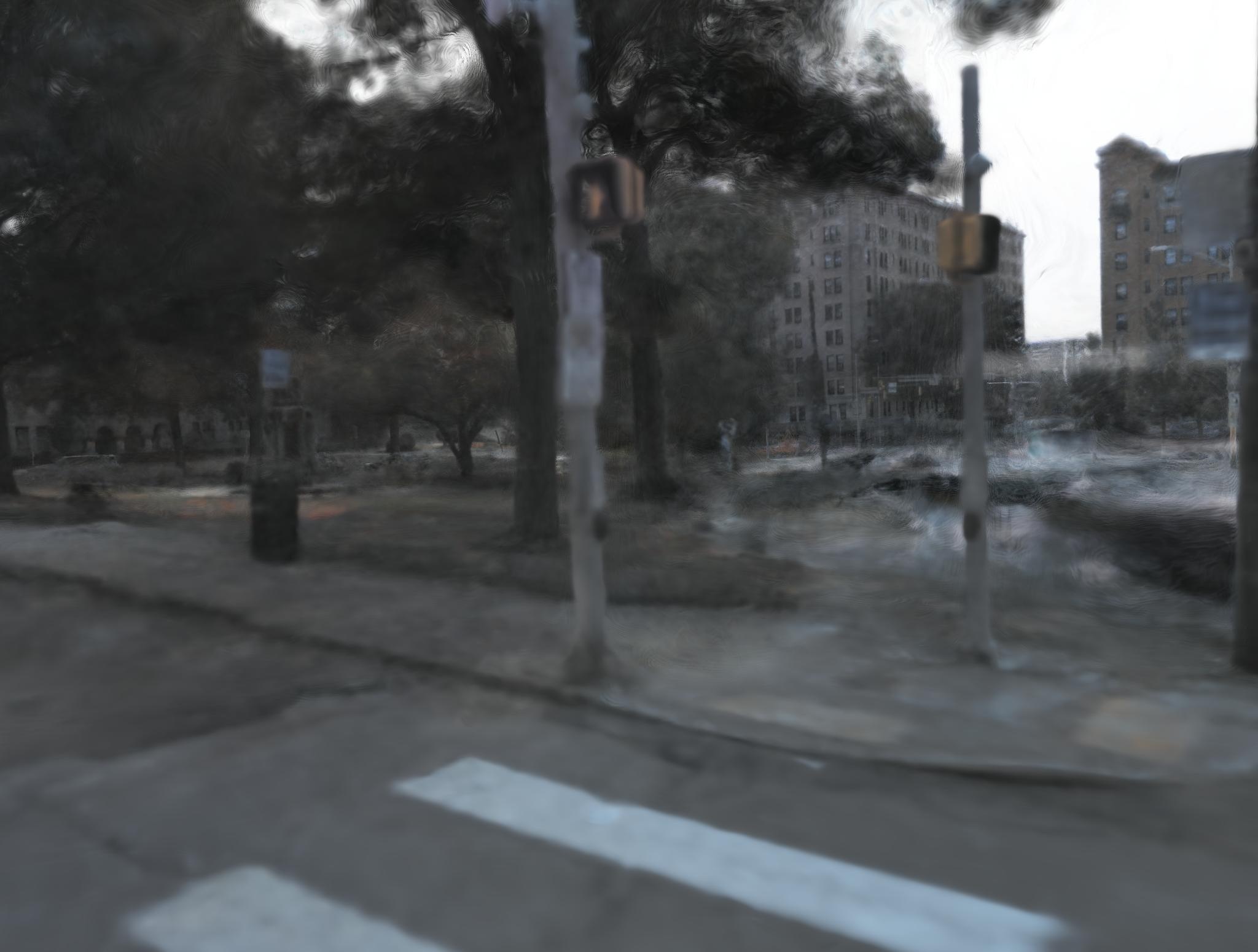} &
\includegraphics[width=0.2\linewidth]{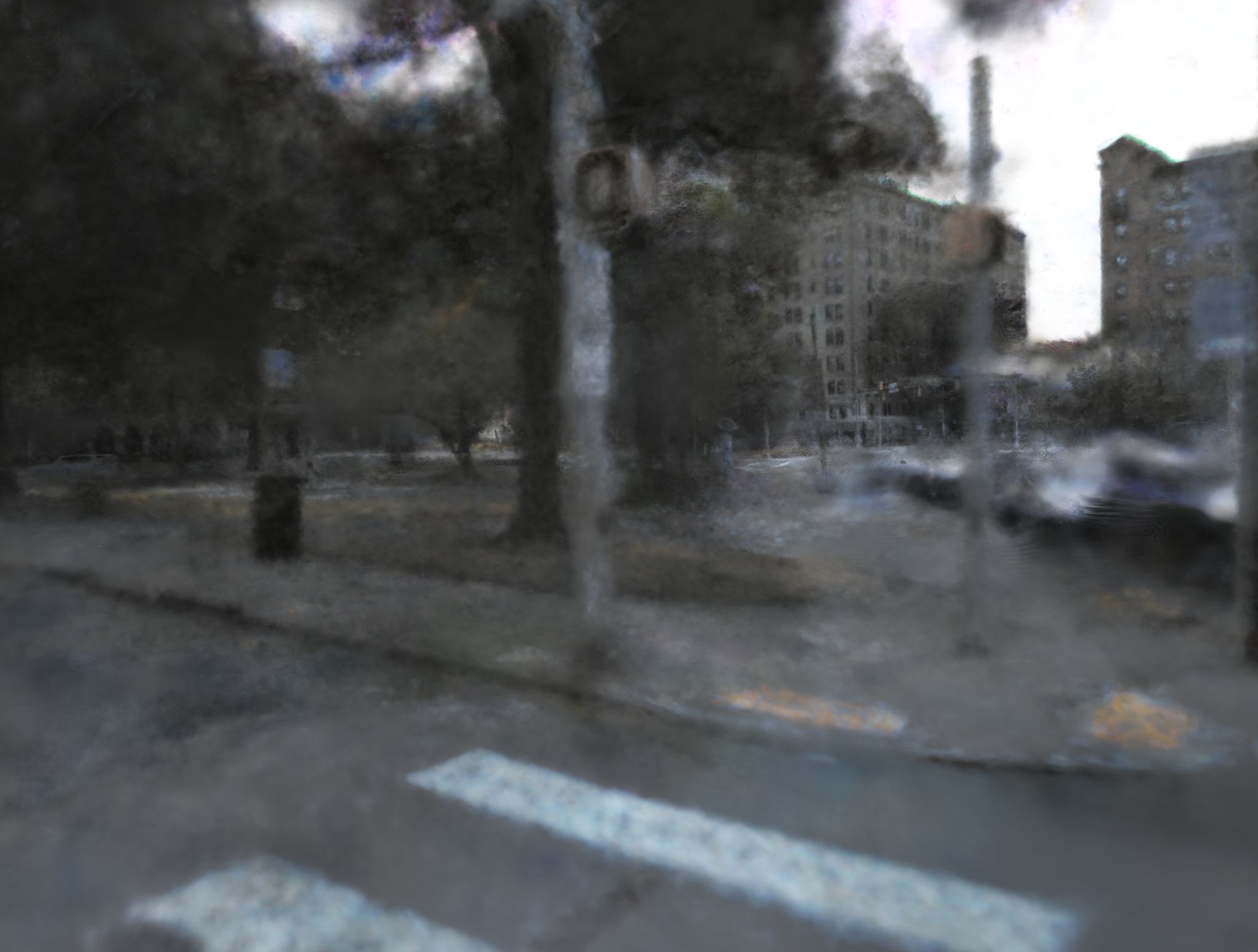} &
\includegraphics[width=0.2\linewidth]{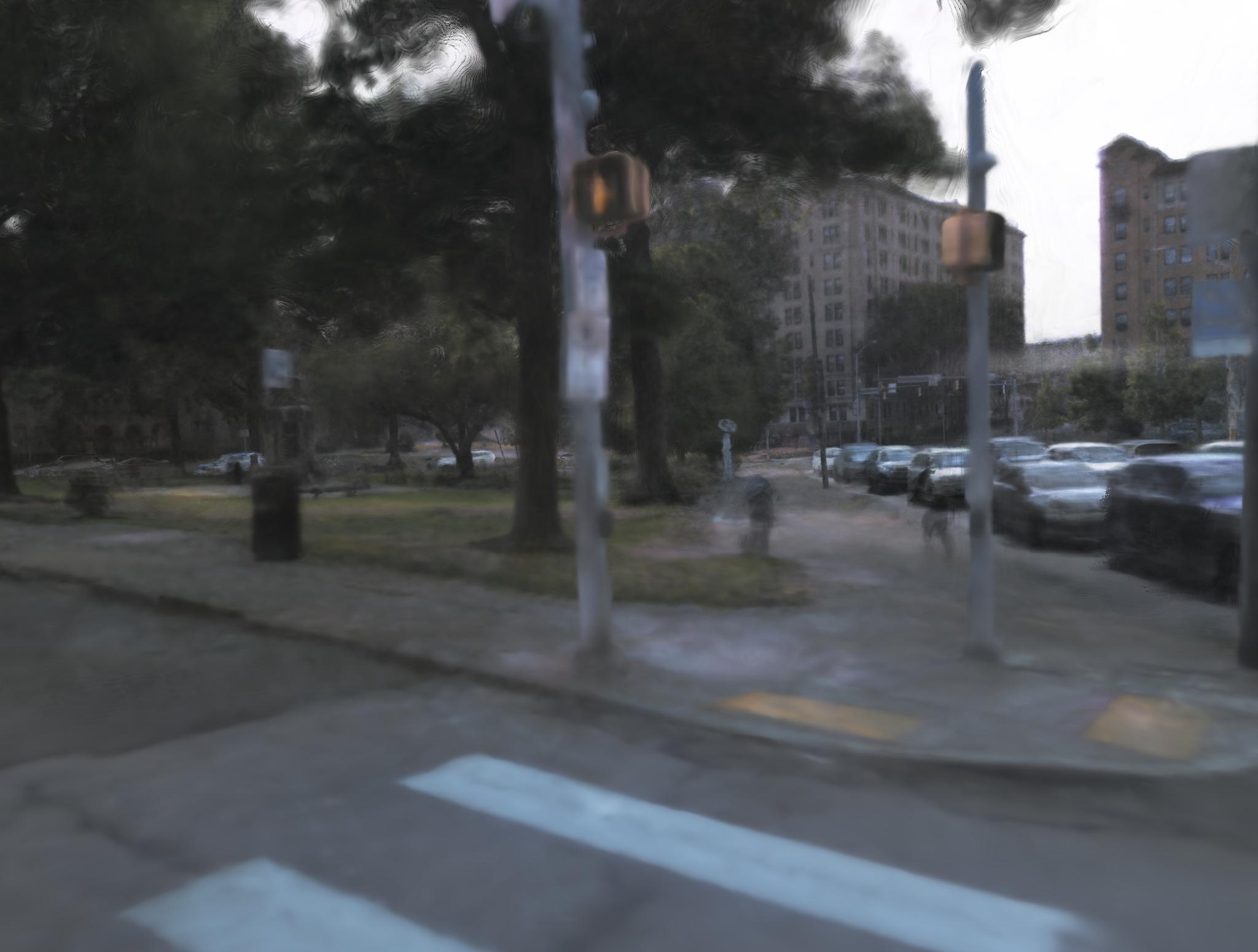} &
\includegraphics[width=0.2\linewidth]{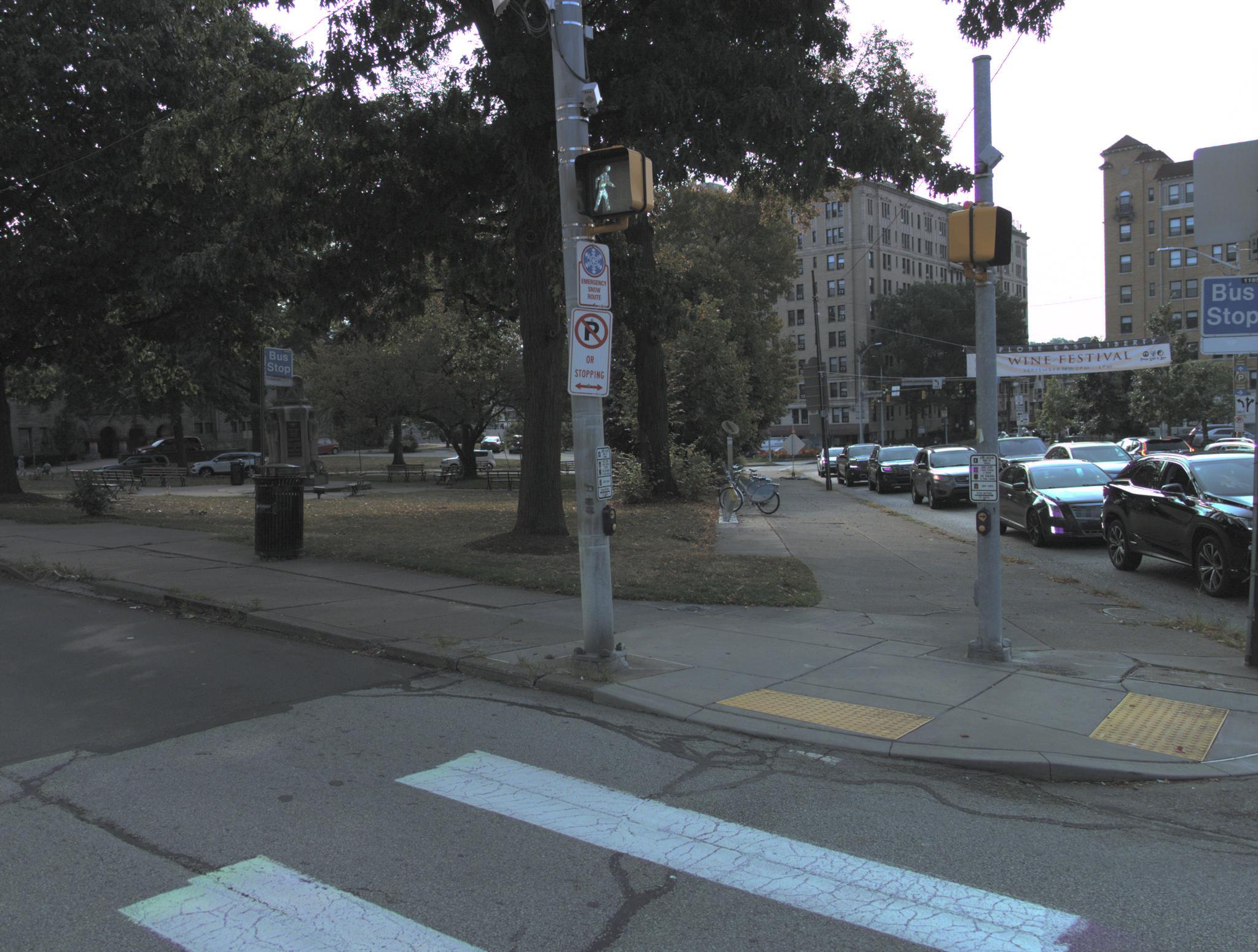} \\
\includegraphics[width=0.2\linewidth]{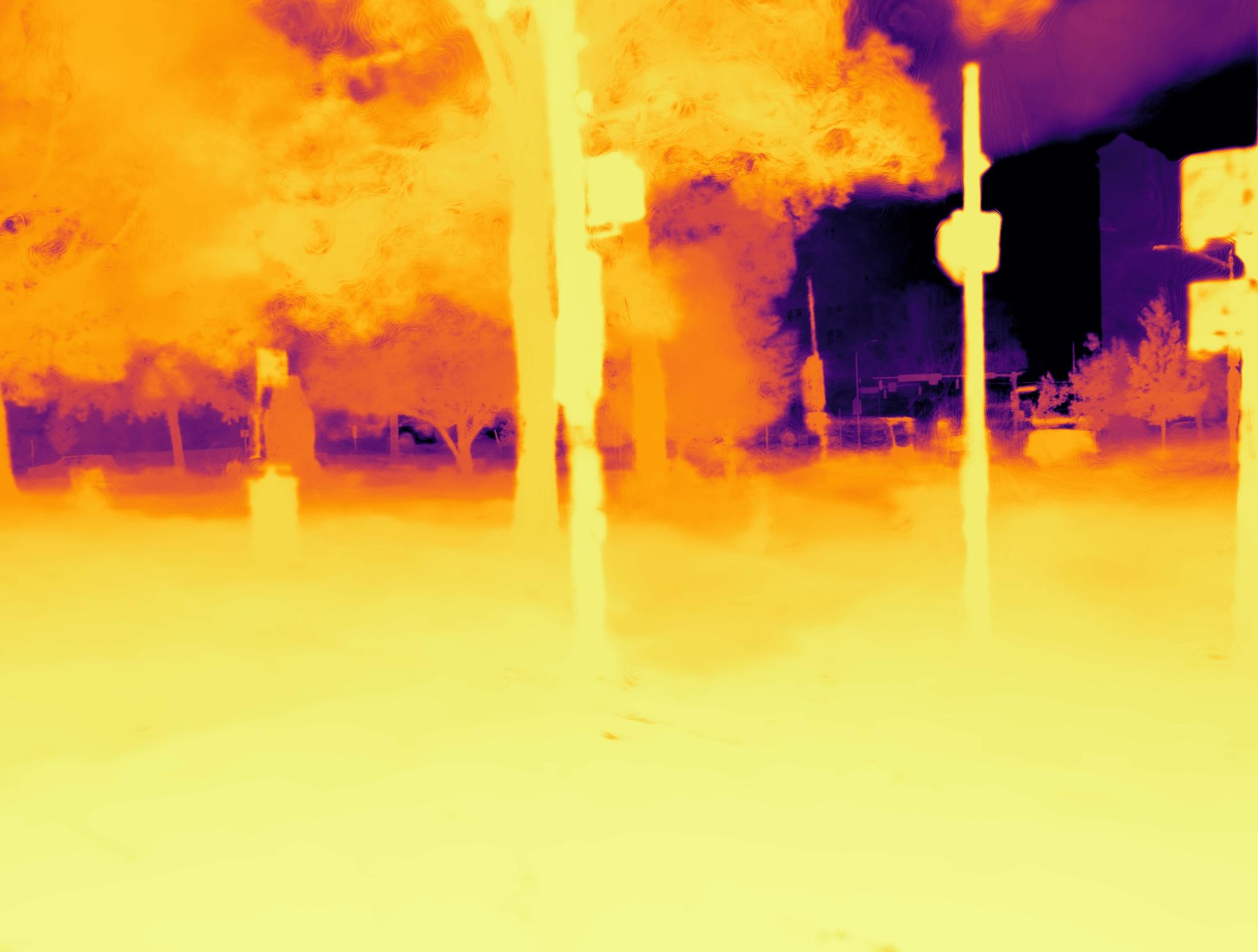} &
\includegraphics[width=0.2\linewidth]{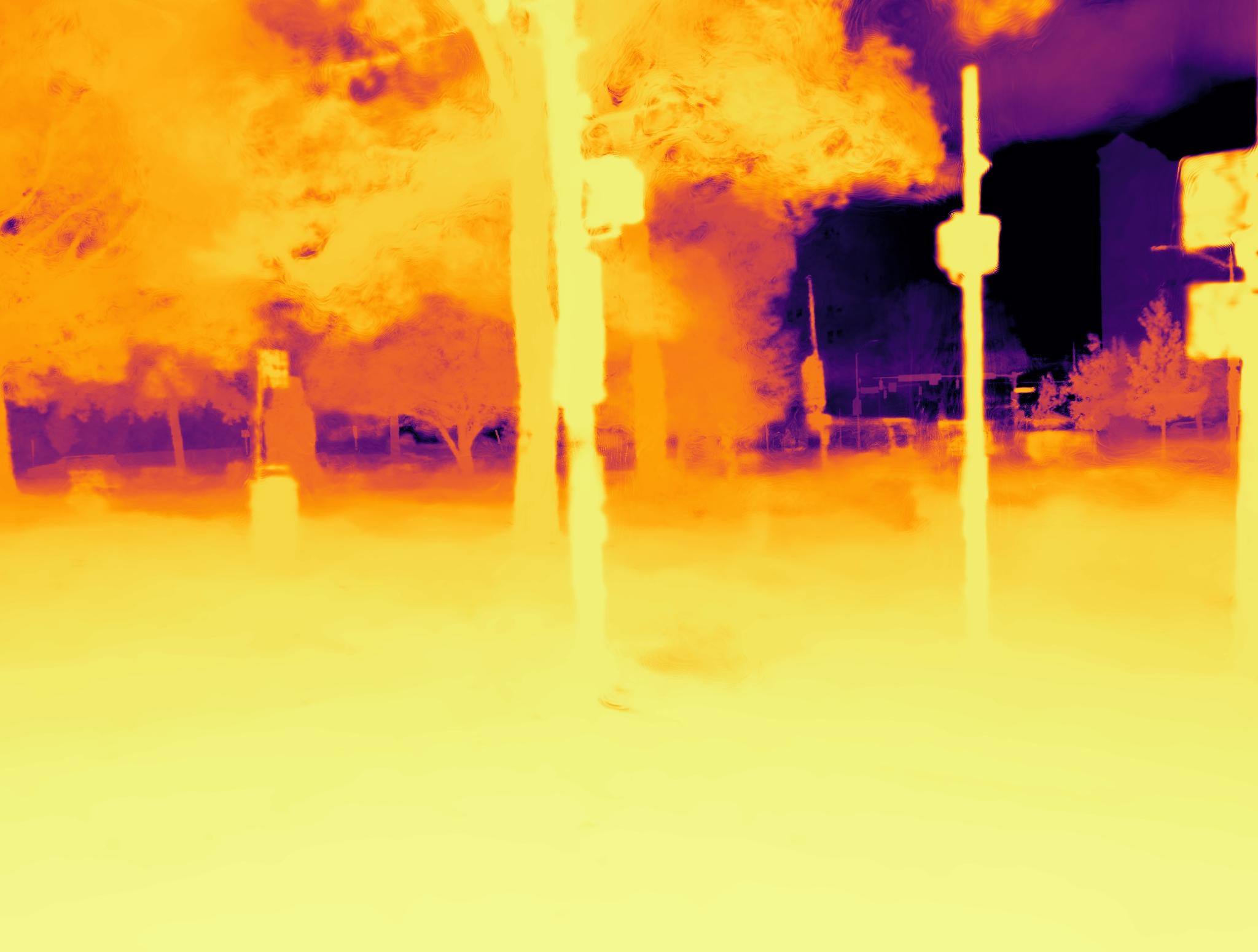} &
\includegraphics[width=0.2\linewidth]{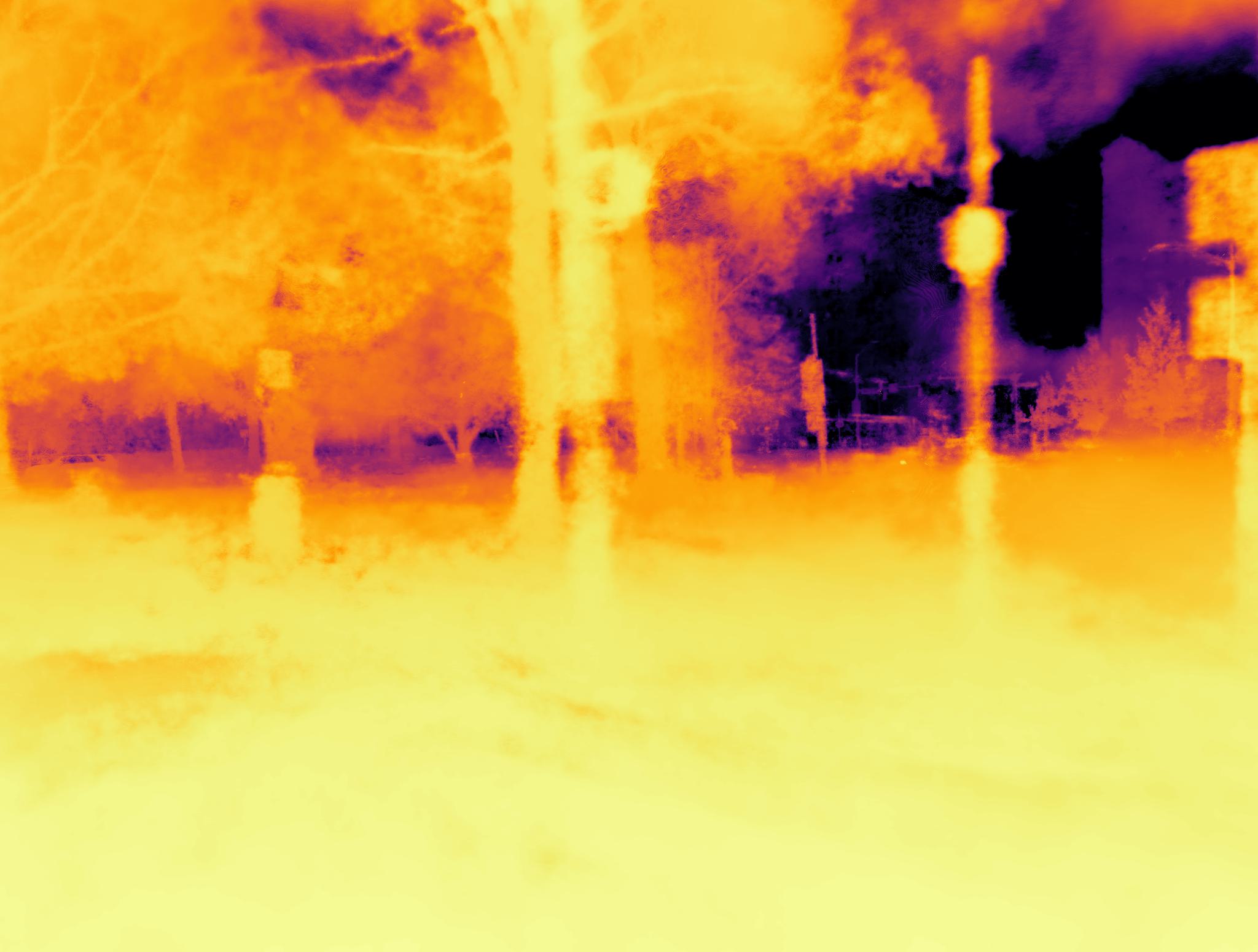} &
\includegraphics[width=0.2\linewidth]{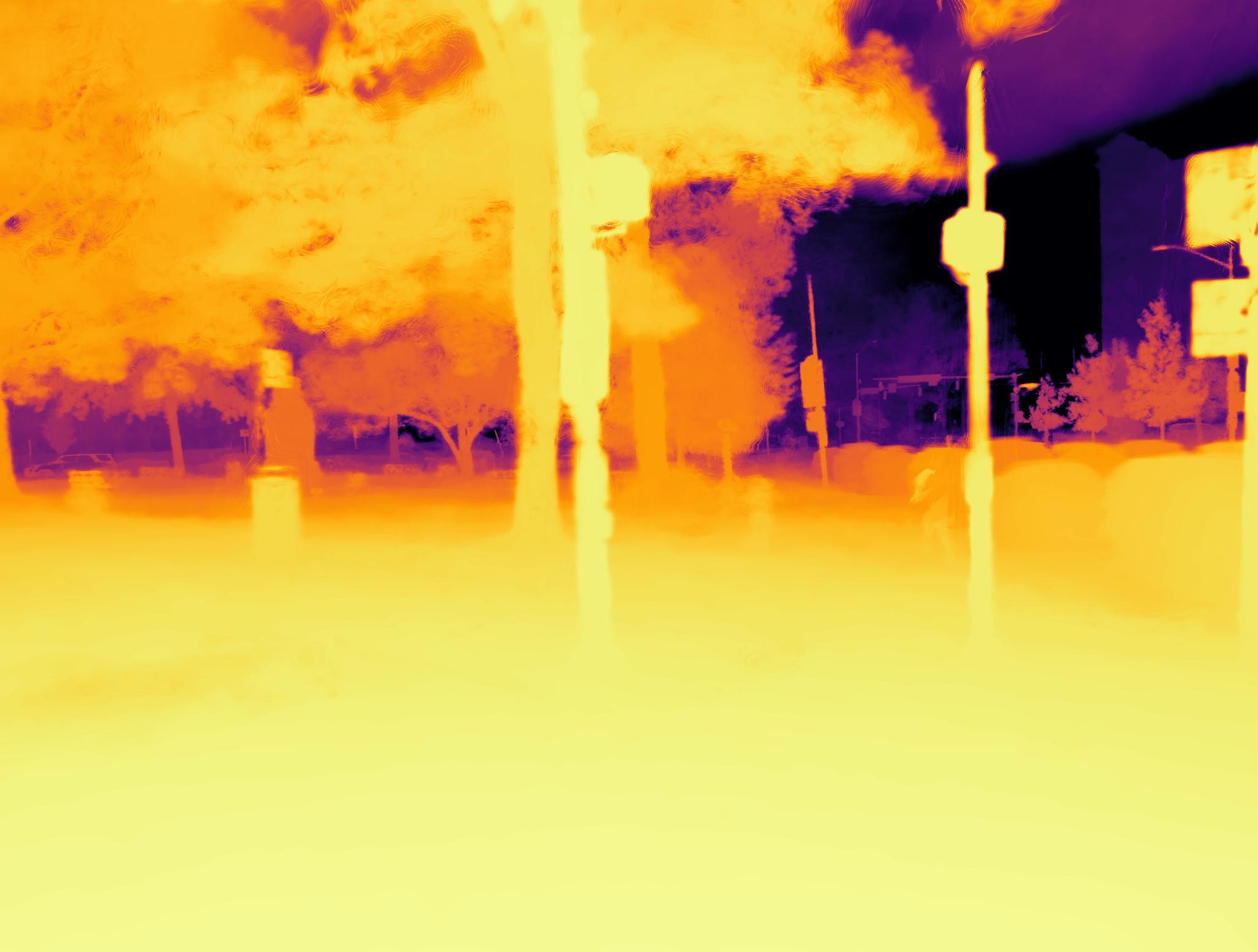} &
\includegraphics[width=0.2\linewidth]{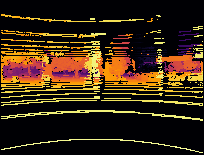} \\

\includegraphics[width=0.2\linewidth]{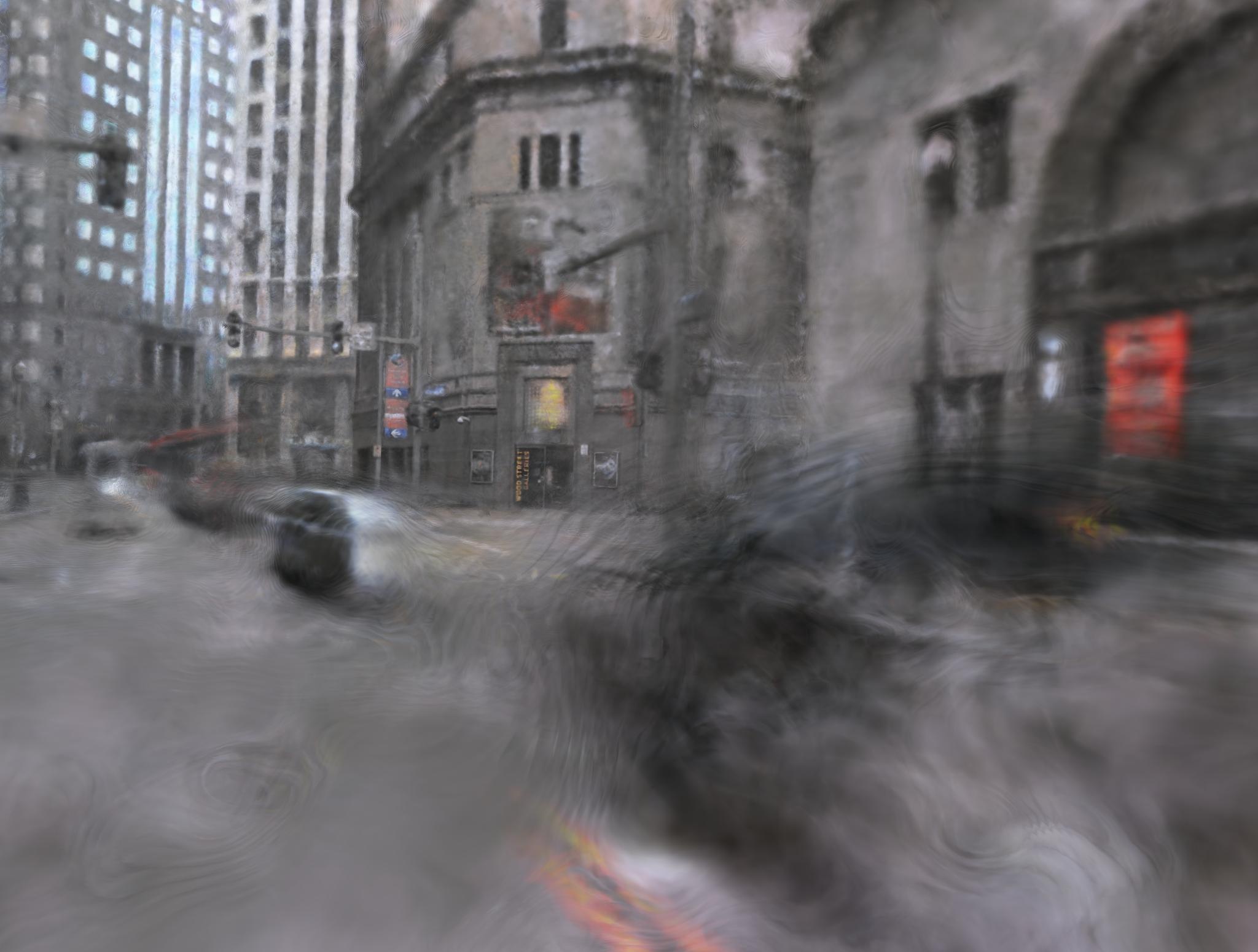} &
\includegraphics[width=0.2\linewidth]{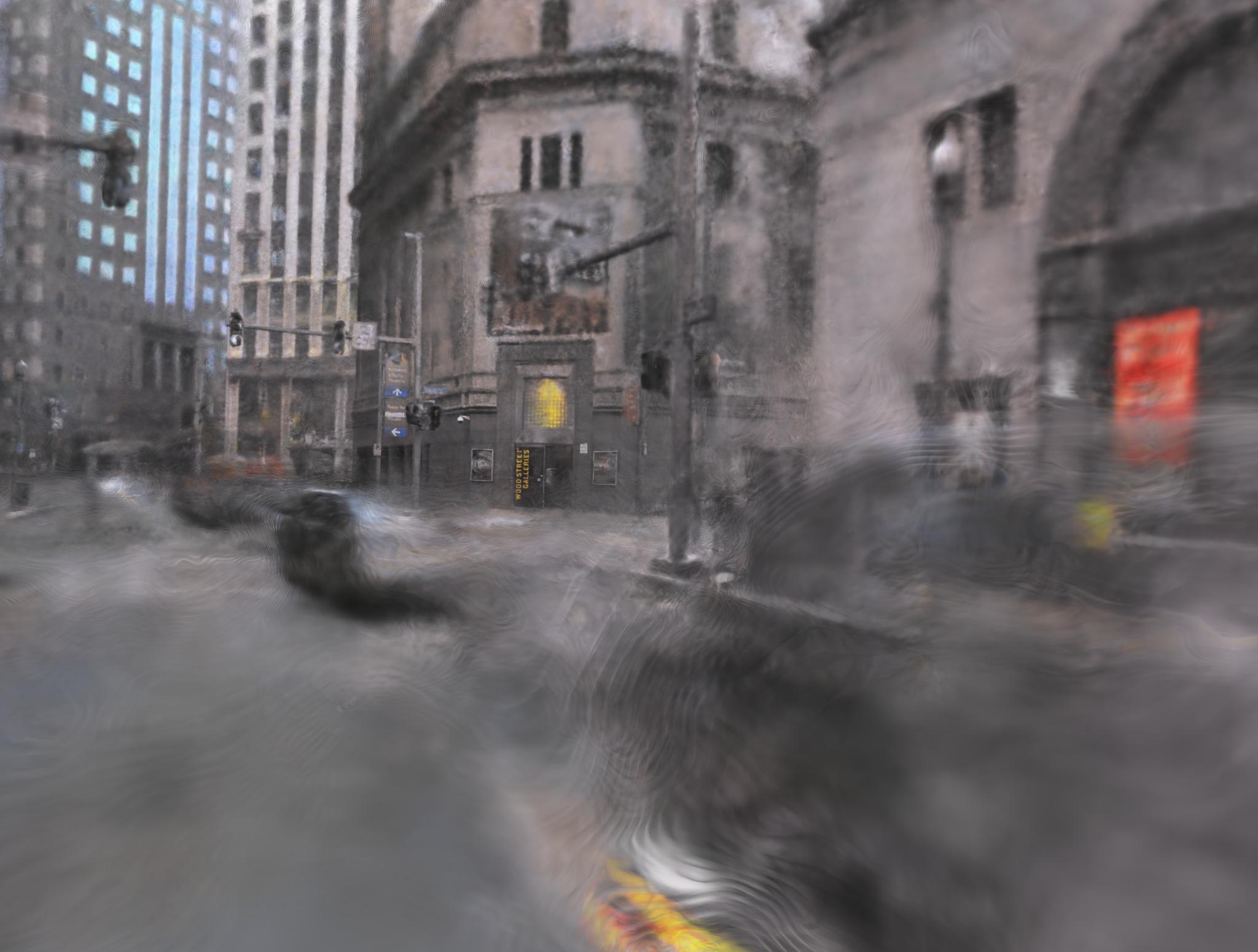} &
\includegraphics[width=0.2\linewidth]{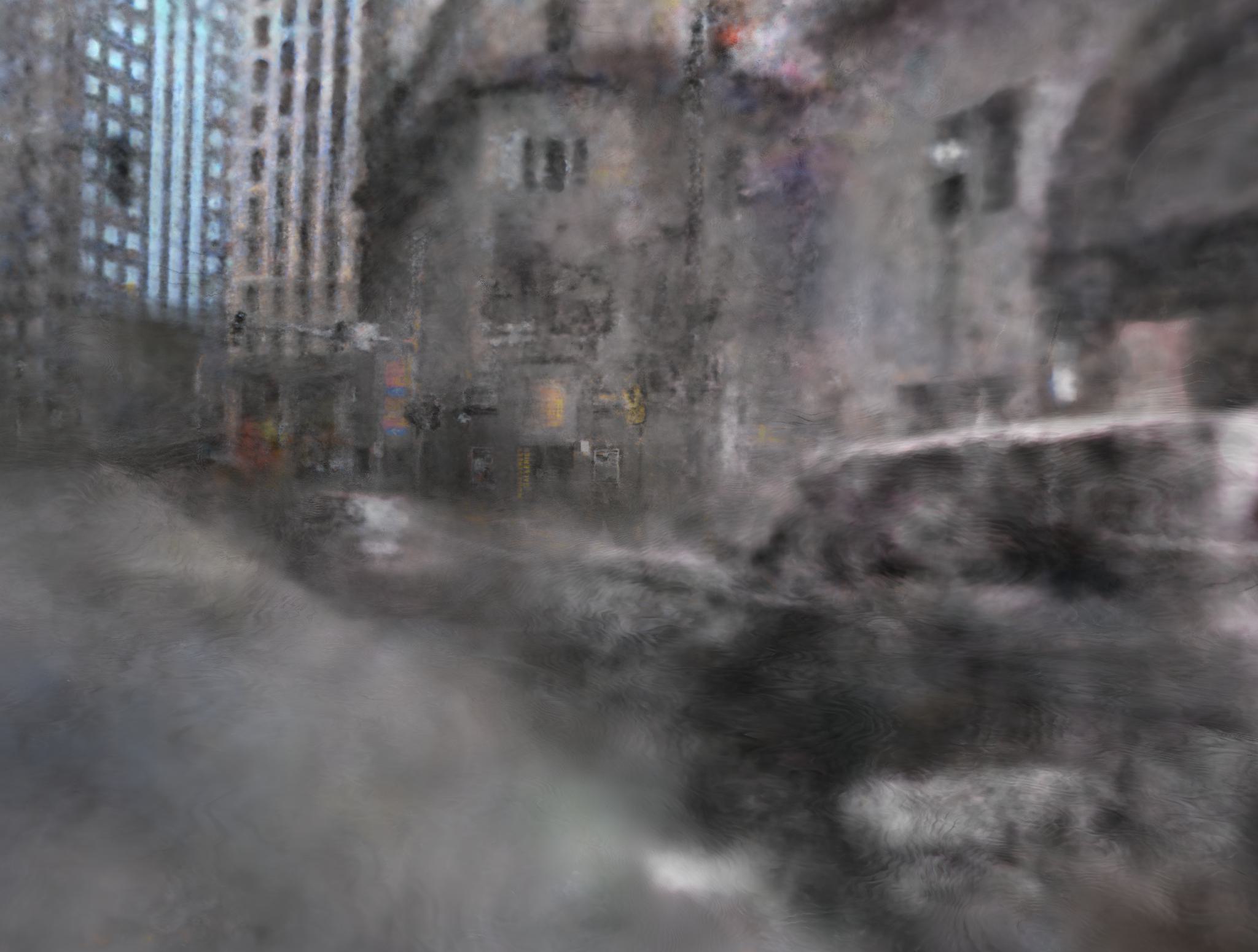} &
\includegraphics[width=0.2\linewidth]{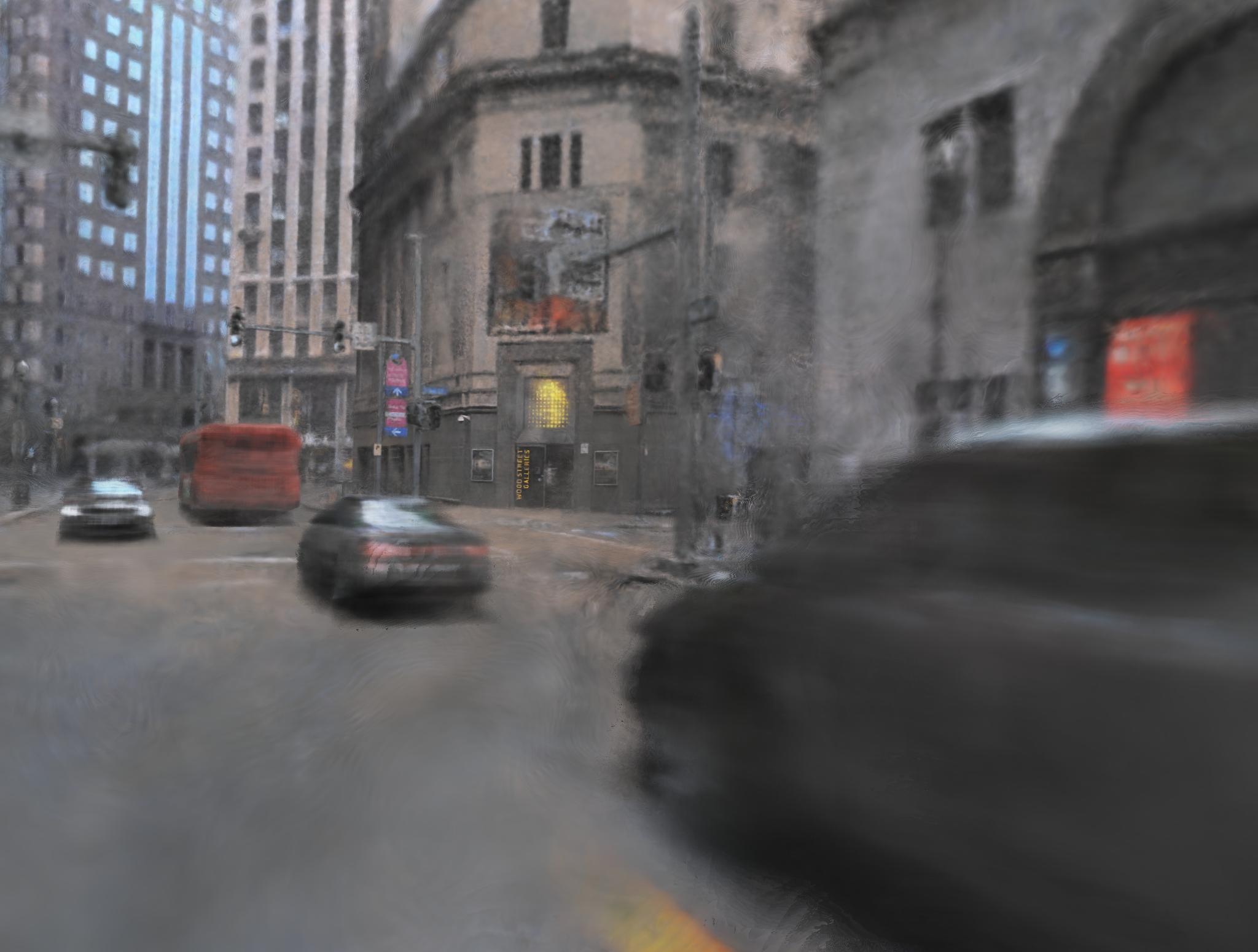} &
\includegraphics[width=0.2\linewidth]{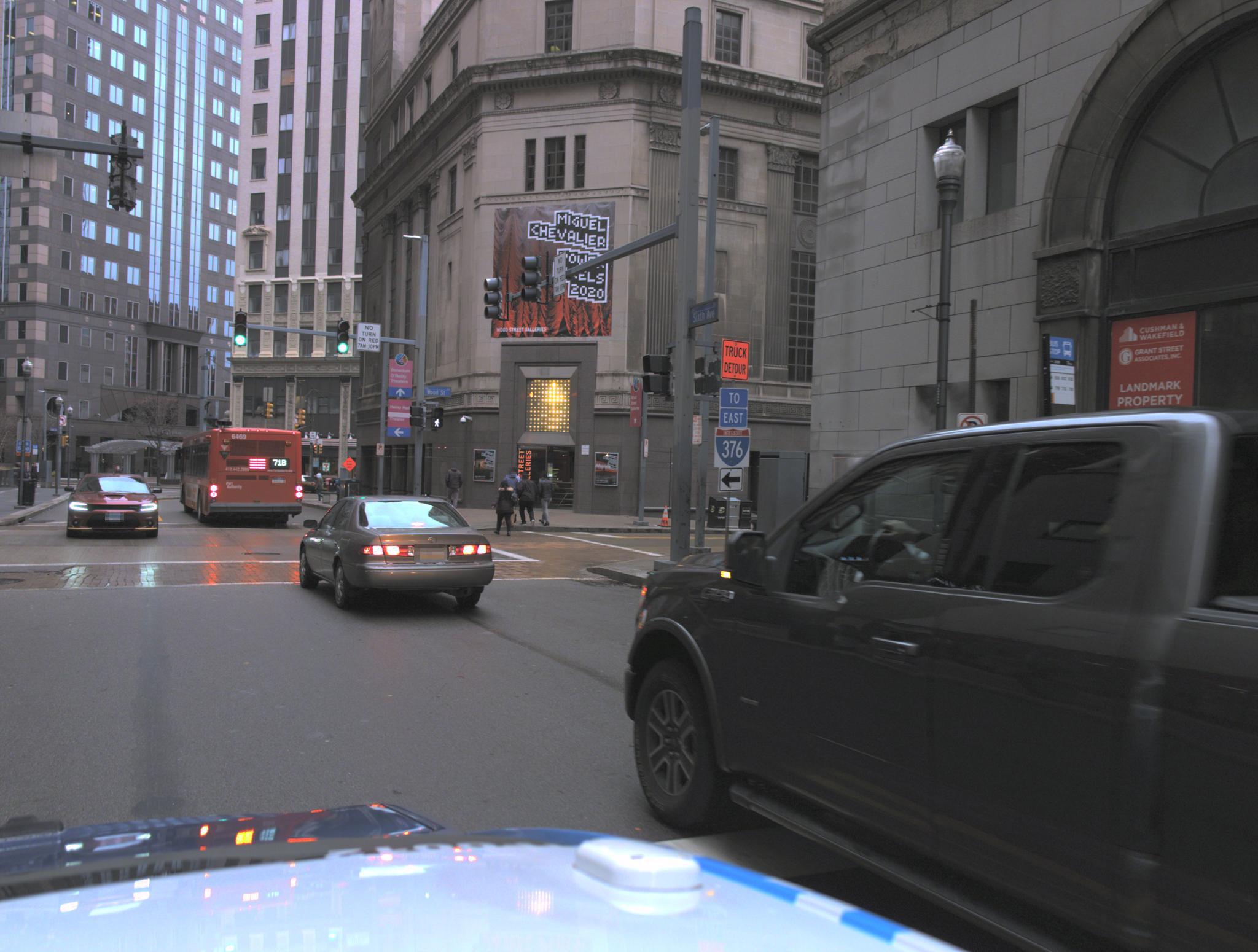} \\
\includegraphics[width=0.2\linewidth]{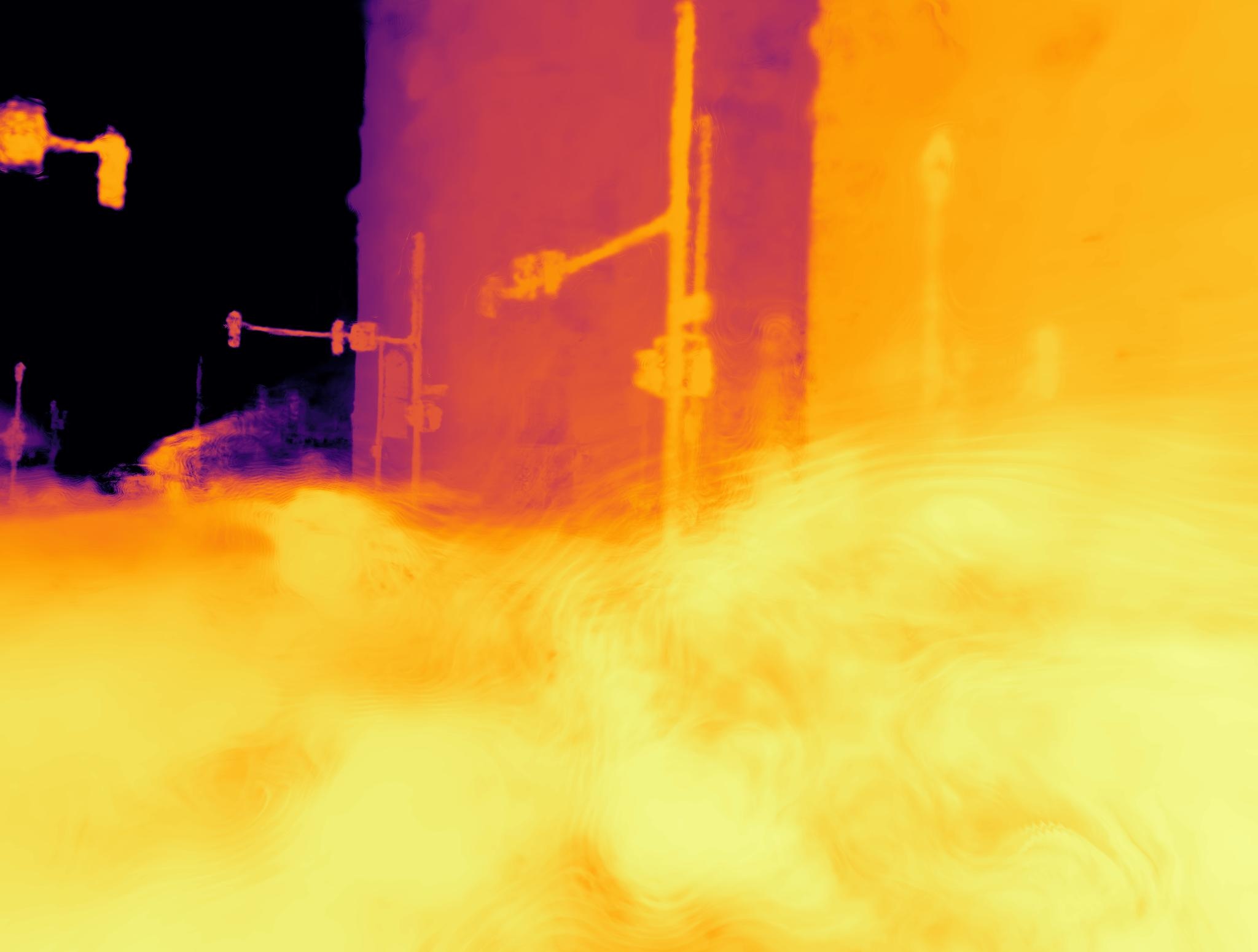} &
\includegraphics[width=0.2\linewidth]{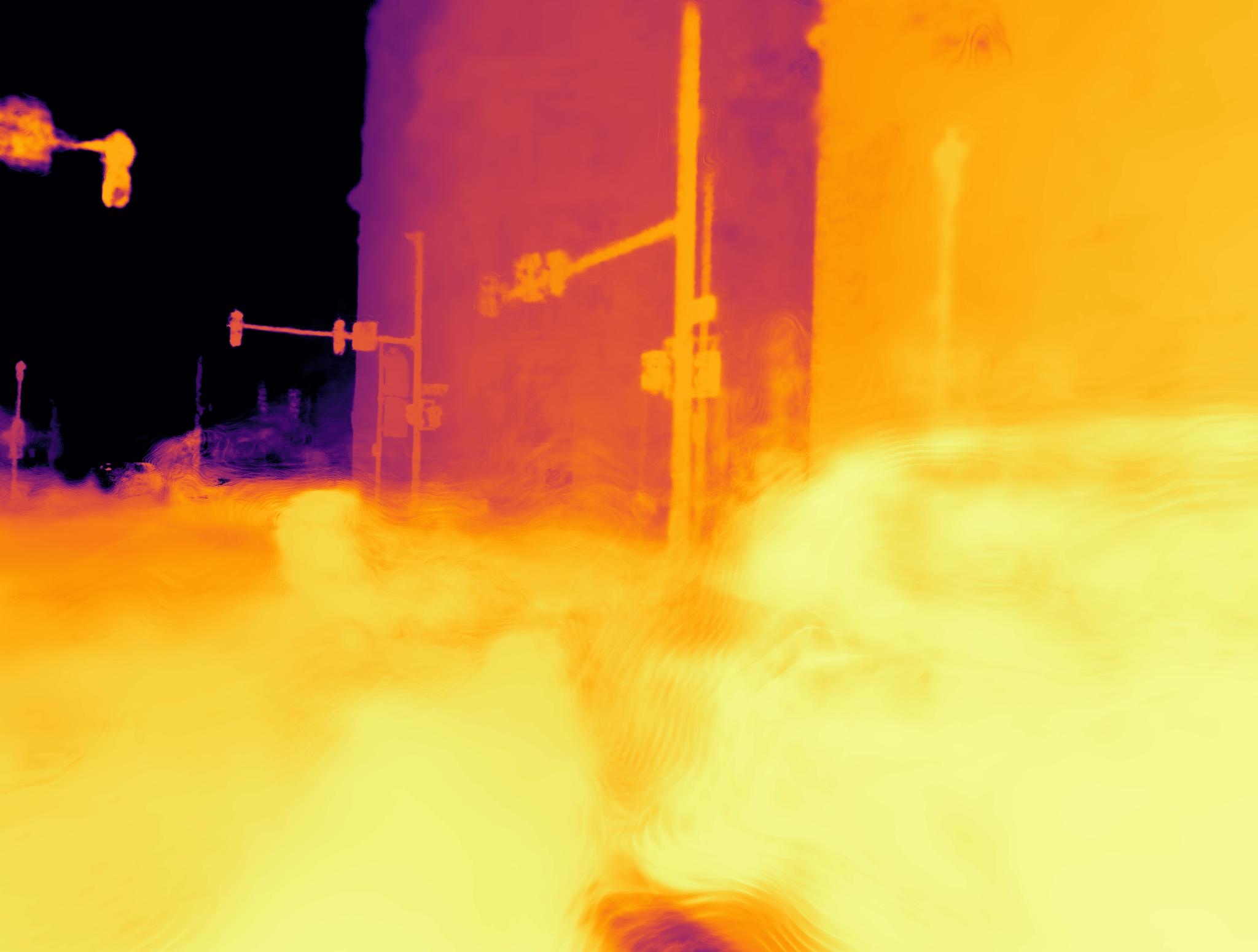} &
\includegraphics[width=0.2\linewidth]{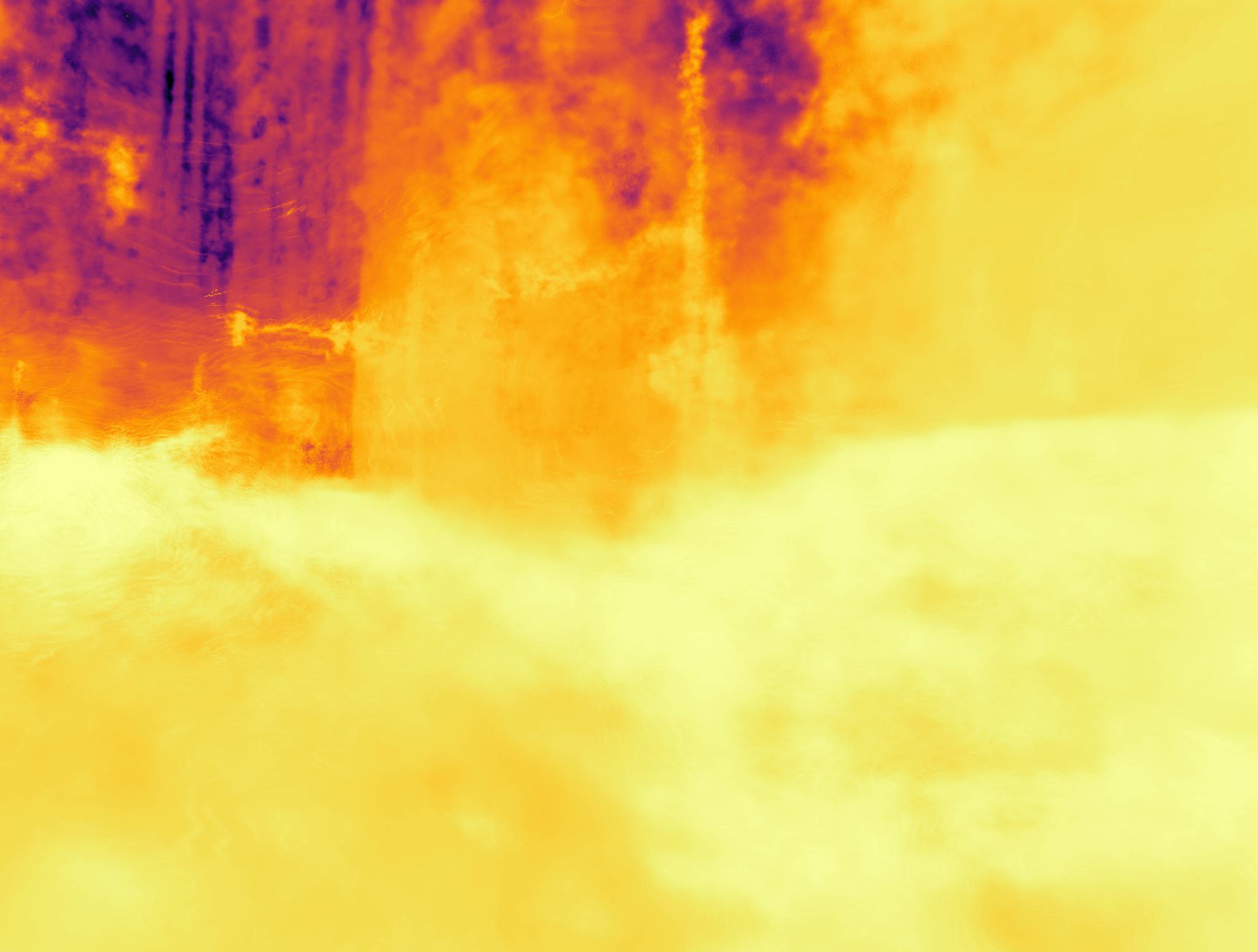} &
\includegraphics[width=0.2\linewidth]{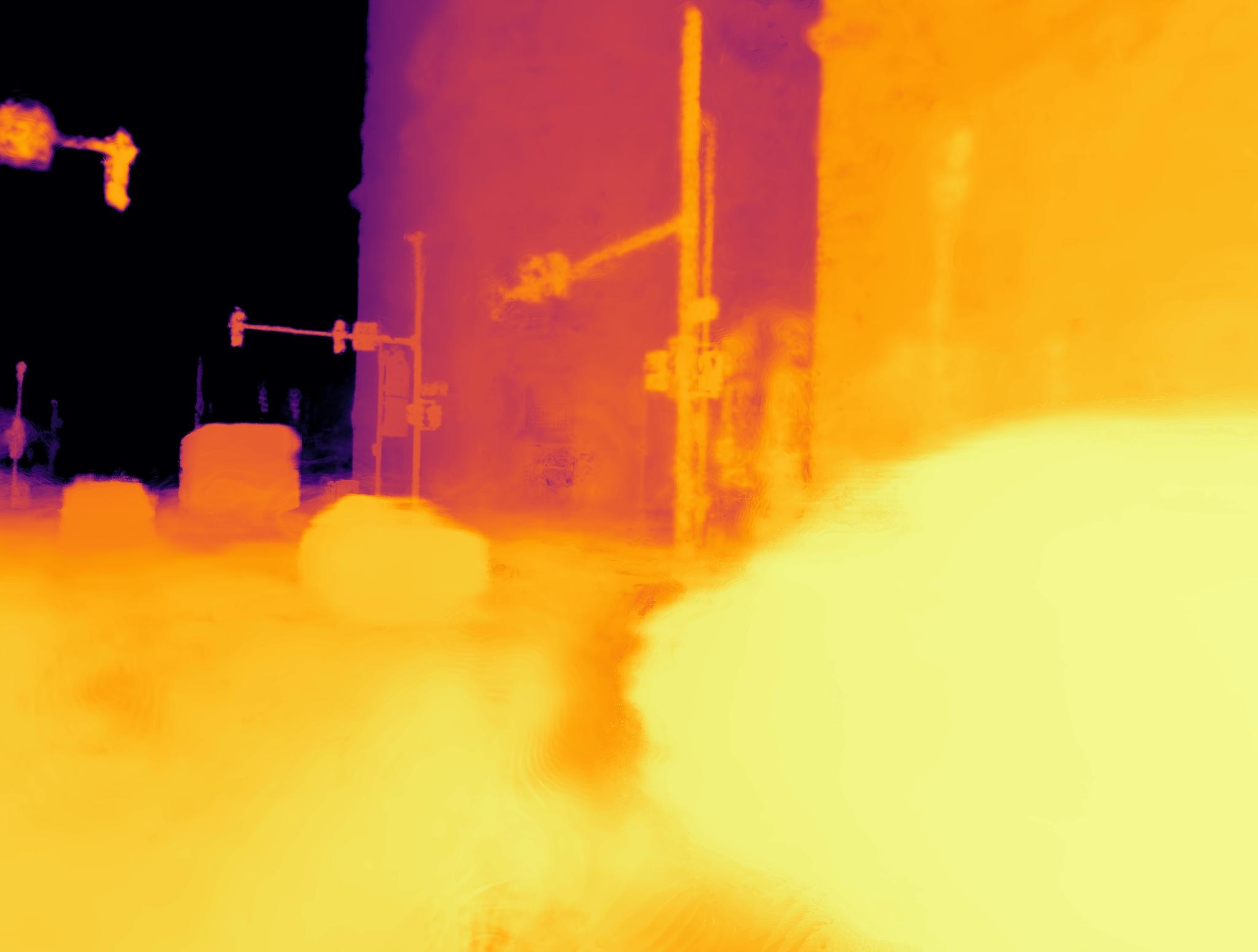} &
\includegraphics[width=0.2\linewidth]{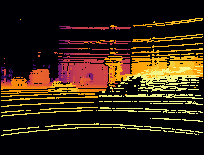} \\

\includegraphics[width=0.2\linewidth]{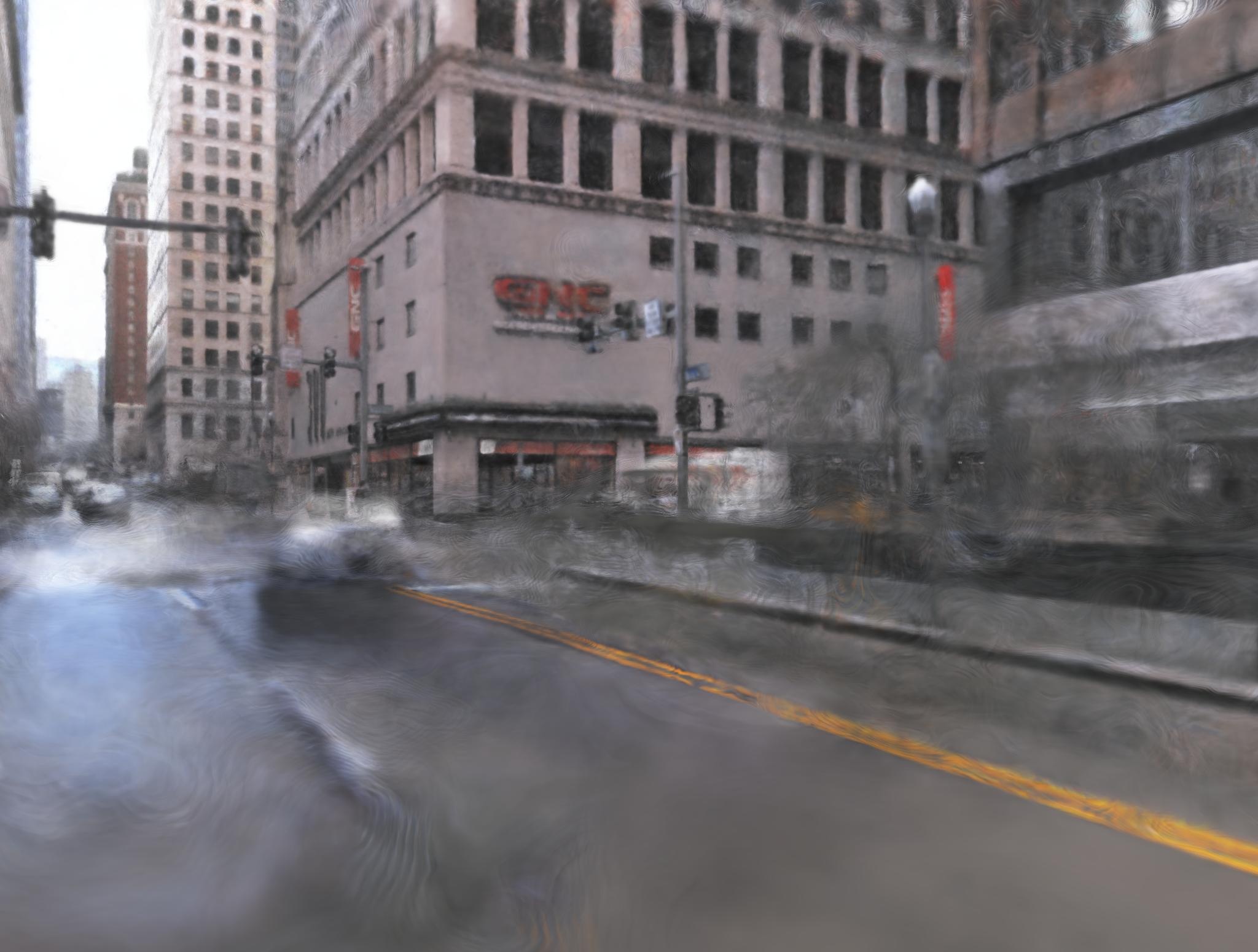} &
\includegraphics[width=0.2\linewidth]{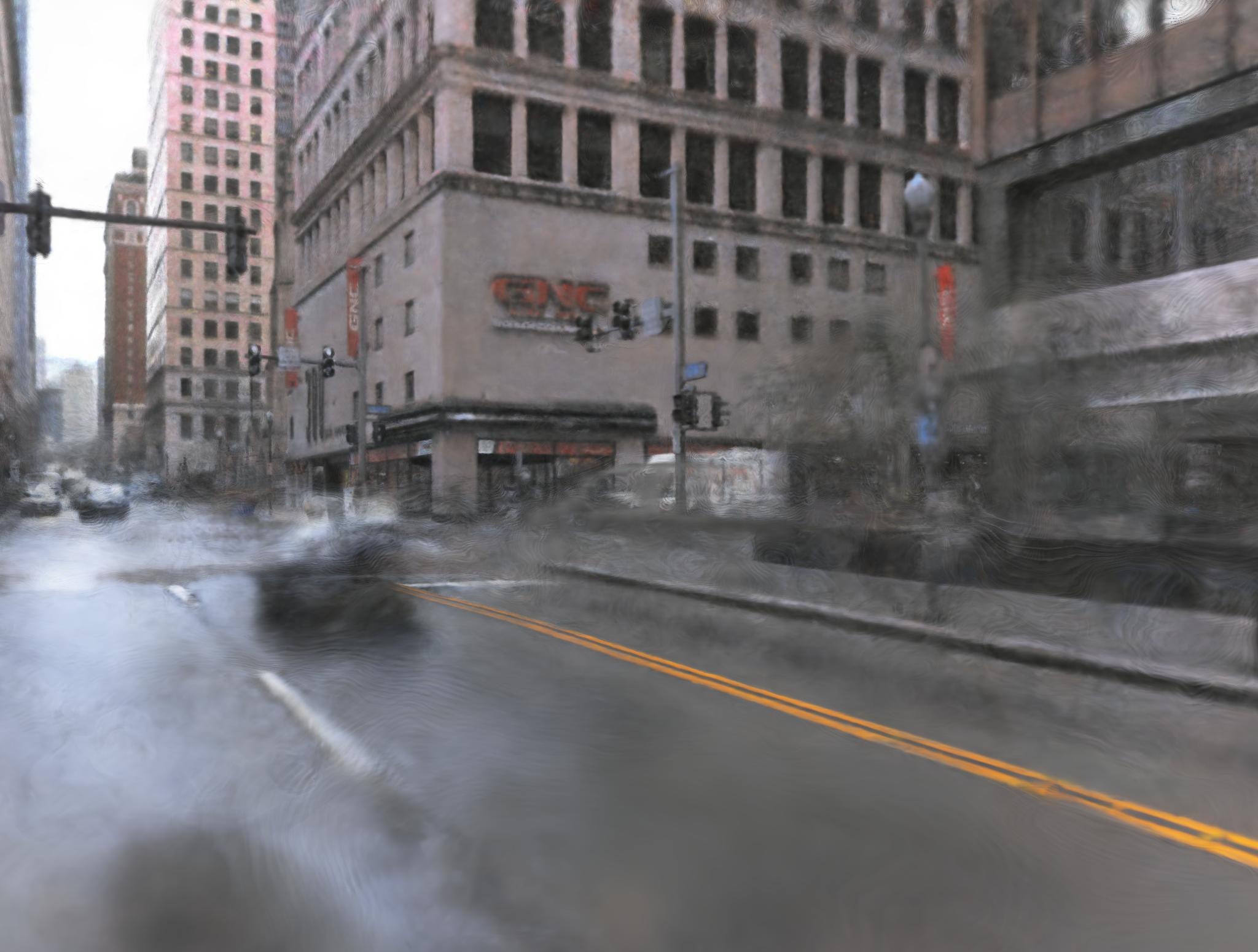} &
\includegraphics[width=0.2\linewidth]{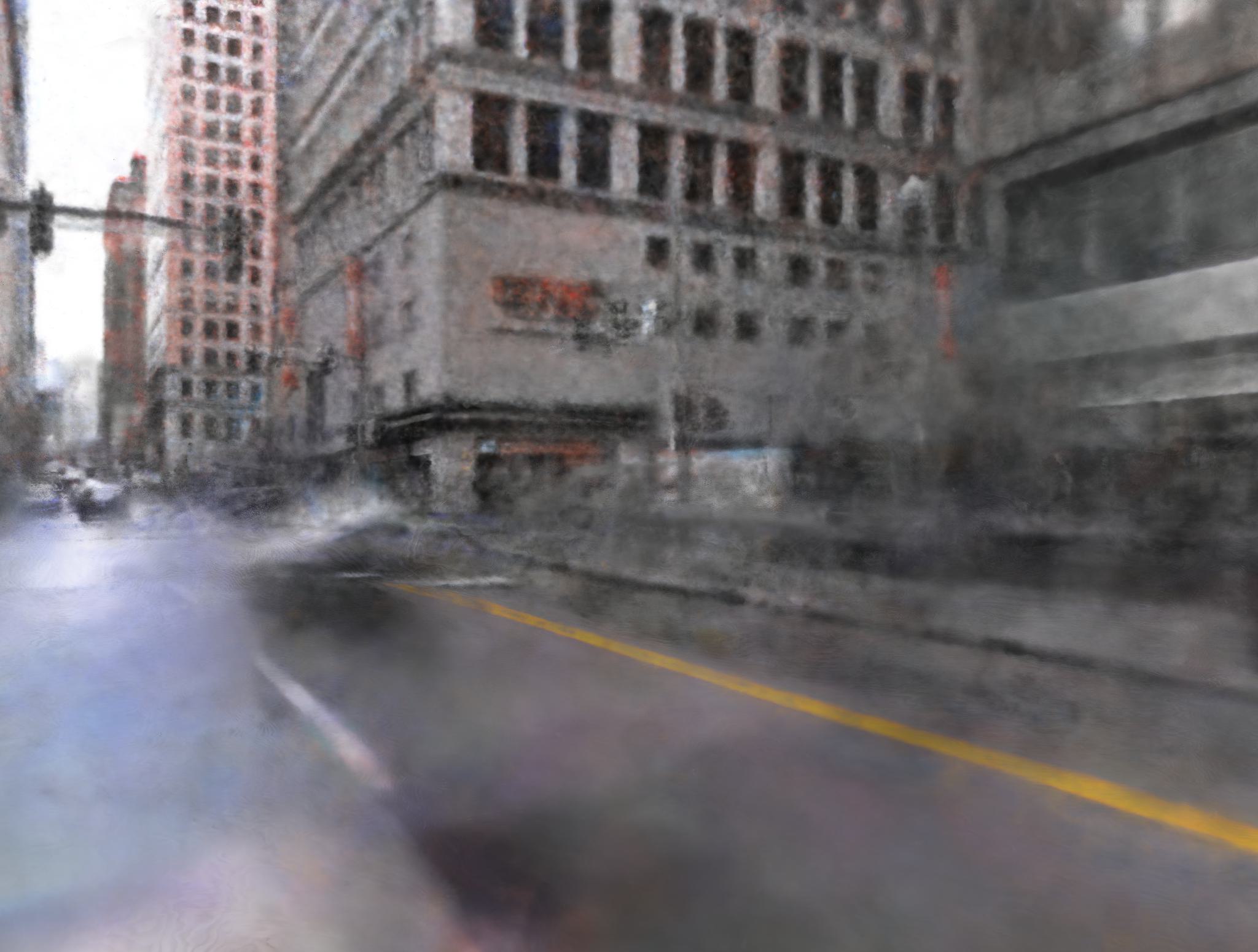} &
\includegraphics[width=0.2\linewidth]{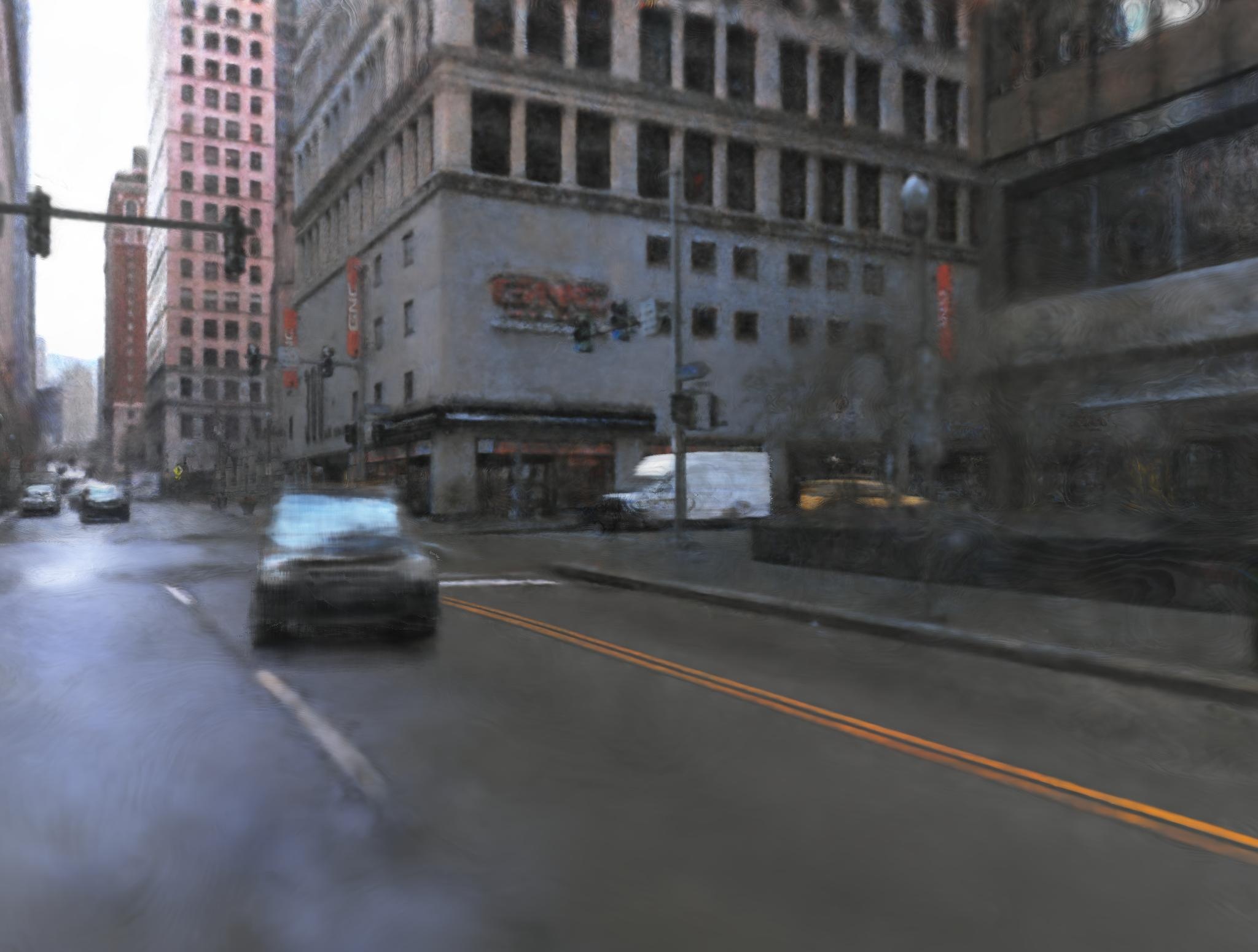} &
\includegraphics[width=0.2\linewidth]{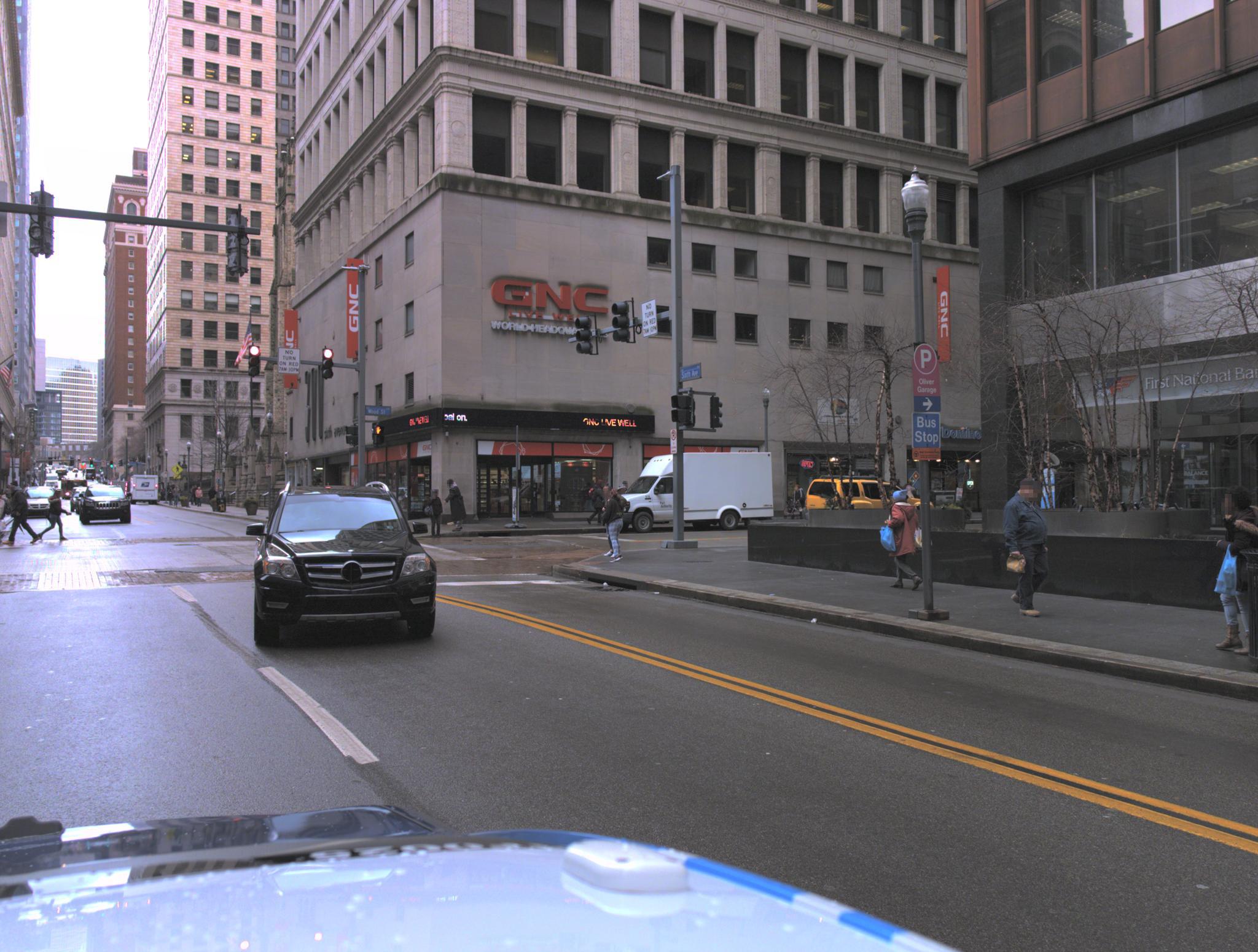} \\
\includegraphics[width=0.2\linewidth]{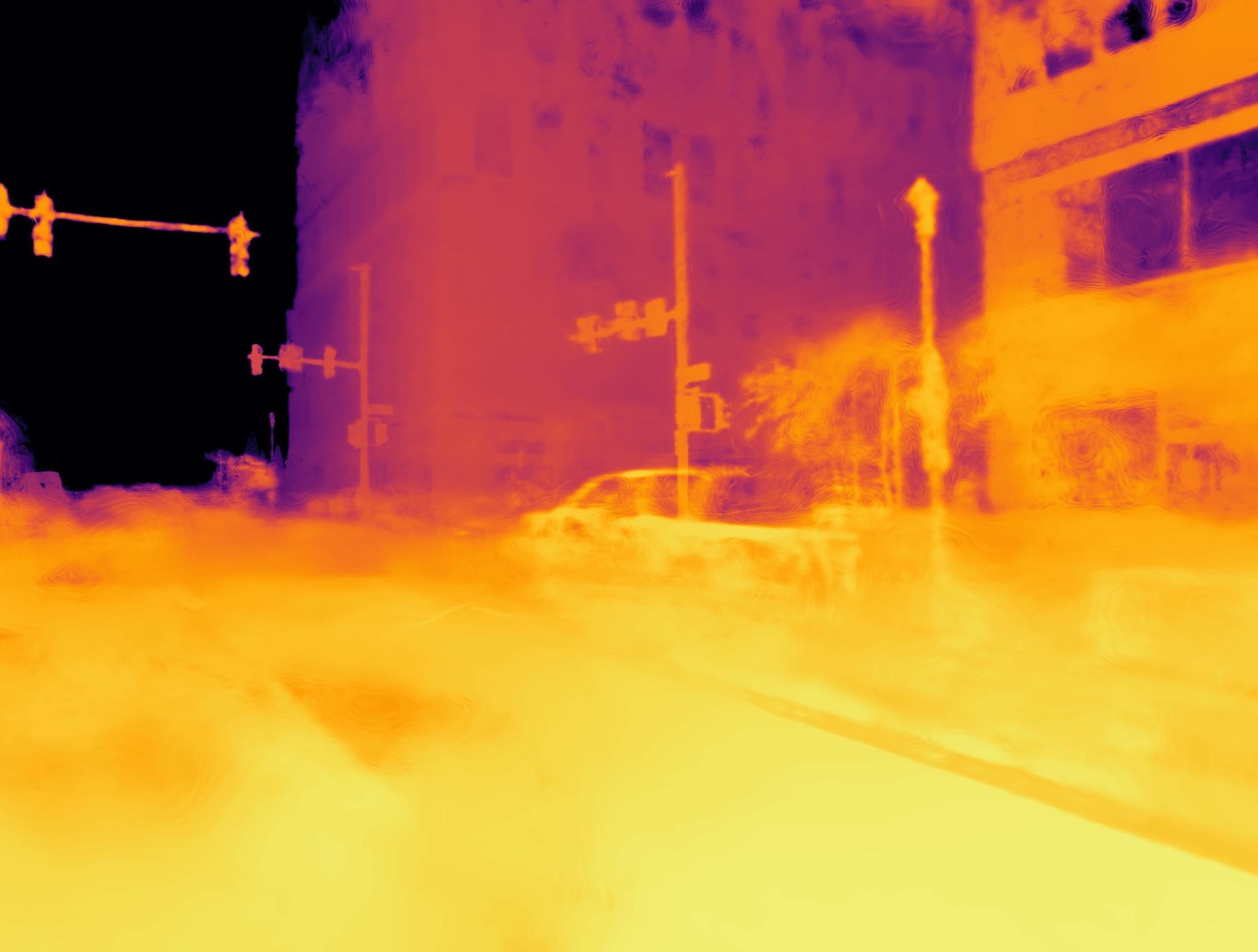} &
\includegraphics[width=0.2\linewidth]{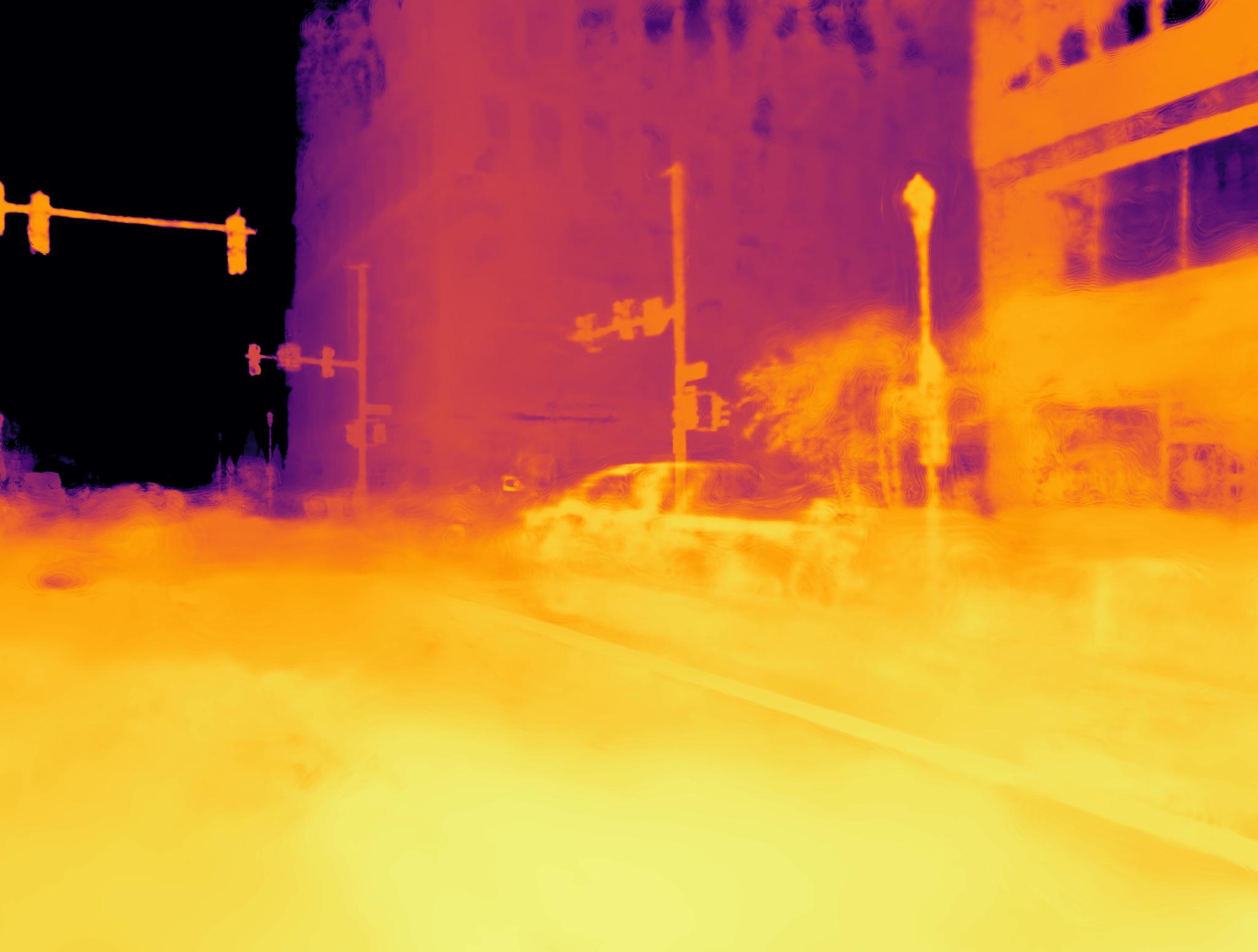} &
\includegraphics[width=0.2\linewidth]{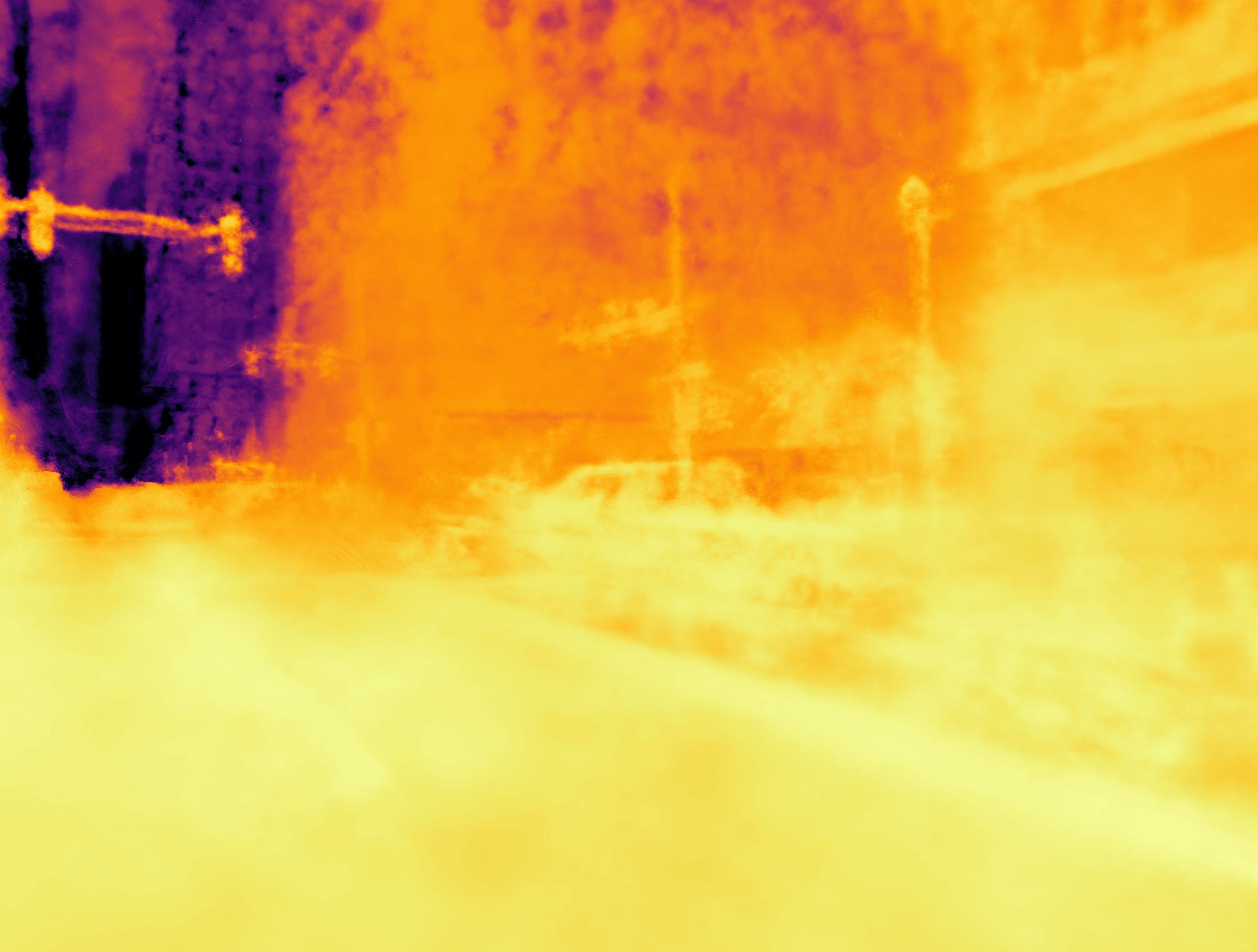} &
\includegraphics[width=0.2\linewidth]{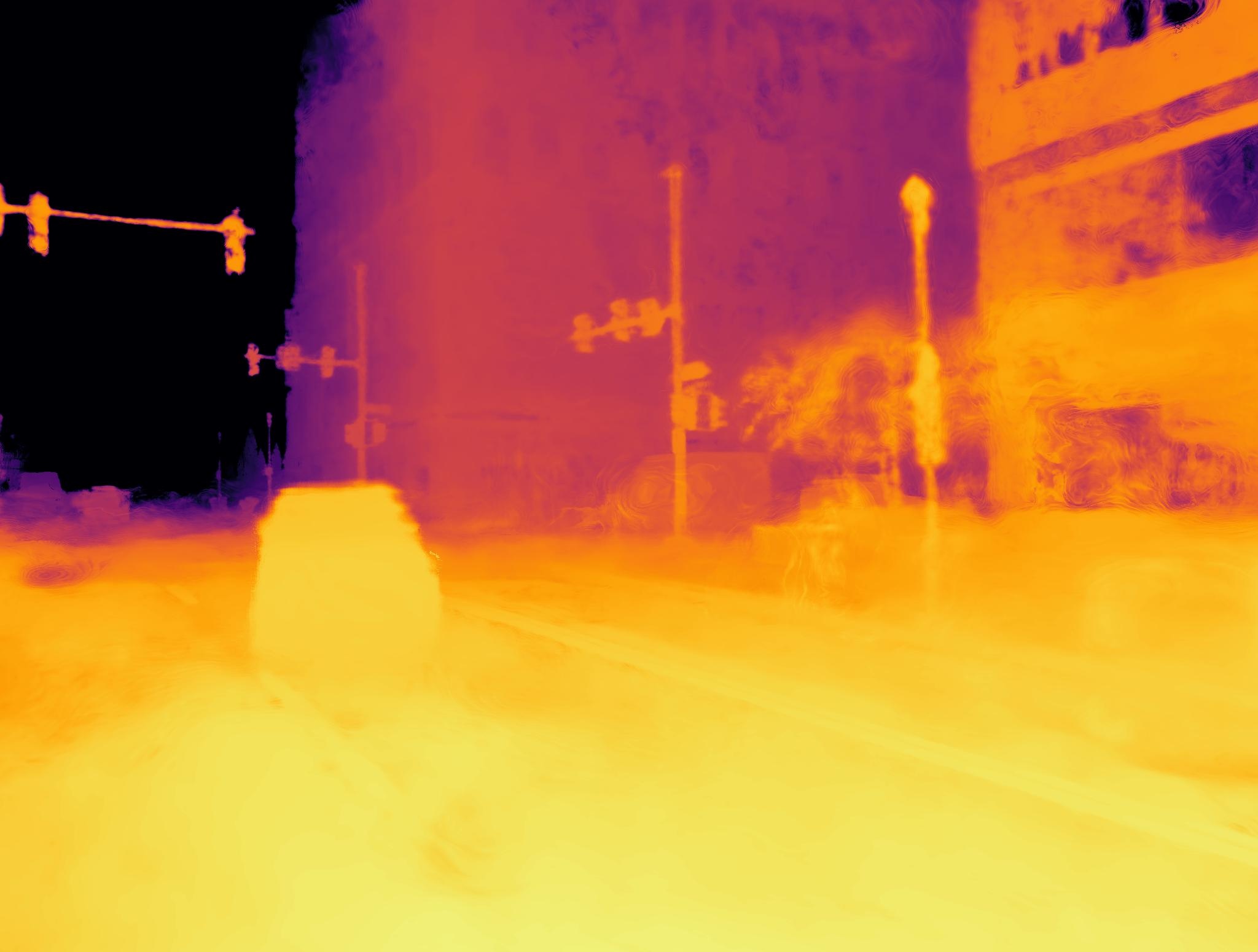} &
\includegraphics[width=0.2\linewidth]{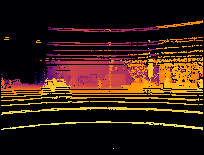} \\ \bottomrule
\end{tabular}
}\vspace{-2mm}
\caption{\textbf{Additional qualitative results on Argoverse 2.} We illustrate four examples, where the upper two are from the residential area and the lower two are from the downtown area in our benchmark. We observe that our method exhibits better view quality and cleaner depth maps, particularly in dynamic areas.}
\label{fig:av2_qualitative_supp}
\end{figure*}

\parsection{Additional qualitative comparison.}
We include further qualitative results of our method compared to the state-of-the-art in Fig.~\ref{fig:av2_qualitative_supp}. We observe that, similar to Fig,~\ref{fig:av2_qualitative} of the main paper, our method exhibits superior view synthesis of dynamic areas and better captures seasonal variations in terms of, for example, tree leaves. Further, the synthesized views of our method are sharper compared to prior art, and the depth maps are less noisy. We include qualitative results from both the residential and downtown areas of our benchmark.

\end{document}